\RequirePackage{fix-cm}
\documentclass[smallextended]{svjour3}       
\smartqed  
\usepackage{caption}
\usepackage{subcaption}
\usepackage{etex}
\usepackage{natbib}
\usepackage{rotating}
\usepackage{eqnarray} 
\usepackage{array,multirow,booktabs} 
\usepackage{enumitem} 
\usepackage{amsfonts,bbm,bm,courier} 

\usepackage{mathtools}
\usepackage{pdflscape} 
\usepackage{graphicx,epstopdf} 
\usepackage[graphicx]{realboxes}
\usepackage{tikz,pgfplots} 
\usetikzlibrary{shapes,arrows}

\newcommand{\lzero}{\ell_0}
\newcommand{\lone}{\ell_1}

\newcommand{\for}{\textnormal{ for }}

\newcommand{\lambdab}{\bm{\lambda}}

\newcommand{\xb}{\bm{x}}
\newcommand{\Lset}{\mathcal{L}}
\newcommand{\X}{\mathcal{X}}
\newcommand{\Y}{\mathcal{Y}}

\newcommand{\B}{\{0,1\}}

\newcommand{\Z}{\mathbb{Z}}
\newcommand{\R}{\mathbb{R}}

\newcommand{\vnorm}[1]{\left\|#1\right\|}
\newcommand{\indic}[1]{\mathbbm{1}[#1]}
\newcommand{\wplus}{W^{+}} 
\newcommand{\wminus}{W^{-}}
 
\newcommand{\mprange}[3]{{#1}={#2}\textnormal{,...,}{#3}}
\newcommand{\mpdes}[1]{\textit{\tiny #1}}

\newcommand{\breastcancer}{\texttt{breastcancer}}
\newcommand{\haberman}{\texttt{haberman}}
\newcommand{\internetad}{\texttt{internetad}}
\newcommand{\mammo}{\texttt{mammo}}

\newcommand{\spambase}{\texttt{spambase}}
\newcommand{\tictactoe}{\texttt{tictactoe}}
\newcommand{\violentcrime}{\texttt{violentcrime}}

\newcommand{\pkg}[1]{{\fontseries{b}\selectfont #1}} 

\begin{document}

\title{Supersparse Linear Integer Models for Interpretable Classification}

\titlerunning{SLIM for Interpretable Classification}        

\author{Berk Ustun\and
        Stefano Trac\`a\and 
        Cynthia Rudin
}

\authorrunning{Ustun, Trac\`a and Rudin} 

\institute{   Berk Ustun \at
              Department of Electrical Engineering and Computer Science \\ 
              Massachusetts Institute of Technology \\
              \email{ustunb@mit.edu}           
           \and
           Stefano Trac\`a \at
              Operations Research Center \\ 
              Massachusetts Institute of Technology \\
              \email{stet@mit.edu}    
               \and
           Cynthia Rudin \at
              Sloan School of Management \\ 
              Massachusetts Institute of Technology \\
              \email{rudin@mit.edu}        
}
\date{Last Updated: \today}

\maketitle

\begin{abstract}
Scoring systems are classification models that only require users to add, subtract and multiply a few meaningful numbers to make a prediction. These models are often used because they are practical and interpretable. In this paper, we introduce an off-the-shelf tool to create scoring systems that both accurate and interpretable, known as a Supersparse Linear Integer Model (SLIM). SLIM is a discrete optimization problem that minimizes the 0-1 loss to encourage a high level of accuracy, regularizes the $\ell_0$-norm to encourage a high level of sparsity, and constrains coefficients to a set of interpretable values. We illustrate the practical and interpretable nature of SLIM scoring systems through applications in medicine and criminology, and show that they are are accurate and sparse in comparison to state-of-the-art classification models using numerical experiments.
\end{abstract}

\hbox{\hspace{\textwidth}}
\hbox{\textbf{Keywords}: machine learning, data mining, interpretability, classification, scoring systems}
\hbox{\hspace{\textwidth}}

\clearpage

\section{Introduction}\label{Sec::Introduction}

Scoring systems are classification models that make predictions using a sparse linear combination of variables with integer coefficients. These systems are widely used in our society: to assess the risk of medical outcomes in hospitals \citep{antman2000timi, morrow2000timi, gage2001validation}; to predict the incidence of violent crimes \citep{andrade2009handbook,steinhart2006juvenile}; to gauge marine safety for military vessels \citep{abs2002marine}; and to rate business schools \citep{business}. 

On one hand, the popularity of scoring systems reflects their \textit{practicality}: scoring systems only require users to add, subtract, and multiply a few numbers to make a prediction. This allows for quick, comprehensible predictions, without the use of a computer, and without formal training in statistics. On the other hand, the popularity of scoring systems is inherently tied to their \textit{interpretability} (i.e. the fact that they are easy to understand). Interpretability is an essential component of applied predictive modeling: domain experts intrinsically dislike black-box predictive models as they would like to understand how a model makes its predictions. This transparency is useful in that it can identify important factors and help troubleshoot faulty models. Nevertheless, the true value of interpretability lies in the fact that a model that is easy to understand is also far more likely to yield insights into datasets and be used in practice. 

Recent research in statistics and machine learning has primarily focused on designing accurate and scalable black-box models to address complex problems such as spam prediction and computer vision, in which computers must generate quick and accurate predictions on a massive scale. In turn, the goal of creating interpretable models -- once recognized as a holy grail in the fields of expert systems and artificial intelligence -- has mostly been neglected over the last two decades. One reason for this oversight is the belief that there is necessarily a trade-off between accuracy and interpretability. Even if this were the case, there remains a strong societal need for a formalized approach to interpretable classification -- evidenced by the fact that many popular medical scoring systems were built using heuristic procedures that did not optimize for accuracy \cite{le1993new,knaus1985apache, gage2001validation}. 

In this paper, we present an off-the-shelf tool to create scoring systems that are both accurate \textit{and} interpretable, which we refer to as a Supersparse Linear Integer Model (SLIM). SLIM is designed to produce classification models are appealing to domain experts, and suitable for hands-on prediction. Given that interpretability is an inherently multifaceted notion, SLIM is built to train models that are:
\begin{itemize}[leftmargin=0.45cm,topsep=4pt,parsep=2pt]
\item \textit{Sparse}: It is well-known that humans can only handle a few cognitive entities at once ($7\pm 2$ according to \citep{miller1984selection}). Accordingly, SLIM can produce scoring systems that are \textit{sparse}. In statistics, sparsity refers to the number of terms in a model and constitutes the standard way of measuring model complexity \citep{ruping2006learning,sommer1996theory}.
\item \textit{Meaningful}: Humans are seriously limited in estimating the association between three or more variables \citep{jennings1982informal}. To help users gauge the influence of one predictive factor with respect to the others, SLIM can produce scoring systems that use integer coefficients or coefficients with a few significant digits. Many medical scoring systems \cite{antman2000timi, morrow2000timi, gage2001validation} and criminology risk assessment tools \cite{psych,webster1995hcr} have integer coefficients. US News business school ratings \cite{business} have coefficients with 1 to 3 significant digits between 0 and 1; multiplying these values by 1000 produces integer coefficients.
\item \textit{Intuitive}: R\"uping \cite{ruping2006learning} warns that domain experts tend to find a fact understandable if they are already aware of it. He illustrates this idea using the statement ``rhinoceroses can fly," - a very understandable assertion that no one would believe. Unfortunately, the signs of coefficients in many linear classification models may be at odds with intuition of domain experts due to dependent relationships between variables. As such, these models are not sufficiently intuitive to be used in practice. SLIM can avoid producing scoring systems that no one would believe by constraining the sign of their coefficients to agree with prior knowledge or intuition. That is, if we believe as Dawes \citep{dawes1979robust} does that the rate of fighting in a marriage has a negative effect on marital happiness, then we can constrain its coefficient to be negative.
\end{itemize}
SLIM is formulated as a discrete optimization problem, which minimizes the 0-1 loss to encourage a high level of accuracy, regularizes the $\ell_0$-norm to encourage a high level of sparsity, and constrains coefficients to a set of meaningful and intuitive values. In this work, we solve this optimization problem using mixed-integer programming. The resulting approach is computationally challenging but can realistically produce scoring systems for datasets with thousands of training examples and hundreds of features - larger most datasets in the medical field, where scoring systems are typically used. Our work suggests that SLIM produces predictive models that are just as accurate as the those produced by state-of-the-art methods in statistics, but are also far more practical and interpretable.

Our paper is structured as follows. In Section \ref{Sec::RelatedWork}, we discuss related work in medicine and statistics. In Section \ref{Sec::FormulationAndMethodology}, we motivate the discrete optimization problem underlying SLIM and show to solve it as a mixed-integer program. In Section \ref{Sec::TheoreticalInsights}, we provide theoretical guarantees to explain why SLIM can produce scoring systems that are interpretable without necessarily sacrificing accuracy. In Section \ref{Sec::Applications}, we present applications of SLIM scoring systems in medicine and criminology to highlight their interpretability and practicality. In Section \ref{Sec::NumericalExperiments}, we report experimental results to show that SLIM scoring systems are accurate and sparse compared to state-of-the-art classification models. In the Appendix, we include: a SLIM formulation for imbalanced classification problems (Appendix \ref{Appendix::SLIMImbalancedModel}); tables of our experimental results in Section \ref{Sec::NumericalExperiments} (Appendix \ref{Appendix::TableOfExperimentalResults}); an empirical study on the effect of $\ell_0$- and $\ell_1$-regularization in SLIM (Appendix \ref{Appendix::C0C1Contours}); and an empirical of the computational performance of SLIM (Appendix \ref{Appendix::ComputationalPerformance}).
\section{Related Work}\label{Sec::RelatedWork}
SLIM produces scoring systems that strike a delicate balance between accuracy and interpretability. In the past, this task has been approached differently in medicine and statistics. On one hand, the medical community has produced highly interpretable scoring systems using heuristic techniques that are not optimized for predictive accuracy. On the other hand, the statistics community has developed black-box classification models that are optimized for predictive accuracy but have mostly ignored interpretability (with a few exceptions, see \cite{VellidoEtAl12}). In what follows, we review the related work in medicine and statistics separately.
\subsection{Related Work in Medicine}
Some popular medical scoring systems include:
\begin{itemize}[leftmargin=0.45cm,topsep=4pt,parsep=2pt]
\item SAPS I, II and III, to assess the mortality of patients in intensive care \cite{le1984simplified,le1993new,metnitz2005saps,moreno2005saps};
\item APACHE I, II and III, to assess the mortality of patients in intensive care \cite{knaus1981apache,knaus1985apache,knaus1991apache};
\item CHADS$_2$, to assess the risk of stroke in patients with atrial fibrillation \cite{gage2001validation};
\item TIMI, to assess the risk of death and ischemic events in patients with certain types of heart problems \cite{antman2000timi,morrow2000timi};
\item SIRS, to detect Systemic Inflammatory Response Syndrome \cite{bone1992american};
\item Wells Criteria for pulmonary embolisms
\cite{wells2000derivation}, and deep vein thrombosis \cite{wells1997value};
\item Ranson Criteria for acute pancreatitis \cite{ranson1974prognostic};
\item Light's Criteria for transudative from exudative pleural effusions \citep{light1972pleural}.
\end{itemize}

All of the scoring systems listed above are highly interpretable models, in that they are sparse models with meaningful and intuitive coefficients. The CHADS$_2$ scoring system, for instance, uses 5 features whose coefficients take on values of 1 and 2 to represent 5 well-known risk factors for strokes.
Unfortunately, many of these models were constructed without optimizing for predictive accuracy. 

In some cases, medical practitioners have built scoring systems using existing classification methods that were tweaked for interpretability. The SAPS II score, for instance, was constructed by rounding logistic regression coefficients. Specifically, Le Gall, Lemeshow and Saulnier \cite{le1993new} write that ``the general rule was to multiply the $\beta$ for each range by 10 and round off to the nearest integer." This approach is at odds with the fact that rounding coefficients is known to produce suboptimal solutions in the field of integer programming.

In other cases, medical scoring systems were constructed using consensus opinion from a panel of physicians, and not learned from data at all. In Knaus et al. \cite{knaus1985apache}, for instance, it is revealed that a pool of experts used their prior beliefs to determine the features and coefficients of the APACHE II scoring system: 
``[There] was general agreement by the group on where cutoff points should be placed."
This also appears to have been the case for the CHADS$_2$ scoring system as suggested by Gage et al. \cite{gage2001validation}:
``To create CHADS$_2$, we assigned 2 points to a history of prior cerebral ischemia and 1 point for the presence of other risk factors because a history of prior cerebral ischemia increases the relative risk (RR) of subsequent stroke commensurate to 2 other risk factors combined. We calculated CHADS$_2$, by adding 1 point each for each of the following - recent CHF, hypertension, age 75 years or older, and DM - and 2 points for a history of stroke or TIA." 

In order to illustrate the dangers of constructing predictive models by hand, we note that an attempted improvement to CHADS$_2$, known as CHA$_2$DS$_2$-VASc \citep{Lip10}, performs \textit{worse} than CHADS$_2$. This is not to say that CHADS$_2$ cannot be improved: recent work has shown that an approach that explicitly optimizes for accuracy and interpretability can produce a predictive model that is as interpretable as CHADS$_2$, but far more accurate \citep{LethamRuMcMa13}.

The medical community is not alone in asserting that domain expertise can be used to construct accurate and interpretable predictive models. Consider the classic work of Robyn Dawes entitled ``The robust beauty of improper linear models in decision making." \citep{dawes1979robust}. Dawes points out that scoring systems whose coefficients are determined using a heuristic, ``improper" method may outperform models whose coefficients were ``obtained upon cross-validating... upon half the sample." Dawes provides several examples of well-performing ``improper" classifiers whose weights are chosen intuitively as $-1$, 0, or $+1$. Given that many methods do not always optimize the correct objective on the training data  (i.e. the classification accuracy), that they are not optimized directly for sparsity (i.e. the number of non-zero terms), and that they do not contain information about the correctness of the sign of the coefficients, it is entirely possible for Dawes to be correct.
\subsection{Related Work in Statistics}
Many off-the-shelf classifiers that have been developed in recent years have sought to achieve a balance between scalability and accuracy without accounting for interpretability (e.g. neural networks \cite{turing2004intelligent}, support vector machines \cite{vapnik1998statistical}, random forests \cite{breimanRF}, and AdaBoost \cite{freund1997decision}). Part of the reason for this is because interpretability is an inherently multifaceted notion that is difficult to measure. In the context of applied predictive modeling, we define an interpretable model as a model that is easy to understand (i.e. a model that clearly relates variables and outcomes). This may be different from a practical model, which is a model that is easy to use.

Complexity measures are not interpretability measures, though there is sometimes overlap: for example, the number of nodes in a decision tree, the number of rules in a rule base, or the maximum depth of a rule are all reasonable ways to measure both the complexity and interpretability of a classification model. Other complexity measures such as the Vapnik-Chervonenkis-dimension \cite{vapnik1998statistical}, the Akaike Information Criterion \cite{akaike1998information}, and the Bayesian Information Criterion \cite{schwarz1978estimating} are useful for hypotheses about generalization, but are not good criteria for interpretability.

Of the top ten algorithms in data mining \citep{top10} only decision-tree methods such as CART \citep{quinlan1986induction,utgoff1989incremental} and C4.5 \citep{quinlan1993c4} can attain the same degree of practicality as the classifiers that we consider in this paper. Unfortunately, these methods produce decision-trees that are only optimized for accuracy. This tends to produce decision trees with a large number of nodes that are not practical enough for hands-on prediction. It is possible to prune trees until they are practical enough to be used by humans, all the while still achieving high accuracy. Even so, Bratko\cite{bratko1997machine} points out that these shorter trees are often unnatural and unintuitive, even when their measured accuracy is higher than domain experts' accuracy. These issues also affect decision lists \cite{rivest1987learning}, which are widely used for interpretable classification because their structure mimics the way in which humans make decisions. In this case, however, some recent work has aimed to design decision lists that are more interpretable, and fully optimized for accuracy \cite{LethamRuMcMa13}. 

A popular approach for making a model more interpretable has consisted of making sparse. Optimizing for sparsity is a well-studied problem in the literature. We note, however, that sparsity only constitutes a single aspect of interpretability. 

Current linear methods such as Lasso \citep{tibshirani1996regression}, elastic net \citep{zou2005regularization} and LARS \citep{efron2004least,hesterberg2008least} use $\ell_1$-regularization (the sum of absolute values of the coefficients) as a convex proxy for $\ell_0$-regularization (the number of coefficients) for computational reasons. The $\ell_1$-regularization is only guaranteed to produce the correct sparse solution (the one which minimizes the $\ell_0$-norm) under very restrictive conditions that are rarely satisfied in practice (see \cite{zhao2007model,liu2009estimation}). It is possible to adjust the regularization parameter throughout its full range to obtain a regularization path \citep{friedman2010regularization,hastie2005entire} that yields coefficients at all levels of sparsity. However, this is not the same as using the $\ell_0$-norm directly as the $\ell_1$-norm produces a substantial amount of additional regularization on the coefficients at each level of sparsity along the path.
 
Sparsity can also be induced using feature selection algorithms \cite{guyon2003introduction,kohavi1997wrappers,mao2004orthogonal,mao2002fast,tipping2001sparse,xu2001comparison}. Some feature selection algorithms rely on analysis of relevance and redundancy \cite{yu2004efficient}, which could yield a more interpretable feature set. Nevertheless, most feature selection relies on greedy optimization and cannot guarantee an optimal balance between accuracy and sparsity (with some exceptions, see e.g. \cite{bradley1999mathematical}). Even if feature selection algorithms could provide such a guarantee, a combination of feature selection and regularized classification would not naturally produce scoring systems with meaningful or intuitive coefficients. In practice, this would require rounding or post-processing the coefficients, which can lead to suboptimal results. Other methods to produce sparse linear models with real coefficients include those of Tipping \cite{tipping2001sparse}, Bi et al. \cite{bi2003dimensionality}, Neylon \cite{neylon2006sparse}, Giacobello et al. \cite{giacobello2012sparse}, Bazerque and Giannakis \cite{mateos2010distributed}, and Balakrishnan and Madigan \cite{balakrishnan2008algorithms}.

There has been work that aims to directly optimize for sparsity using $\ell_0$-regularization. Goldberg and Eckstein \cite{GoldbergEc2012} present a mixed-integer optimization formulation similar to the one we consider in this paper, but they do not advocate solving it. Instead they advocate relaxing this formulation and including additional constraints so as to reduce the integrality gap from exponential to linear. In practical applications, this gap can be unreasonable, and even if the full problem were solved, the coefficients could still be uninterpretable. There are similar asymptotic results in other works \cite{greenshtein2006best} which are theoretically interesting but not necessarily relevant to the kind of applied problems we consider in this paper.

There is classic work supporting the idea that simple, interpretable models have the capability to perform well \citep{Holte93}. However, the most recent work on interpretable classification is mainly motivated by the fact that interpretability and transparency are crucial for domain experts to accept and use a prediction model.  Carrizosa, Mart\ ́ın-Barrag\ ́an and Morales \cite{carrizosa2010binarized,carrizosa2011detecting} and Carrizosa, Nogales-G\ ́omez and Romero Morales \cite{carrizosaDILSVM13} suggest elegant ways to improve the interpretability of SVM classifiers by limiting coefficients to a very small set of meaningful values. The review paper of Carrizosa and Romero Morales \cite{carrizosa2013supervised} mentions a way to extract easy-to-understand ``if, then" rules from SVMs. Interpretability is also addressed in a novel way by Bien and Tibshirani \cite{bien2011prototype}, who present a mechanism to extract a small set of ``representative" samples that can help domain experts understand the workings of any classification model.
\section{Methodology}
\label{Sec::FormulationAndMethodology}
\subsection{Motivation}\label{Sec:SLIMMotivation}
Our strategy for producing a classifier that balances accuracy and interpretability is to formulate an optimization problem with the following structure:
\begin{align}
\label{Eq::InterpretableOptimizationProblem}
\begin{split}
\max_f & ~~~ \textrm{Accuracy}(f) + C \cdot \textrm{InterpretabilityScore}(f)\\
\textrm{s.t} & ~~~ \textrm{InterpretabilityConstraints}(f)>0
\end{split}
\end{align}
The optimization problem in \eqref{Eq::InterpretableOptimizationProblem} can produce an interpretable classifier using two mechanisms: first, an interpretability score, which promotes interpretable classifiers through regularization; second, a set of interpretability constraints, which restrict classifiers to a user-defined interpretable set. We will use the interpretability score to induce qualities that are desirable but not strictly necessary, and interpretability constraints to enforce qualities that are strictly necessary.

We consider linear classifiers of the form $\hat{y}=\mathrm{sign}(\mathbf{x}^T\lambdab)$ because they mimic scoring systems in their ability to make predictions through addition, subtraction and multiplication. Here, $\mathbf{x}\in\mathcal{X}\subseteq \mathbb{R}^P$ denotes a vector of $P$ features (which includes an intercept term), $\hat{y}\in\mathcal{Y} = \{-1,1\}$ denotes a predicted label, and $\lambdab\in \mathbb{R}^P$ denotes a vector of coefficients. Given a dataset with $N$ training examples, $\{(\xb_i,y_i)\}_{i=1}^N$, we produce a scoring system that balances accuracy and interpretability by writing the optimization problem in \eqref{Eq::InterpretableOptimizationProblem} as:
\begin{align}
\label{Eq::SLIMGeneralFormulation}
\begin{split}
\min_{\lambdab} & ~~~ \textrm{Loss}\big(\lambdab;\{(\xb_i,y_i)\}_{i=1}^N\big) + C \cdot \textrm{InterpretabilityPenalty}(\lambdab) \\ 
\textrm{s.t} & ~~~ \lambdab \in \Lset
\end{split}
\end{align}
A Supersparse Linear Integer Model (SLIM) is a special case 
\footnote{The framework that we present in \eqref{Eq::SLIMGeneralFormulation} can be used to construct a wide variety of interpretable models. We present version where users can define their own interpretability score and interpretability constraints in Appendix \ref{Appendix::PLIM}}
of \eqref{Eq::SLIMGeneralFormulation}, expressed as:
\begin{align}
\label{Eq::SLIMInitialFormulation}
\begin{split}
\min_{\lambdab} & ~~~ \frac{1}{N} \sum_{i=1}^{N}\indic{y_i \mathbf{x}_i^T \lambdab \leq 0} + C\vnorm{\lambdab}_0 + \epsilon \vnorm{\lambdab}_1 \\ 
\textrm{s.t.} & ~~~ \lambdab \in \Lset
\end{split}
\end{align}
Here, the objective induces accuracy using the 0-1 loss function, and regularizes for interpretability using the penalty function, $\vnorm{\lambdab}_0 + \epsilon \vnorm{\lambdab}_1$. These choices are meant to achieve a high degree of accuracy \textit{and} interpretability.
\footnote{Optimizing the 0-1 loss and the $\lzero$-norm is computationally challenging. However, there are many applications for which the extra computation may be worthwhile. We address computational considerations in Section \ref{Sec::ComputationalConsiderations} and Appendix \ref{Appendix::ComputationalPerformance}, and show SLIM can handle real-world applications in Sections \ref{Sec::Applications} and \ref{Sec::NumericalExperiments}.}

The high degree of accuracy stems from the 0-1 loss function, which produces a classifier that is robust to outliers and provides the best learning-theoretic guarantee for a finite hypothesis space. Other loss functions, such as the hinge loss (SVM), the exponential loss (AdaBoost), and the logistic loss (logistic regression), are often used as convex surrogates for the 0-1 loss for computational reasons. 

The high degree of interpretability is primarily achieved through two means: first, a $\ell_0$-penalty, which tunes the sparsity of our models; second, a set of interpretability constraints, $\lambda \in \Lset$, which restricts coefficients to a user-defined set of meaningful and intuitive values, such as integers or sign-constrained values. 

The $\lone$-penalty in the objective of \eqref{Eq::SLIMInitialFormulation} also adds to the interpretability to our models, but this merits further discussion - especially because many methods use an $\lone$-penalty to induce sparsity.  \textbf{SLIM only uses a tiny $\lone$-penalty to discard the large number of equivalent classifiers that arise when we induce sparsity with an $\ell_0$-penalty}. To illustrate this point, consider a classifier such as $\hat{y}=\mathrm{sign}(x_1 + x_2)$. If the objective in \eqref{Eq::SLIMInitialFormulation} only minimized the 0-1 loss and an $\ell_0$-penalty, then classifiers such as $\hat{y}=\mathrm{sign}(2 x_1 + 2 x_2)$ or $\hat{y}=\mathrm{sign}(3 x_1 + 3 x_2)$ would attain the same objective value as $\hat{y}=\mathrm{sign}(x_1 + x_2)$ because they make the same predictions and have the same number of features. We therefore include a tiny $\ell_1$-penalty in the objective of \eqref{Eq::SLIMInitialFormulation} so that SLIM chooses the classifier with the smallest and most interpretable coefficients within an equivalence class of solutions
\footnote{This does not imply that the solution to \eqref{Eq::SLIMInitialFormulation} is unique. Consider a case on $\mathbb{R}^2$: $\xb_1 = (-1,1)$, where $y_1 = +1$; and $\xb_2 = (1,-1)$ where $y_2 = -1$. Here, SLIM produces two optimal classifiers when coefficients are restricted to integers: $\lambdab^*=(-1,0)$ and $\lambdab^{**}$ = (0,1).}
: $\hat{y} = \textrm{sign}(x_1+x_2)$. In models where we constrain coefficients to a set of bounded integer, this strategy restricts coefficients to a coprime set
\footnote{A vector $\lambda \in \mathbb{Z}^P$ is coprime if $\gcd(\lambda_1,\ldots,\lambda_P) = 1$. Any  feasible solution for \eqref{Eq::SLIMInitialFormulation} that is not coprime is not optimal because it can be evenly divided by a positive integer to produce a solution with a smaller $\ell_1$-norm, which achieves a lower objective value.}
, which improves generalization (see Theorem \ref{Thm::CoprimeBound}) and computational performance
\footnote{If we solve SLIM using a branch and bound algorithm, as we do in this paper, then this strategy prunes nodes pertaining to equivalent classifiers without the use of call-back functions.}

For the sake the clarity, we denote the values of the $\ell_0$-penalty and $\ell_1$-penalty as $C_0$ and $C_1$ in the remainder of this paper. This leads to the standard formulation of SLIM:
\begin{align}
\label{Eq::SLIMFormulation}
\begin{split}
\min_{\lambdab} & ~~~ \frac{1}{N} \sum_{i=1}^{N}\indic{y_i \mathbf{x}_i^T \lambdab \leq 0} + C_0\vnorm{\lambdab}_0 + C_1 \vnorm{\lambdab}_1 \\ 
\textrm{s.t.} & ~~~ \lambdab \in \Lset
\end{split}
\end{align}

\subsection{Values of $C_0$ and $C_1$}
One overlooked benefit of optimizing an objective function in which accuracy and sparsity are optimized directly, without the use of proxy functions, is that it provides a meaningful interpretation for the regularization parameters. These interpretations allow practitioners to set the values of SLIM's regularization parameters purposefully, before training the model, and without necessarily using grid search.

Seeing how SLIM minimizes a combination of the zero-one loss (i.e. the misclassification rate) and the $\ell_0$ norm (i.e. the number of non-zero terms), the value of $C_0$ represents the \textit{exact} trade-off between accuracy and the number of features. That is to say, users may think of $C_0$ as the minimum decrease in training error required to include a single feature in the final classifier. To illustrate this point, consider a case where user trains a SLIM model where $C_0 = 0.01$ and obtains a classifier with $p\leq P$ features for which $\lambda_j=0$. In this case, any feature for which $\lambda_j = 0$ would have yielded less than a $1\%$ improvement in training error. Moreover, there exists no combination of $p \leq P$ features with $\lambda_j = 0$ that would yield at least a $p \times C_0$ improvement in the training error. 

This insight $C_0$ restricts the interesting values of $C_0$ to the range,
\begin{align*}
C_0 \in \left[\frac{1}{N},1\right].
\end{align*}
Choosing $C_0 \in (0,\frac{1}{N}$ will prevent any real $\ell_0$-regularization as minimizing the objective will set $\lambda_j \neq 0$ for any feature $j$ that improves the training error by $\frac{1}{N}\%$ (i.e. any feature that helps classify at least one point correctly). In contrast, choosing $C_0 \in (1,\infty)$ will produce full $\ell_0$-regularization as minimizing the objective will set $\lambda_j \neq 0$ for any feature $j$ that improves the training error by at least $1$; given that no feature that yield more than a 100\% increase in accuracy. This will set all features to $0$ and yield $\lambdab=0$.

Clearly, this interpretation of $C_0$ assumes that $C_1$ is set to value that is small enough to restrict coefficients to coprime values without affecting the accuracy nor the sparsity of the SLIM classifier. For any given $C_0$ and $\Lset$, we can set $C_1$ to such a value by ensuring that the maximum value of the $\ell_1$-penalty in SLIM's objective (i.e. $C_1\max{\vnorm{\lambdab}_1}$) is smaller than the unit value of accuracy in the objective (i.e. $\frac{1}{N}$) as well as the unit value of sparsity in the objective (i.e. $C_0$). That is, choosing 
\begin{align}
C_1 \in \left(0, \frac{ \min{(\frac{1}{N},C_0})}{\max_{\lambdab\in\Lset}({\vnorm{\lambdab}_1})} \right]
\end{align}
will ensure that we never choose a classifier that has a smaller $\ell_1$-norm if it changes the accuracy or sparsity of the classifier.
\newpage
\subsection{MIP Formulation}\label{Sec::SLIMBalancedModel}
In this paper, we solve SLIM using the following mixed-integer program (MIP):
\begin{equationarray}{crcl>{\hspace{1cm}}l>{\hspace{1cm}}r}
\min_{\bm{z},\lambdab} &\frac{1}{N}\sum_{i=1}^{N} z_i & + & \sum_{j=1}^P I_j  \notag \\
\textrm{s.t.}          & M_i z_i                  & \geq & \gamma -\sum_{j=1}^P y_ix_{ij}\lambda_j          &\mprange{i}{1}{N} & \mpdes{0-1 loss} \label{Con::SLIMLoss} \\
& & & & & \notag \\
& \lambdab & \in & \Lset & & \mpdes{coefficient values} \notag \\ 
& & & & & \notag \\
& I_j & = & C_0\alpha_j + C_1\beta_j &\mprange{j}{1}{P}& \mpdes{int. penalty} \label{Con::SLIMIntPenalty} \\
&\Lambda_j\alpha_j    & \geq & \lambda_j   &\mprange{j}{1}{P} & \mpdes{$\ell_0$-norm \#2} \label{Con::SLIML0NormUpper} \\
&\Lambda_j\alpha_j    & \geq & -\lambda_j   &\mprange{j}{1}{P} & \mpdes{$\ell_0$-norm \#2} \label{Con::SLIML0NormLower} \\
&\beta_j              & \geq & \lambda_j   &\mprange{j}{1}{P} & \mpdes{$\ell_1$-norm \#1} \label{Con::SLIML1NormUpper} \\
&\beta_j              & \geq & -\lambda_j   &\mprange{j}{1}{P} & \mpdes{$\ell_1$-norm \#2} \label{Con::SLIML1NormLower} \\
& & & & & \notag \\
& z_i & \in & \B &  \mprange{i}{1}{N} & \mpdes{0-1 loss indicators} \notag  \\
& I_j  & \in & \R_+  & \mprange{j}{1}{P} & \mpdes{int. penalty values} \notag \\
& \alpha_j  & \in & \B  & \mprange{j}{1}{P} & \mpdes{$\ell_0$ indicators} \notag \\
& \beta_j    & \in & \R_+ & \mprange{j}{1}{P} & \mpdes{abs. value variables} \ \notag
\end{equationarray}
We use the variables $z_i = \indic{y_i\neq \hat{y}_i}$ to indicate a misclassification, $\alpha_j = \indic{\lambda_j\neq 0}$ to indicate a non-zero coefficient, and $\gamma_j = |\lambda_j|$ to represent the absolute value of a coefficient. Constraints \eqref{Con::SLIMLoss} compute the $0-1$ loss by ensuring that $\alpha_i=1$ if example $i$ is misclassified by using a Big-$M$ formulation that depends on the scalar parameters, $\gamma$ and $M_i$. By default, we set $\gamma = 0.1$ and $M_i = \gamma - \max{\sum_{j=1}^P y_i x_{ij}\lambda_j}$.
\footnote{Readers who are familiar with mixed-integer programming may notice that we have used a Big-$M$ formulation to compute the zero-one loss. This is a standard formulation for the zero-one loss that does not suffer from numerical issues associated with Big-$M$ formulations. Here, the choice of $M_i$ reflects the smallest value of $M_i$ such that the condition ``if $y_i \mathbf{x}_i^T \leq 0$ then $\alpha_i=1$" holds for all $i = 1,\ldots,N$. We note that we could entirely avoid numerical issues by formulating the zero-one loss using logical constraints. Here, we present a Big-$M$ formulation because it is tightly constrained, can be solved using a wide range of MIP solvers, and can be solved faster using default settings in CPLEX 12.4}
Constraints \eqref{Con::SLIML0NormUpper} through and \eqref{Con::SLIML1NormLower} compute the $\ell_0$-norm and $\ell_1$-norm of $\lambdab$, respectively. By default, we restrict all coefficients $\lambdab$ to the set:
\begin{align*}
\Lset= \left\{ \lambdab \in \mathbb{Z}^P \big| |\lambda_j| \leq \Lambda \; \text{ for } j = 1 \ldots P\right\},
\end{align*}
where $\Lambda$ represents the largest value that SLIM may assign to any coefficient; we typically set $\Lambda$ = 100 to restrict our coefficients to integers between -100 and 100. 

We note that this formulation is not appropriate for highly imbalanced classification problems. In such cases, training a classifier that maximizes classification accuracy is likely to produces degenerate classifiers (i.e. if the probability of heart attack is 1\%, for instance, any model that never predicts a heart attack is 99\% accurate). In Appendix \ref{Appendix::SLIMImbalancedModel}, we present a SLIM formulation to produce meaningful scoring systems for imbalanced problems by specifying misclassification costs for positive and negative labels.

SLIM can be expressed using several different MIP formulations. These formulations may exhibit different computational properties based on the software that is used to solve them. In developing SLIM, we tested several formulations and found that they only exhibited minor differences in computational performance using CPLEX 12.4. Here, we have presented a formulation that is parsimonious and uses variables that are related to the structure of our problem. The latter point is especially important as it allows practitioners to further improve the way that a commercial solver tackles SLIM (e.g. by bounding the variables using prior knowledge, or setting branching priorities based on coefficients or data points).%
\subsection{Useful $\Lset$ Sets}\label{sec::UsefulLSets}
SLIM enhances the interpretability of scoring systems by allowing users to restrict coefficients to any discrete and finite set, such as a set of integers with a few significant digits. In some cases, this restriction can be seamlessly incorporated into the MIP formulation in Section \ref{Sec::SLIMBalancedModel}. In other cases, this may require additional \textit{interpretability constraints}. In general, users can restrict the coefficient $j$ to the set $\Lset=\{ l_1, l_2, \ldots, l_{\Omega_j}\}$ by defining a new set of $\Omega_j$ variables $u_{j\omega}\in\{0,1\}$ and adding the following constraints to the MIP:
\begin{align*}\centering
&\lambda_j = \sum_{\omega=1}^{\Omega_j} l_\omega u_{j\omega} \hspace{1cm} \sum_{\omega=1}^{\Omega_j} u_{j\omega} \leq 1 .
\end{align*}
In what follows, we provide examples of $\Lset$ to reproduce coefficients from medical scoring systems. Although our examples apply to all of the coefficients in a scoring system, practitioners may mix and match our guidelines to restrict different coefficients to different interpretable sets.
\subsubsection{Basic Integers}
In the default formulation, we set $\Lambda = 100$ so that SLIM chooses coefficients from:
\begin{align*}
\Lset= \left\{ \lambdab \in \mathbb{Z}^P \big| |\lambda_j| \leq \Lambda \; \text{ for } j = 1 \ldots P\right\},
\end{align*}
This produces integer coefficients that range between $-100$ and $100$ without the use of interpretability constraints. The Wells score for DVT \citep{wells1997value} uses both positive and negative integer coefficients.
\subsubsection{Sign-Constrained Integers}
SLIM can force the sign of coefficients to be positive or negative so as to capture established relationships between the data and the outcome variable. This may be important for producing models that are intuitive. For instance, the CHADS$_2$ score \citep{gage2001validation} and the TIMI score \citep{morrow2000timi} both use positive coefficients.
Suppose that we wanted the coefficients with indices in the set $S_{pos}$ to be non-negative, the coefficients with indices in the set $S_{neg}$ to be non-positive, and the remaining coefficients in the set $S_{free}$ to take on either sign. We may then express $\Lset$ as:
\begin{align*}
\Lset = \Lset_{pos} \cup \Lset_{neg} \cup \Lset_{free}
\end{align*}
%
%
\begin{alignat*}{5}\centering
\Lset_{pos} &= \Big\{ \lambdab \in \mathbb{Z}^{|S_{pos}|} &:&  &0  \leq \lambda_j \leq \Lambda &\hspace{0.3cm} \forall j \in S_{pos} &\Big\},\\
\Lset_{neg} &= \Big\{ \lambdab \in \mathbb{Z}^{|S_{neg}|}  &:& \; -&\Lambda \leq \lambda_j \leq 0 &\hspace{0.3cm} \forall j \in S_{neg} &\Big\},\\
\Lset_{free} &= \Big\{ \lambdab \in \mathbb{Z}^{|S_{free}|} &:& \; -&\Lambda \leq \lambda_j \leq \Lambda &\hspace{0.3cm}\forall j \in S_{free} &\Big\}.
\end{alignat*}
These sets can be implemented without interpretability constraints, by using bounds for the coefficient variables $\lambda_j$. Note that a sign constrained formulations may produce a more accurate predictive model by incorporating prior knowledge, and may improve the computational performance of SLIM by restricting the feasible region of the MIP.
%
%
\subsubsection{One Significant Digit}
Sometimes, the features of a dataset have wildly different orders of magnitude. In such cases, we might want a model similar to the following: \textit{predict violent crime in neighborhood next year if sign(0.0001$\#$residents -3$\#$parks +60$\#$thefts$\_$last$\_$year)$>$0}. Forcing the leading digit to be non-zero produces a scoring system that is practical enough for hands-on prediction and that synchronizes with the units of the different features. Consider a case where the coefficients have one significant digit and range between $10^{-3}$ and 900. In such a case, we could define the set $\Lset$ as:
\begin{align*}\centering
\Lset = 
\left\{
	\begin{array}{l|c}
	\multirow{3}{*}{$\lambdab \in \mathbb{Z}^P$} & \lambda_j = d \times 10^E \, \for j = 1,\ldots, P  \\ 
	 & d \in \{0, \pm 1,\pm 2\ldots \pm 9\} \\ 
	 & E\in \{-3, -2, -1 ,0, 1, 2\}
	\end{array}
\right\}.
\end{align*}
\subsubsection{Two Significant Digits}
We may wish to consider two significant digits in our coefficients rather than one, similar to the Wells score \citep{wells2000derivation}. The following set contains coefficients that range from -9900 to 9900 where the first two digits are significant:
\begin{align*}\centering
\Lset = 
\left\{
	\begin{array}{l|c}
	\multirow{4}{*}{$\lambdab \in \mathbb{Z}^P$} &  \lambda_j = d_1 \times 10^{E_1} + d_2 \times 10^{E_2} \, \for \, j = 1,\ldots, P  \\ 
	 & d_1,d_2 \in  \{0, \pm 1,\pm2, \ldots, \pm9\} \\ 
	 & E_1, E_2 \in \{  0, 1, 2, 3 \} \\
	 & E_2=E_1-1 
	 \end{array}
\right\}.
\end{align*}
\subsection{Computational Considerations}
\label{Sec::ComputationalConsiderations}
Given that SLIM is a discrete optimization problem, computation is an important consideration. It is well-known that discrete optimization problems are NP-hard. However, this does \textit{not} mean that we should avoid  solving them. Over the last two decades, we have been able to tackle exponentially larger discrete optimization problems using mixed-integer programming (MIP) due to two reasons: first, a steady increase in computational power; second, the emergence of commercial solvers that incorporate state-of-the-art MIP research.
\footnote{In Mixed-Integer Programming: A Progress Report, for example, it is shown that CPLEX 8 yields a 12 to 528-fold improvement in solution times over CPLEX 5 for 758 MIP models; this represents an order of magnitude improvement in solution times for many problems.}

In this paper, we allocated at most one hour of computing time to train a single scoring system using CPLEX 12.4. This translated into at most 3 hours of computing time when we ran a 5-fold cross-validation on 6 distinct values of $C_0$.
\footnote{We ran 30 training instances of SLIM. We solved 12 instances at a time, in parallel, on a 12-core 2.7 GhZ Intel Nehalem processor with 48 GB RAM.}
In practice, users can expect shorter computing times as:
\begin{itemize}[leftmargin=0.45cm,topsep=4pt,parsep=2pt]
\item Our one-hour time limit was unnecessary and self-imposed. As shown in Appendix \ref{Appendix::ComputationalPerformance}, SLIM can produce accurate and interpretable scoring systems within minutes. In many cases, CPLEX had found the optimal solution to our instance early on and used the remaining time to obtain a certificate of optimality.
\item We trained SLIM using default settings in CPLEX 12.4 to ensure that our results were reproducible and generalizable. Users can easily improve the computational performance by warm-starting the MIP with rounded values of the coefficients from LARS Lasso or Logistic Regression, by using a branching strategy that produces a diverse set of feasible solutions instead of narrowing the optimality gap for a single solution, and by running a self-tuning procedure that is included in many commercial solvers.
\end{itemize}
\section{Theoretical Insights}\label{Sec::TheoreticalInsights}
%
According to the principle of structural risk minimization \citep{vapnik1998statistical}, training a classifier that belongs to a simpler class of models can lead to an improved guarantee on predictive accuracy. 

Consider training a classifier $f:\X\rightarrow\Y$ with a dataset containing $N$ i.i.d samples $\{(\xb_i,y_i)\}_{i=1}^N,$, where each sample consists of a vector of $P$ features, $\xb_i \in \X \subseteq \R^P$, and a class label, $y_i \in \mathcal{Y} = \{-1,1\}$. In what follows, we provide generalization guarantees that bound the predictive accuracy of all classifiers within a class of models, $f \in \mathcal{F}$. These guarantees bound the true risk of a classifier,
\begin{align}
R^{\text{true}}(f) &\equiv \mathbb{E}_{(\xb,y)\sim\X,\Y}\mathbbm{1}[f(\xb) \neq y],
\intertext{in terms of its empirical risk,}
R^{\text{emp}}(f) &\equiv \frac{1}{N}\sum_{i=1}^N \mathbbm{1}[f(\xb_i) \neq y_i].
\end{align}

We begin with a bound on the predictive accuracy of any linear classifier whose coefficients belong to $\Lset$,
\newtheorem{thm}{Theorem}
\begin{thm}\label{Thm::NormalBound}
Let $\mathcal{F}$ denote the set of linear classifiers with coefficients $\lambdab \in \Lset$. That is,
\begin{align*}
\mathcal{F} \equiv \left\{f :\X\to\Y \;\big|\; f(\mathbf{x})=\textnormal{sign}(\mathbf{x}^T\lambdab) \textnormal{ and } \lambdab \in \Lset \right\}.
\end{align*}
For every $\delta > 0,$ with probability at least $1-\delta$, every classifier $f\in\mathcal{F}$ obeys:
\begin{align*} 
R^{\textnormal{true}}(f) \leq R^{\textnormal{emp}}(f) + \sqrt{\frac{\log(|\Lset| + 1) - \log(\delta)}{2N}}.
\end{align*}
\end{thm}
The proof uses Hoeffding's inequality for a single function $f$, combined with the union bound over all functions $f$ such that $\lambdab \in \Lset$.

In the default case, we set $\Lset$ as a set of bounded integers. As we explain in Section \ref{Sec:SLIMMotivation}, however, SLIM uses a tiny $\ell_1$-penalty to further restrict coefficients to sets of coprime integers. This improves the generalization bound presented in Theorem \ref{Thm::NormalBound} as follows.
\begin{thm}\label{Thm::CoprimeBound}
Let $\mathcal{F}$ denote the set of linear classifiers with coprime integer coefficients, $\lambdab$, bounded by $\Lambda$. That is,
\begin{align*}
\mathcal{F} &\equiv \left\{f :\X\to\Y \;\big|\; f(\mathbf{x})=\textnormal{sign}(\mathbf{x}^T\lambdab) \textnormal{ and } \lambdab \in \Lset \right\},
\intertext{where,}
\Lset &\equiv \{\lambdab \in \mathbb{\hat{Z}}^P : |\lambda_j| \leq \Lambda \textnormal{ for } j=1,\ldots,P\},\\
\hat{\Z}^P &\equiv \left\{\bm{z}\in\mathbb{Z}^P:\textnormal{gcd}(\bm{z}) = 1\right\}.
\end{align*}
For every $\delta > 0,$ with probability at least $1-\delta$, every classifier $f\in\mathcal{F}$ obeys:
\begin{align*} 
R^{\textnormal{true}}(f) &\leq R^{\textnormal{emp}}(f) + \sqrt{\frac{\log(\mathcal{C}_{\Lambda,P}) - \log(\delta)}{2N}},
\intertext{where,}
\mathcal{C}_{P,\Lambda} &= \left\{ \frac{\lambdab}{q} \in [0,1)^P: (\lambdab,q) \in \mathbb{\hat{Z}}^{P+1} \text{ and } 0 < q \leq \Lambda \right\}, 
\intertext{represents the set of Farey points of level $\Lambda$}
\end{align*}
\end{thm}
This bound can be significantly tighter than that one in Theorem \ref{Thm::NormalBound}, especially when $P$ and $\Lambda$ are small. As before, the proof uses Hoeffding's inequality for a single function $f$, combined with the union bound over all functions $f$ such that $\lambdab \in \Lset$. The fact that the set of bounded coprime integers can be expressed in terms of Farey points suggests an interesting relationship between discrete classification and the geometry of numbers. A recent study on these sequences can be found in \cite{marklof2012fine}.

The fact that more restrictive hypothesis spaces can lead to better generalization provides some motivation for using more interpretable models without necessarily expecting a loss of accuracy. As the amount of data $N$ increases, these bounds indicate that we can refine the set $\Lset$ to include more functions. When a large amount of data are available, we would be able to reduce the empirical error by including, for instance, one more significant digit within each coefficient $\lambda_j$.
\section{Applications}\label{Sec::Applications}
In this section, we present three applications of SLIM scoring systems in medicine and criminology so as to illustrate their practical and interpretable nature.
\subsection{Detecting Breast Cancer using Data from a Biopsy}\label{Sec::DemoBreastCancer}
Our first application is a scoring system to detect malignant breast tumors using features from a biopsy, which we built using the \breastcancer\, dataset \citep{MangaWo90,Bache+Lichman:2013}.

The \breastcancer\, dataset contains $N=683$ examples and $P=10$ features. The classification task is to predict if a breast tumor is malignant (Class $ = +1$) or benign (Class = $-1$). Here, we trained SLIM using 80\% of the examples and used remaining examples to assess predictive accuracy. We set regularization penalties to $C_0=0.006$ and restricted coefficients to the set,
$\Lset = \{0,\pm 1, \pm 5, \pm 10, \pm 50, \pm 100, \pm 500 \}^{10}.$
This setup produced the following linear model:
\begin{align}\small
\label{Fig::DemoBreastCancerSLIMClassifier}
\begin{split}
\textrm{Score} = & \;\; \textrm{ClumpThickness} + \textrm{UniformityOfCellShape} + \textrm{BareNuclei} - 10 \\
\textrm{Class} = & \;\; \textrm{sign(Score)}.
\end{split}
\end{align}
which we can represent as the following scoring system in Figure \ref{Fig::DemoBreastCancerSLIM}.
\begin{figure}[htbp] 
\caption{SLIM scoring system for the \breastcancer\, dataset.}
\label{Fig::DemoBreastCancerSLIM}\centering
\begin{tabular}{| r | r c |}
\hline
\rule{0pt}{3ex} Clump Thickness  (1 to 10) & & $\cdots\cdots$  \\
Uniformity of Cell Size (1 to 10) & $+$ & $\cdots\cdots$  \\
Bare Nuclei  (1 to 10) & $+$ & $\cdots\cdots$ \\
 &  $-$ & 10  \\ 
\hline
\textbf{Total} & $=$ & $\cdots\cdots$ \\
\hline
\end{tabular}
\end{figure}

The scoring system in Figure \ref{Fig::DemoBreastCancerSLIM} is both sparse and accurate in comparison to the models produced by state-of-the-art methods in Section \ref{Sec::NumericalExperiments}, as it only uses 4 features and achieves an error of 3.7\% on the training set and 2.2\% on the test set. A medical practitioner can use this model by adding the values of three features for a given tumor, and subtracting 10 from the total. If the score is positive, then the model predicts that the tumor is malignant with 97.8\% accuracy.

For the sake of comparison, we show the model produced by LARS Lasso in Figure \ref{Fig::DemoBreastCancerLARSLasso} as it represents the sparsest
\footnote{Lasso's $\ell_1$-penalty is set by the \pkg{glmnet} package to the value that produces the sparsest model within 1 standard error of the $\ell_1$-penalty that minimizes the 5-fold CV error.}
linear model produced by the baseline methods in Section \ref{Sec::NumericalExperiments}.
This model achieves an error rate of 3.7\% on both the training and test sets, which is comparable to the accuracy of the scoring system in Figure \ref{Fig::DemoBreastCancerSLIM}. However, it also involves 7 features whose coefficients that range between -4.98 to 0.34. These qualities make the model far less interpretable (as it is harder to grasp the impact of each variable on the prediction) and far less practical (as it is hard to compute the actual prediction by hand).
%
%
\begin{align}\small
\label{Fig::DemoBreastCancerLARSLasso}
\begin{split}
\textrm{Score} = & \;\; 0.24\times\textrm{ClumpThickness}  + 0.15\times\textrm{UniformityOfCellSize}  \\
& + 0.20\times\textrm{UniformityOfCellShape}  + 0.10\times\textrm{MarginalAdhesion}   \\
& + 0.34\times\textrm{BareNuclei} + 0.13\times\textrm{NormalNucleoli}  - 4.98    \\
\textrm{Class} = & \;\; \textrm{sign(Score)}.
\end{split}
\end{align}
\subsection{Detecting Breast Cancer using Data from a Mammogram}\label{Sec::DemoMammo}
Our second application is a scoring system for detect malignant breast tumors, which we have trained using the \mammo\, dataset. This scoring system is significantly less accurate than the one we presented in Section \ref{Sec::DemoBreastCancer}, but can detect malignant tumors without requiring patients to undergo a biopsy, which is invasive and often unnecessary. In this case, our model uses mammographic attributes from the Breast Imaging-Reporting and Data System (BI-RADS), which are recorded by a radiologist after examining a mammogram.

The \mammo\, dataset contains $N = 961$ examples and $P = 12$ features that can be used to predict if a breast tumor is malignant or benign (Class $ = +1/-1$). As before, we trained SLIM using 80\% of the examples, and used remaining examples to assess predictive accuracy. When we set the regularization penalty to $C_0=0.002$ and restricted coefficients to the default set $\Lset= \{ \lambdab: \lambdab \in \mathbb{Z}^{12}, \; |\lambda_j| \leq 100 \for j = 1, \ldots, 12\}$, SLIM produced the model:
\begin{alignat}{2}\small
\begin{split}
\label{Eq::DemoMammoSLIM}
\textrm{Score} &= 1 - 2 \times \textrm{OvalShape} - 2 \times \textrm{CircumscribedMargin} \\
\textrm{Class} &= \textrm{sign(Score)}
\end{split}
\end{alignat}

\begin{figure}[htbp]
\caption{Decision tree induced by the SLIM scoring system for the \mammo\, dataset.}
\vspace{0.2cm}
\label{Fig::MammoTreeSLIM}
 \begin{center}
\tikzstyle{block} = [rectangle, draw, fill=white!20, 
    text width=7.5em, text centered, rounded corners, minimum height=3.5em]
\tikzstyle{line} = [draw, -latex']
\tikzstyle{rcloud} = [draw, ellipse,fill=red!20, node distance=1cm, text width=4em, text centered,
    minimum height=2em]
\tikzstyle{gcloud} = [draw, ellipse,fill=green!20, node distance=1cm, text width=4em, text centered,
    minimum height=2em]
    \begin{tikzpicture}[node distance = 1cm, auto]
    \node [block] (init) {\small{Is the shape oval?}};
    \node [gcloud, right of=init, below of=init, node distance=1.7cm] (end1) {benign};
    \node [block, left of=init, below of=init, node distance=1.7cm] (dec1) {\small{Is the margin circumscribed?}};
    \node [rcloud, left of=dec1, below of=dec1, node distance=1.7cm] (end2) {malignant};
    \node [gcloud, right of=dec1, below of=dec1, node distance=1.7cm] (end3) {benign};
       
    \path [line] (init) -| node [near start] {yes} (end1);
    \path [line] (init) -| node [near start, above] {no} (dec1);
     \path [line] (dec1) -| node [near start] {yes} (end3);
     \path [line] (dec1) -| node [near start, above] {no} (end2);
\end{tikzpicture}
\end{center}
\end{figure}
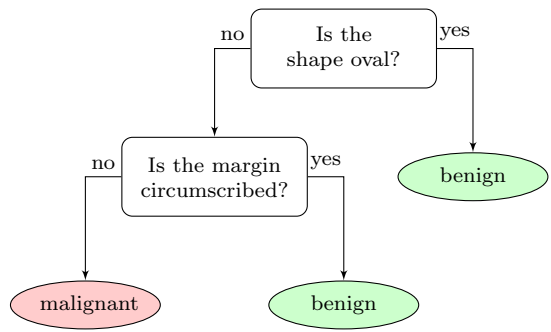

In comparison to the models produced by the baseline methods in Section \ref{Sec::NumericalExperiments}, this scoring system is both accurate and interpretable, as it achieve an error of 20.8\% on both the test and training sets using 3 meaningful features. Given that model in \eqref{Eq::DemoMammoSLIM} only uses binary features, we can also construct a decision tree by considering all possible values of the score. We show this tree in Figure \ref{Fig::MammoTreeSLIM} and note that it is simple enough to be remembered by medical practitioners.

\subsection{Predicting the Incidence of Violent Crime Among Young People}\label{Sec::DemoViolentCrime}
Our final application is a scoring system to predict whether a young person between the ages of 17 and 18 will commit a violent crime in the next three years. In light of the legal and social consequences of violent offenses, we believe that such a predictive model can be used to provide these individuals with preventative services, such as counseling.

We have built this scoring system from a US Department of Justice Statistics study of crime among young people raised in out-of-home care \citep{cusick2010crime}. The study consists of three surveys for each person taken at age 17, 19, and 21  (Wave 1, Wave 2, Wave 3). Individuals were asked detailed questions on their education, employment history, criminal history, relationships with parents, and where they were raised (foster care, group home, etc.). We constructed the \violentcrime\ dataset by deriving $P = 108$ features for $N = 558$ individuals using the responses to the survey in Wave 1 and by assigning each individual a label of $Class = +1$ if they committed a violent crime between Wave 1 and Wave 3.

Given that we observe Class $= +1$ for only 19\% of examples, we used a SLIM formulation designed for imbalanced datasets (see Appendix \ref{Appendix::SLIMImbalancedModel}). We trained SLIM using 80\% of the examples and used remaining examples to assess the predictive accuracy. We set misclassification costs to $\wplus=1.2$ and $\wminus=0.8$ and the regularization penalty to $C_0=0.01$. In addition, we restricted the coefficients to the set $\Lset = \{0,\pm 1, \pm 5, \pm 10, \pm 50, \pm 100, \pm 500 \}^{10}$. With this setup, SLIM produced the scoring system in Figure \ref{Fig::DemoViolentCrimeSLIM}.
\begin{figure}[htbp]
\caption{SLIM scoring system for the \violentcrime\, dataset.}
\label{Fig::DemoViolentCrimeSLIM}
\centering
\begin{tabular}{| l c |  c |}
 \hline
\rule{0pt}{3ex}1) Does the person have a mental health problem? &(10 points) &  $\cdots \cdots$  \\
2) Has the person ever used or threatened to use a weapon? &(5 points)&     $\cdots \cdots$  \\
3) Has the person ever shot or stabbed someone? &(5 points)   &    $\cdots \cdots$ \\
4) Has the person ever stolen something worth over \$50? &(5 points)   &    $\cdots \cdots$ \\
5) Is the person male and distanced from his dad? &(5 points)   &    $\cdots \cdots$ \\
6) Does the person not have a dad or stepdad? &(1 point)   &    $\cdots \cdots$ \\
7) Is the person male and not have a dad or stepdad? &(1 point)   &    $\cdots \cdots$ \\
8) Does the person not have a mom or stepmom? &(1 point)   &    $\cdots \cdots$ \\
9) Is the person male and not have a mom or stepmom? &(1 point)   &    $\cdots \cdots$ \\
\hline
\textbf{Sum points from 1 to 9} & {\bf Total A}&  $\cdots \cdots$  \\
\hline
10) Is the person female and not have a dad or stepdad? &(10 points) &    $\cdots \cdots$ \\
11) Does the person have college plans? &(5 points)   &    $\cdots \cdots$ \\
12) Is the person employed? &(1 point)   &    $\cdots \cdots$ \\
13) Is the person in school and employed? &(1 point)   &    $\cdots \cdots$ \\
14) Likelihood to use child welfare system. &(1-4 points)   &    $\cdots \cdots$ \\
\hline
\textbf{Sum points from 10 to 14} &{\bf Total B}&  $\cdots \cdots$  \\
\hline
\textbf{Subtract Total B from Total A} &{\bf Total C}& $\cdots\cdots $\\
\hline
\end{tabular}
\end{figure}

The scoring system is sparse, in that it uses 14 of the 108 features; practical, in that users can make a prediction by adding and subtracting a few numbers; and interpretable, in that the points are meaningful and agree with intuition of domain experts. This scoring system achieves a sensitivity of 69\% and a specificity of 44\%. Note that sensitivity is the important measure in this problem as we would much rather correctly identify young people who risk committing a violent crime, than falsely identify young people who do not risk committing a violent crime.

Assessing the predictive performance of this scoring system is not straightforward given that entire confusion matrix has to be considered to compare the prediction quality of different classifiers for imbalanced datasets.
\footnote{We could consider statistics such as the AUC. However, the AUC does not take into account the position of the decision boundary, which is problematic as our focus is on constructing classifiers. The AUC is a rank statistic, not a classification statistic.}
%
%
%
%
%
For the sake of comparison, however, we did produce a series of decision-tree classifiers for all possible misclassification costs, $\wplus$ and $\wminus$ using the \texttt{classregtree} function in MATLAB. None of these trees had fewer than 60 nodes, nor did they attain a sensitivity higher than 62\% on the test set (except for one trivial tree that had a single node and 100\% sensitivity). When we set the misclassification costs to $\wplus=1.2$ and $\wminus=0.8$, for instance, we obtained a decision tree with 93 nodes that attained a sensitivity of 62\% and specificity of 79\%. We found these models to be problematic, as they were too large to be used in practice, too complicated to be understood by domain experts, and unable to attain the same level of sensitivity.
\section{Numerical Experiments}\label{Sec::NumericalExperiments}
In this section, we show that SLIM can produce scoring systems that are accurate and sparse in comparison to state-of-the-art classification models. Our experiments compare ten different classification methods on six popular datasets from the UCI Machine Learning repository \cite{Bache+Lichman:2013}. 
\subsection{Methods}
In our experiments, we trained SLIM using default settings for the CPLEX 12.4 API in MATLAB 2012b. As we mention in Section \ref{Sec::ComputationalConsiderations}, we allocated at most one hour of computing time to train each scoring system. This means that it took us at most 3 hours to perform a 5-fold cross validation on 6 distinct values of $C_0$.
\footnote{We trained SLIM 30 times and solved 12 instances at a time, in parallel, on a 12-core 2.7 GhZ Intel Nehalem processor with 48 GB RAM.}
For the sake of comparison, we trained classification models for nine baseline methods in R 2.15, shown in Table \ref{Table::ExperimentalMethods} without any time limits. We set free parameters for most methods to the values that minimized the mean 5-fold cross-validation (CV) errors; for LARS-related methods such as Lasso, Ridge and EN, however, we set the $\ell_1$-penalty to the value that produced the sparsest model within 1 standard error of the $\ell_1$-penalty that minimized the 5-fold CV error.
\begin{table}[htbp]
\caption{Baseline Methods for the Numerical Experiments in Section \ref{Sec::NumericalExperiments}}
\centering
\renewcommand{\arraystretch}{1.25}
\begin{tabular}{| l | c | c |}
\hline
\textbf{Method} & \textbf{Acronym} & \textbf{R Package} \\ 
\hline
C5.0 Decision Trees & C50T & \pkg{c50}, \citep{kuhn2012c50} \\
\hline
C5.0 Decision Rules & C50R & \pkg{c50}, \citep{kuhn2012c50} \\
\hline
CART Decision Trees & CART & \pkg{rpart}, \citep{the2012rpart} \\
\hline
Logistic Regression & LR & N/A (built-in) \\
\hline
LARS Lasso (Binomial Family) & Lasso & \pkg{glmnet}, \citep{friedman2010glmnet, simon2011glmnet} \\
\hline
LARS Ridge (Binomial Family) & Ridge & \pkg{glmnet},\citep{friedman2010glmnet, simon2011glmnet} \\
\hline
LARS Elastic Net (Binomial Family) & EN & \pkg{glmnet},\citep{friedman2010glmnet, simon2011glmnet} \\
\hline
Random Forests & RF & \pkg{randomForest}  \\ 
\hline
Support Vector Machines (RBF Kernel) & SVM & \pkg{e1071}, \citep{meyer2012e1071} \\ 
\hline
\end{tabular}
\label{Table::ExperimentalMethods}
\end{table}

Given that interpretability is difficult to capture using a single metric, we report the interpretability of each model using a measure of sparsity, which we refer to as \textit{model size}. We define model size  in a way that reflects to the interpretability of each model. Model sizes represents the number of coefficients for linear classifiers such as SLIM, LR, Lasso, Ridge and EN, the number of leaves for decision tree classifiers such as C5.0T and CART, and the number of rules for rule-based classifiers such as C5.0R. For RF and SVM, we set the model size to the number of features in each dataset as this metric does not reflect the interpretability of these methods
\subsection{Datasets}
We trained each models on six popular datasets from the UCI Machine Learning Repository \citep{Bache+Lichman:2013}, shown in Table \ref{Table::ExperimentalDatasets}. We chose these datasets to allow a comparison with other works, and to investigate how SLIM behaves as we change the size and nature of the training data. Some datasets also varied in other interesting ways: \internetad, for instance, has a highly sparse feature matrix, while \tictactoe, is a non-linear classification problem. We processed each dataset as follows: we added a feature composed of 1's to act as an intercept; we transformed categorical features into binary features; and we either dropped examples with missing entries (\breastcancer) or imputed these values (\mammo).
\begin{table}[htbp]
\renewcommand{\arraystretch}{2}
\caption{Datasets for the Numerical Experiments in Section \ref{Sec::NumericalExperiments}}
\centering
\scalebox{1}{
\begin{tabular}{| l | c | c | c |}
\hline
\textbf{Dataset} & $N$ & $P$ & \textbf{Classification Task}\\
\hline
\breastcancer & 683 & 10 & \parbox{0.55 \linewidth}{detect breast cancer using a biopsy \vspace{2pt}} \\\hline
\haberman & 306 & 4 &\parbox{0.55 \linewidth}{predict the 5-year survival of patients who have undergone surgery for breast cancer\vspace{2pt}} \\\hline
\internetad & 2359 & 1431 & \parbox{0.55 \linewidth}{predict if an image on the internet is an ad  \vspace{2pt}} \\\hline
\mammo & 961 & 12 & \parbox{0.55 \linewidth}{detect breast cancer using a mammogram \vspace{2pt}} \\\hline
\spambase & 4601 & 58 & \parbox{0.55 \linewidth}{predict if an e-mail is spam or not} \\\hline
\tictactoe & 958 & 28 & \parbox{0.55 \linewidth}{detect if the first player has won at the end of a game of tic-tac-toe\vspace{2pt}} \\\hline
\end{tabular}
\label{Table::ExperimentalDatasets}
}
\end{table}
\subsection{Sparsity and Accuracy of SLIM vs. Baseline Methods}\label{Sec::AccuracyVsSparsity}
\subsubsection{Table of Results}
We compare the accuracy and sparsity of each method on each dataset in Tables \ref{Table::ComparisonAccuracy} and \ref{Table::ComparisonSparsity} in Appendix \ref{Appendix::TableOfExperimentalResults}. In Table \ref{Table::ComparisonAccuracy}, we report the mean and standard deviation of the 5-fold test error and training error (as measures of accuracy). We also report the median, minimum and maximum model size over the 5 folds for each method (as measures of sparsity). The results in Table \ref{Table::ComparisonAccuracy} reflect the performance of each method when we have set free parameters so as to minimize the mean 5-fold cross-validation (CV) error. Although this is the standard way to evaluate algorithms, we wanted to provide the regularized linear methods (Lasso, Ridge, EN and SLIM) with an opportunity to produce sparser models, so we also constructed Table \ref{Table::ComparisonSparsity}. In Table \ref{Table::ComparisonSparsity}, the last 4 columns report results from the sparsest model that was within one standard deviation of the accuracy of the model produced in Table \ref{Table::ComparisonAccuracy} (the remaining columns were reproduced from Table \ref{Table::ComparisonAccuracy} to allow for an easier comparison between methods).
\subsubsection{Graphical Results}
\begin{figure}[htbp]
\caption{Sparsity and accuracy of SLIM vs. baseline algorithms}
\label{Fig::AccuracyVsSparsityPlots}
\centering
\begin{subfigure}[b]{0.475\textwidth}
    \centering
    \includegraphics[width=\textwidth]{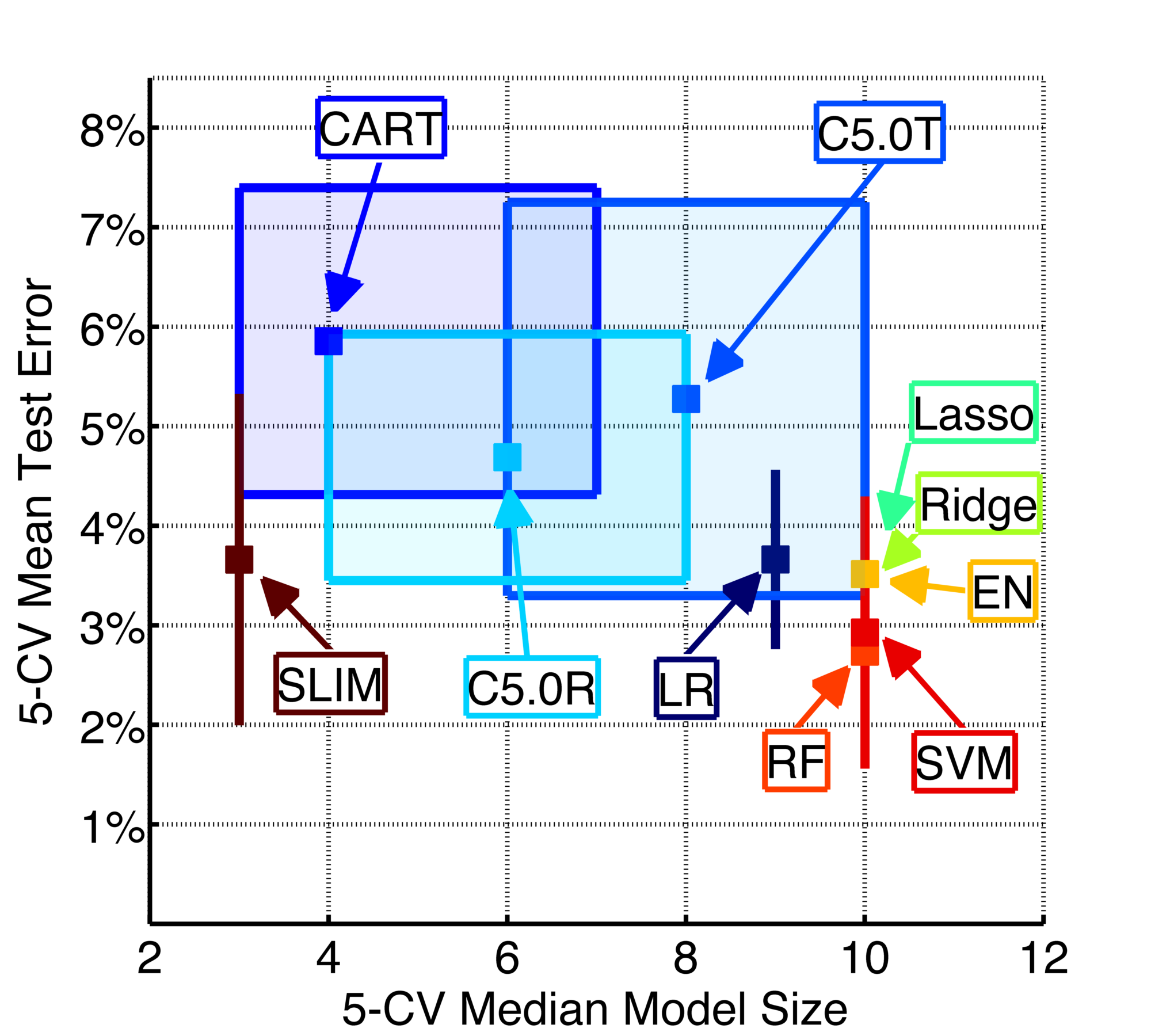}
    \caption{\breastcancer}
    \label{Fig::AccuracyVsSparsity_breastcancer}
\end{subfigure}
\hspace{0.1cm}
\begin{subfigure}[b]{0.475\textwidth}
    \centering
    \includegraphics[width=\textwidth]{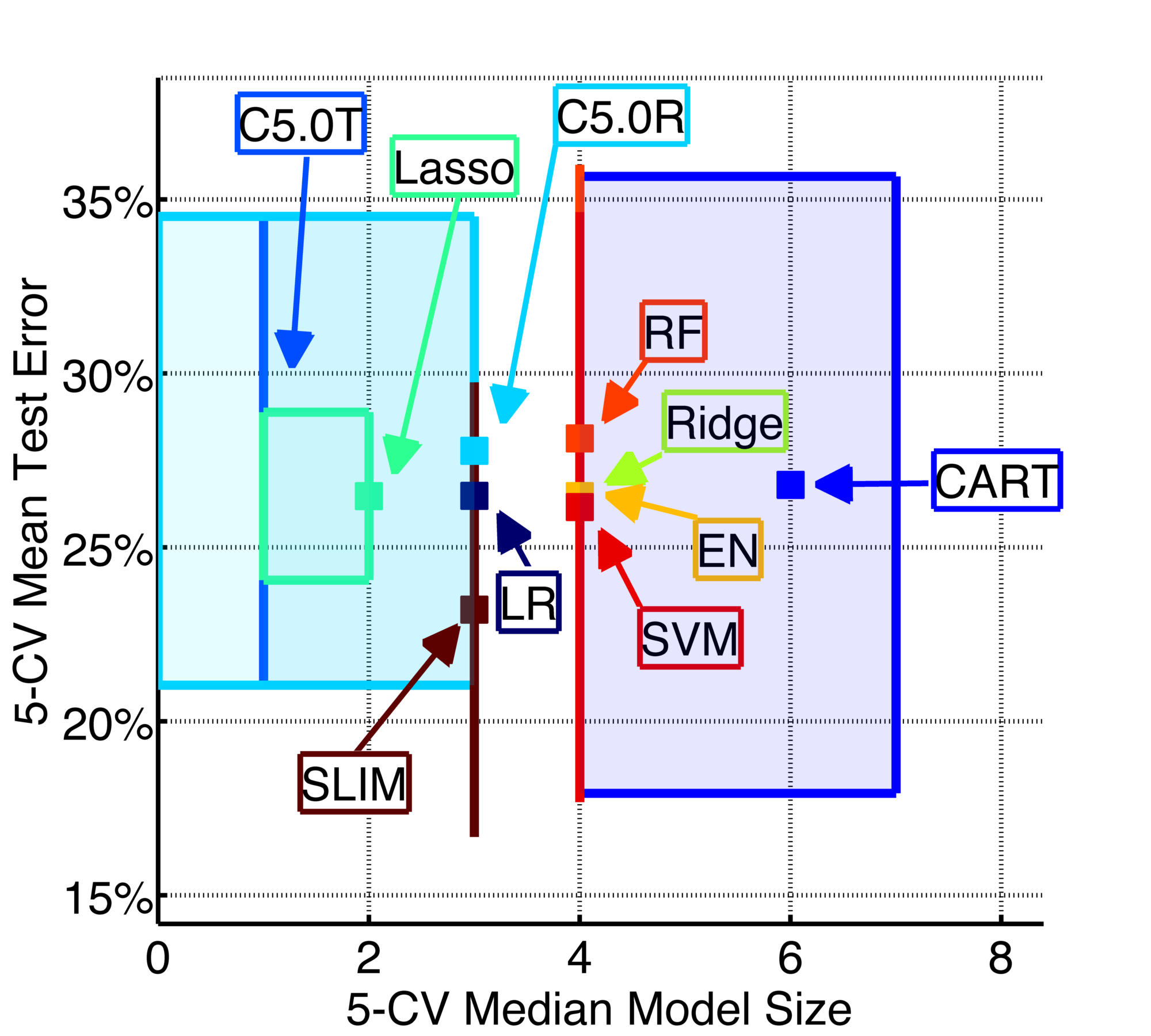}
    \caption{\haberman}
    \label{Fig::AccuracyVsSparsity_haberman}
\end{subfigure}
\begin{subfigure}[b]{0.475\textwidth}
    \centering
    \includegraphics[width=\textwidth]{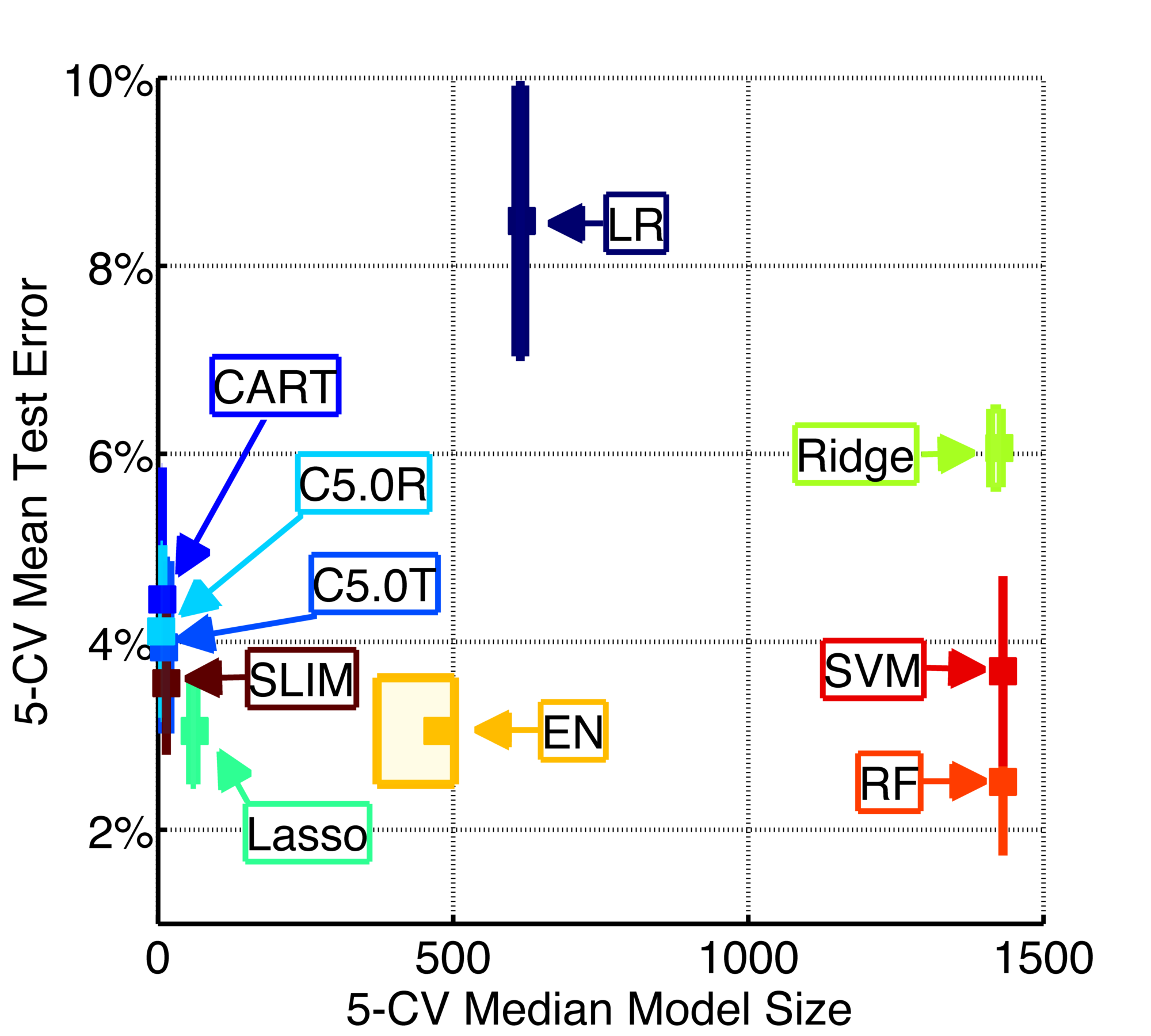}
    \caption{\internetad}
    \label{Fig::AccuracyVsSparsity_internetad}
\end{subfigure}
\hspace{0.1cm}
\begin{subfigure}[b]{0.475\textwidth}
    \centering
    \includegraphics[width=\textwidth]{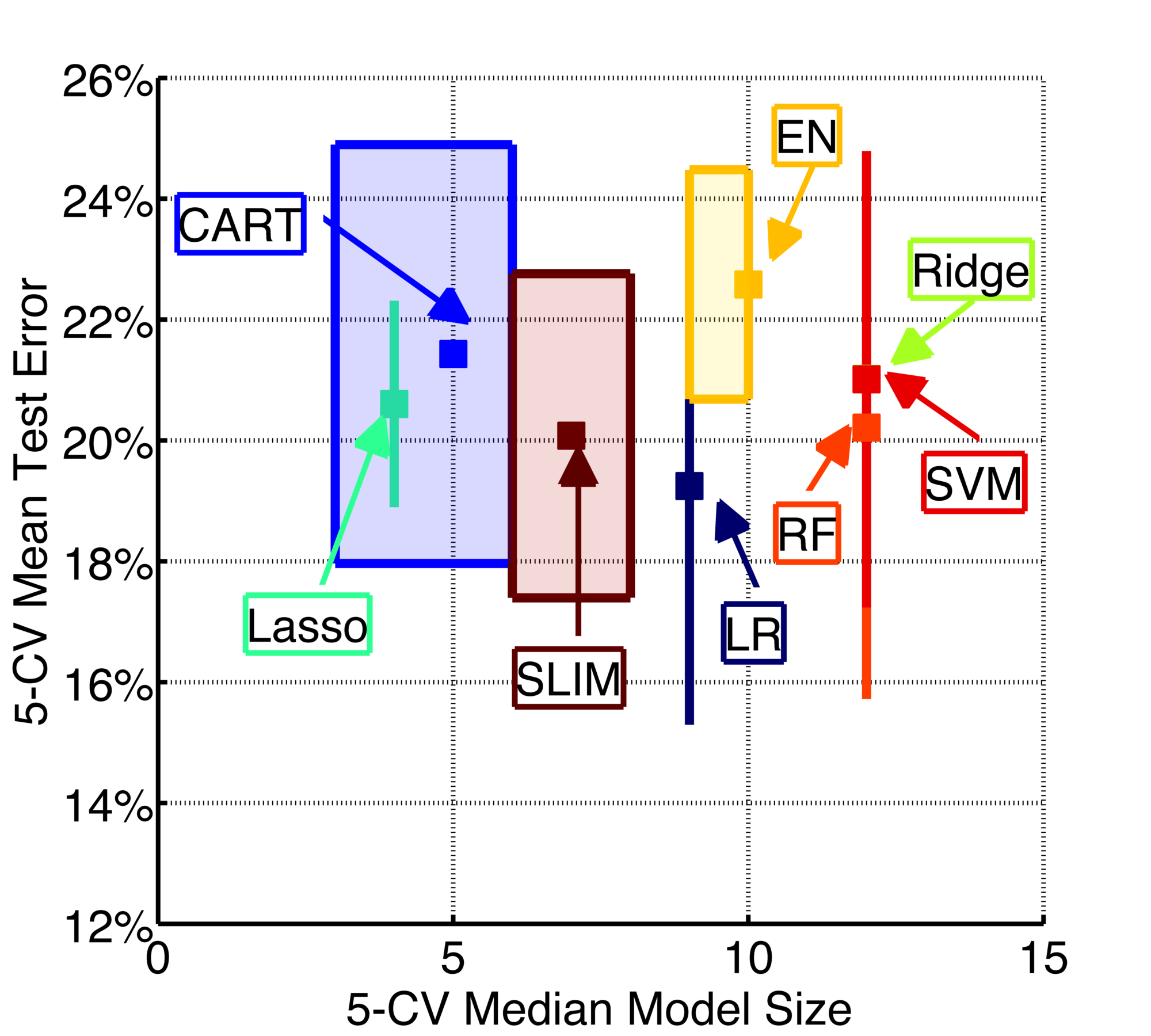}
    \caption{\mammo}
    \label{Fig::AccuracyVsSparsity_mammo}
\end{subfigure}
\begin{subfigure}[b]{0.475\textwidth}
    \centering
    \includegraphics[width=\textwidth]{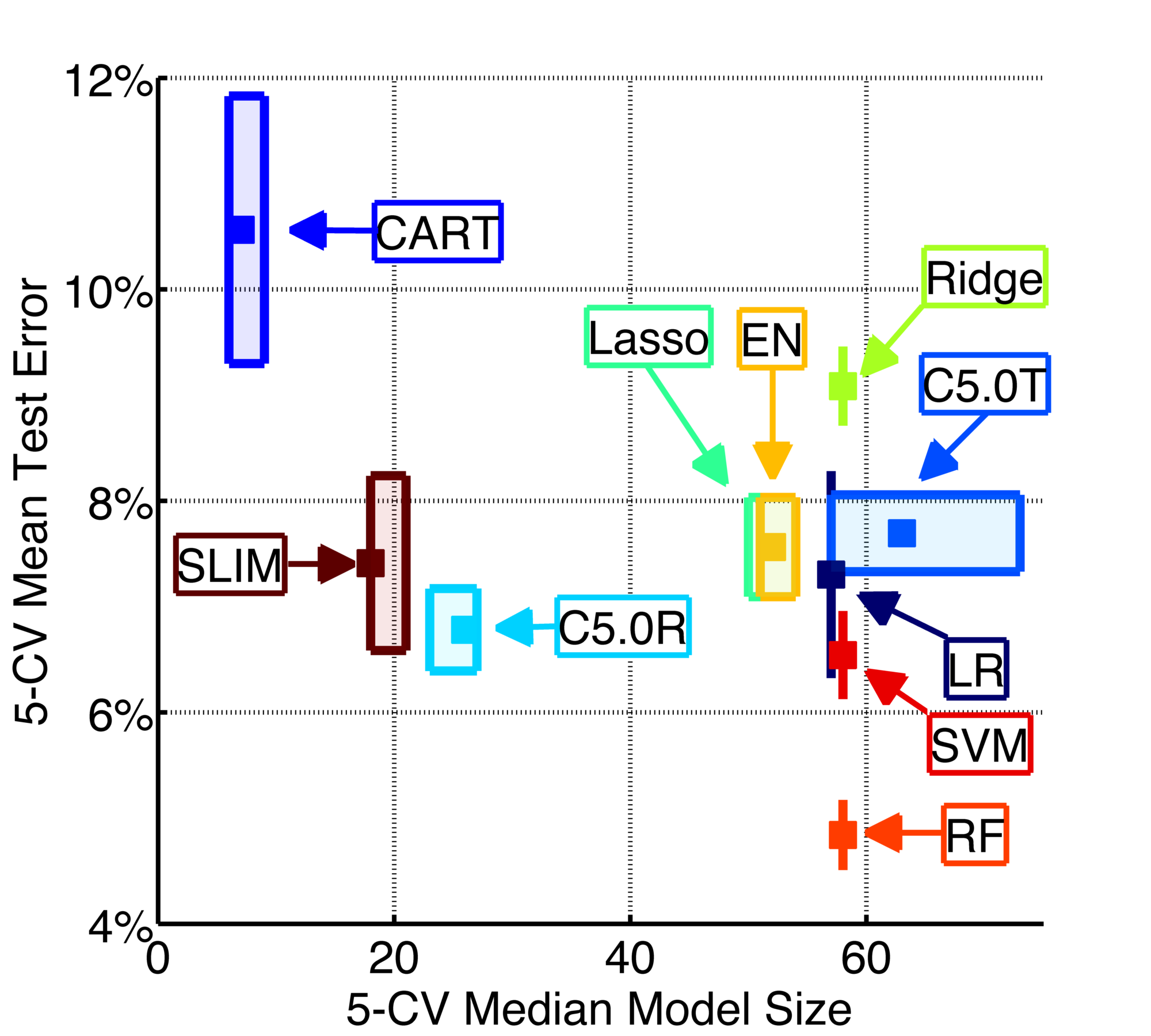}
    \caption{\spambase}
    \label{Fig::AccuracyVsSparsity_spambase}
\end{subfigure}
\hspace{0.1cm}
\begin{subfigure}[b]{0.475\textwidth}
    \centering
    \includegraphics[width=\textwidth]{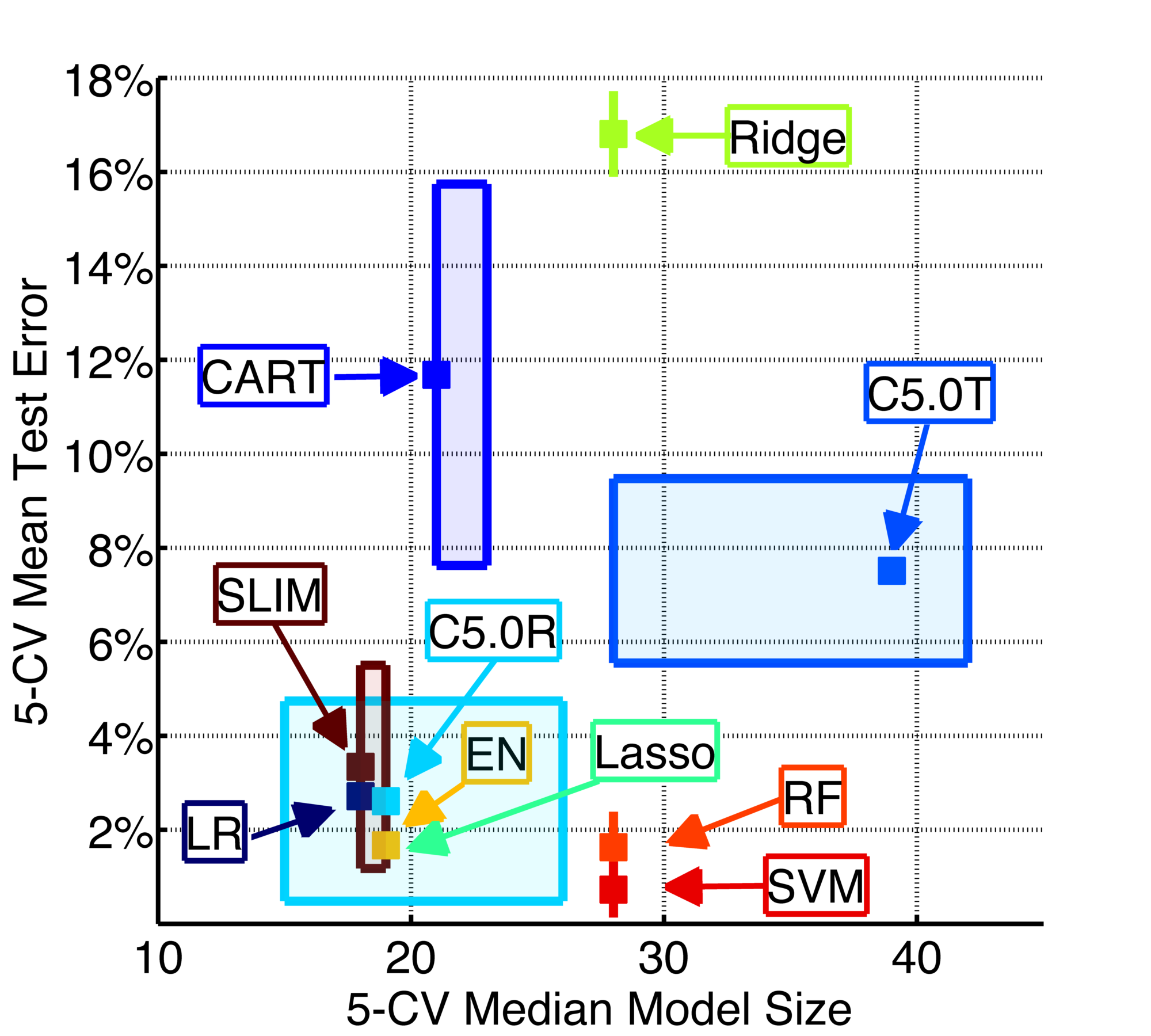}
    \caption{\tictactoe}
    \label{Fig::AccuracyVsSparsity_tictactoe}
\end{subfigure}
\end{figure}

We provide a visual representation of our results in Table  \ref{Table::ComparisonAccuracy} in Figures \ref{Fig::AccuracyVsSparsity_breastcancer}-\ref{Fig::AccuracyVsSparsity_tictactoe}. Each figure plots the accuracy and sparsity of multiple algorithms on a single dataset. In a given figure, we plot a single point for each algorithm corresponding to the mean 5-fold CV test error (as a measure of accuracy) and the median 5-fold CV model size (as a measure of sparsity). We surround this point with a box to highlight the variation in accuracy and sparsity for each algorithm; in this case, the box ranges over the 5-fold CV standard deviation in test error and the 5-fold min/max of model sizes. In situations where an algorithm shows no variation in model size over the 5 folds, we have plotted the algorithm as a vertical line rather than a box (i.e. no horizontal variation). In cases when algorithms produce models with the same size (e.g. Lasso, Ridge and EN on the \breastcancer\, dataset) the boxes or lines will also coincide. When an algorithm produces a model that is not dominated by another algorithm (i.e. no other algorithm can produce a more that is more accurate \textit{and} more sparse), we say it lies on the \textit{efficient frontier}. The methods that consistently lie on the efficient frontier for all datasets achieve the best possible balance between accuracy and sparsity.
\subsubsection{Discussion}
Our main observations on the experimental results are that: (i) SLIM scoring systems often lie on the efficient frontier, meaning other methods were often unable to produce a model that was both more accurate and more sparse; (ii) SLIM scoring systems are generally more sparse than the models produced by other methods; (iii) SLIM's model sizes are more stable (have less variation) than that of other methods; (iv) in comparison, some of the baseline methods have very high variance in model size (e.g., CART, C5.0R and C5.0T); and (v) there is no single method that performs better than all others on all datasets, although many methods produce models that lie on the efficient frontier. Observations (i)-(iv) can be accounted for because SLIM directly optimizes the accuracy and sparsity of its classifiers, without the use of convex loss functions or regularized approximations for sparsity. This allows SLIM to produce interpretable models whose predictive performance does not suffer.
\subsection{SLIM vs. LARS Lasso}
\begin{figure}[htbp]
\caption{Sparsity and accuracy of SLIM vs. models on the regularization path of LARS Lasso}
\label{Fig::SLIMvsLasso}
\centering
\begin{subfigure}[b]{0.475\textwidth}
    \centering
    \includegraphics[width=\textwidth]{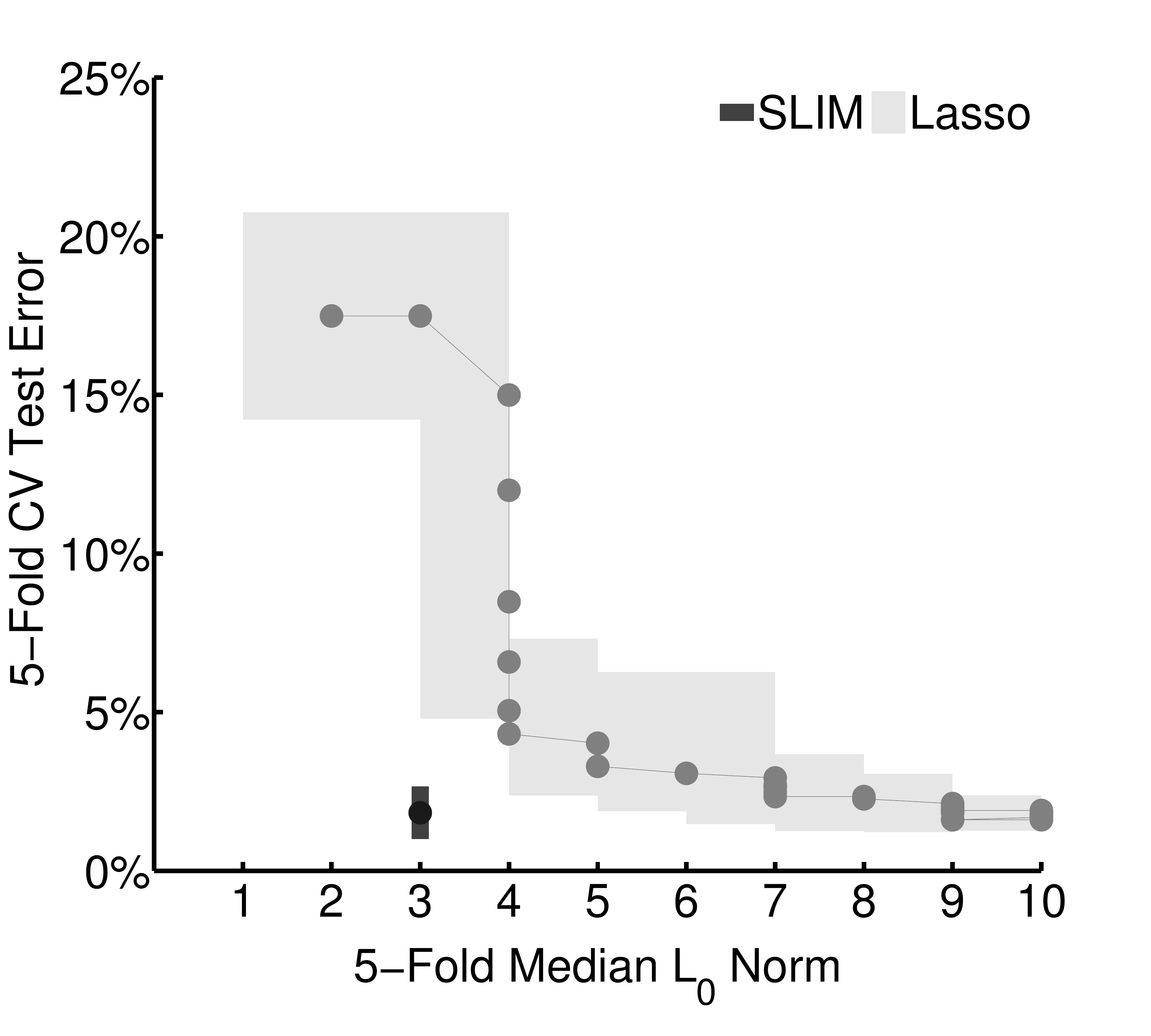}
    \caption{\breastcancer}
    \label{Fig::SLIMvsLasso_breastcancer}
\end{subfigure}
\hspace{0.1cm}
\begin{subfigure}[b]{0.475\textwidth}
    \centering
    \includegraphics[width=\textwidth]{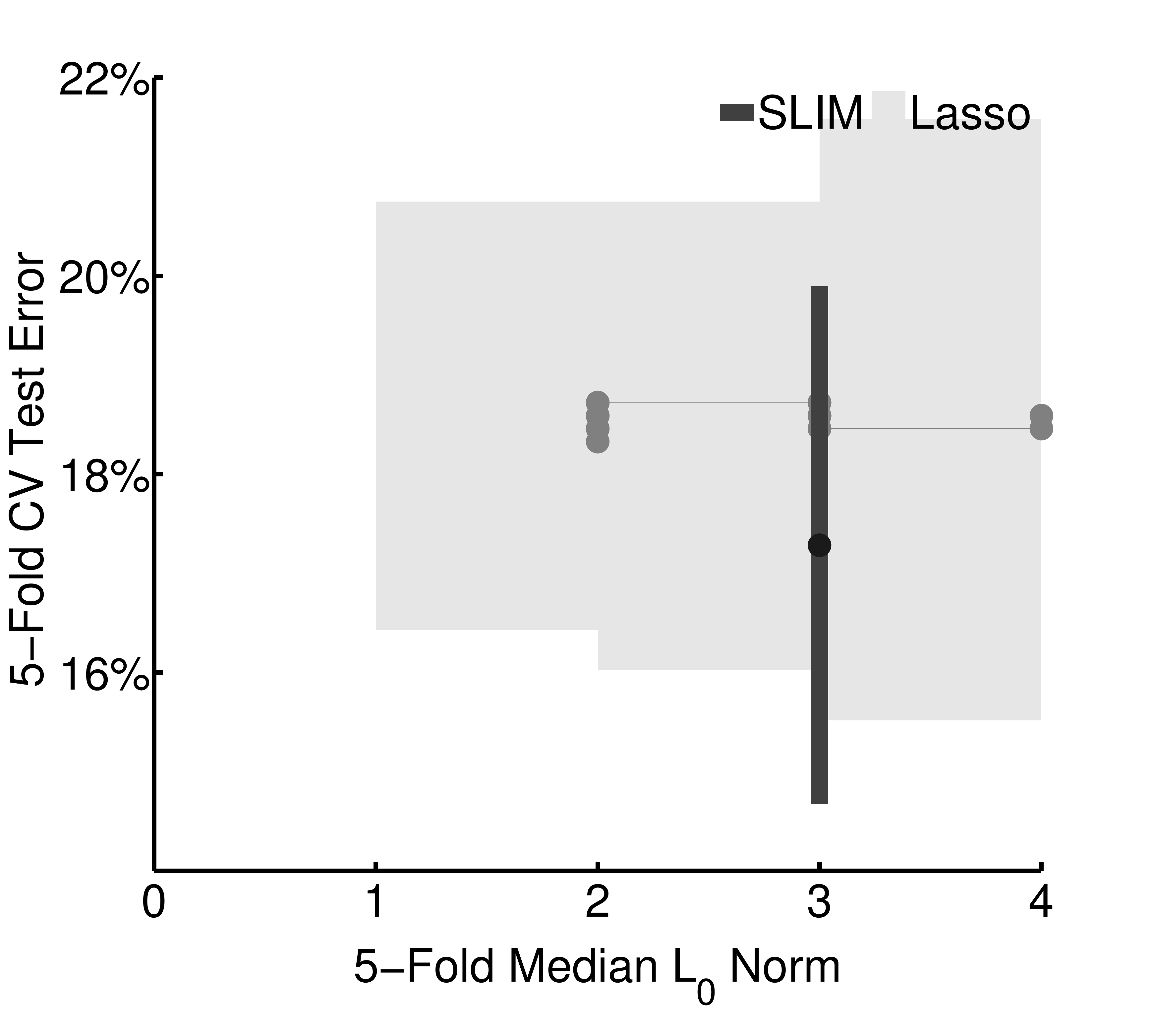}
    \caption{\haberman}
    \label{Fig::SLIMvsLasso_haberman}
\end{subfigure}
\begin{subfigure}[b]{0.475\textwidth}
    \centering
    \includegraphics[width=\textwidth]{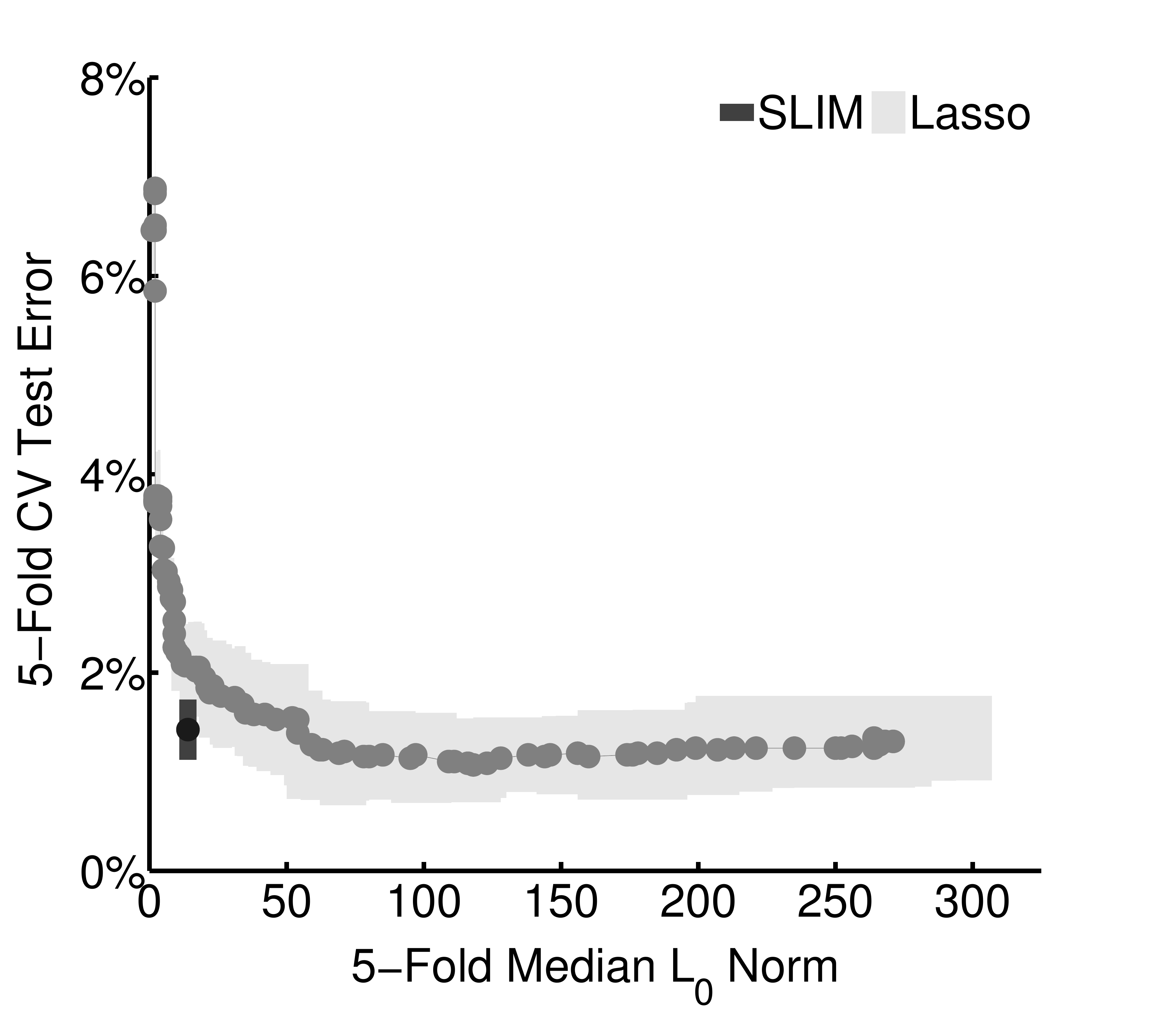}
    \caption{\internetad}
    \label{Fig::SLIMvsLasso_internetad}
\end{subfigure}
\hspace{0.1cm}
\begin{subfigure}[b]{0.475\textwidth}
    \centering
    \includegraphics[width=\textwidth]{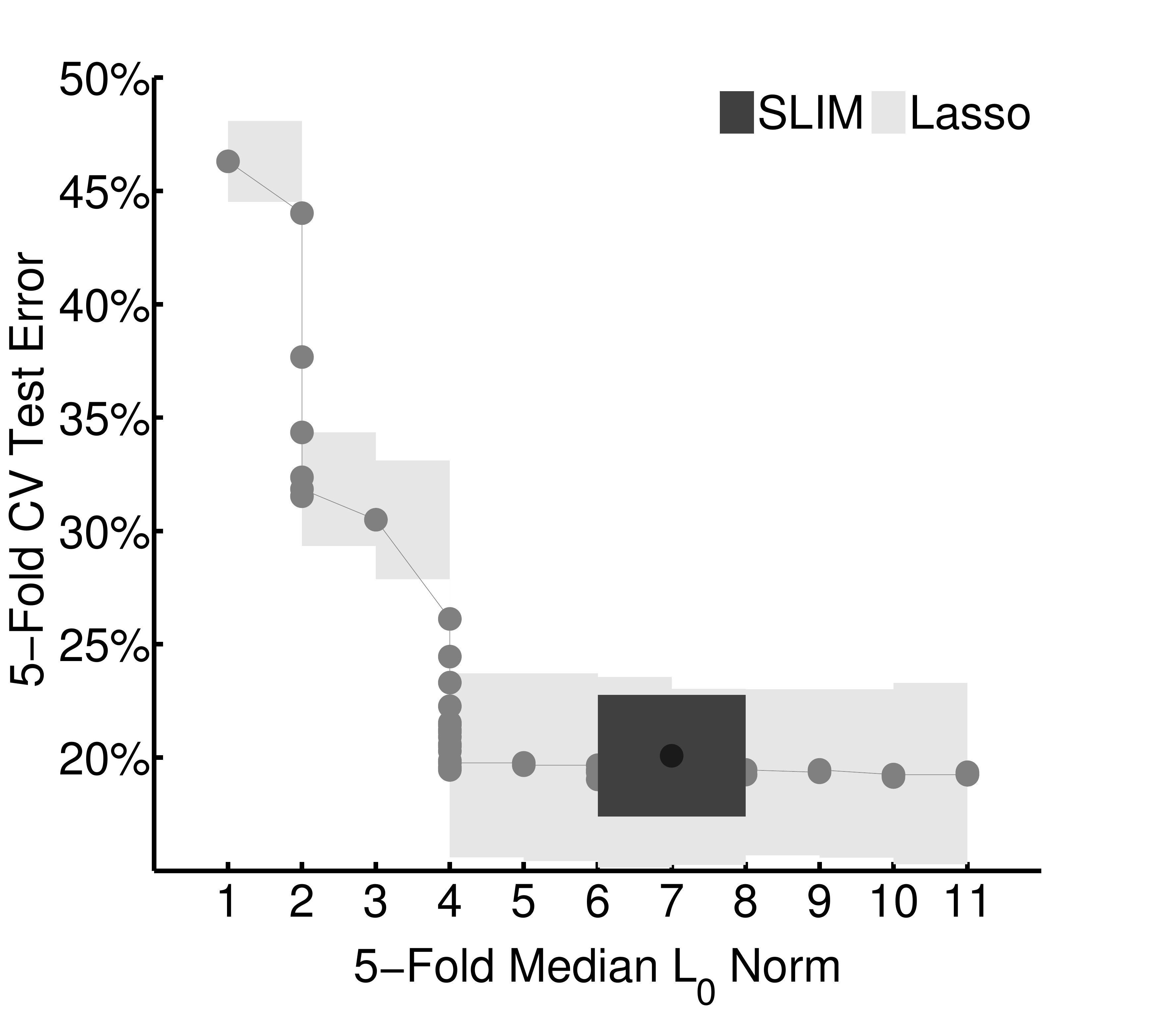}
    \caption{\mammo}
    \label{Fig::SLIMvsLasso_mammo}
\end{subfigure}
\begin{subfigure}[b]{0.475\textwidth}
    \centering
    \includegraphics[width=\textwidth]{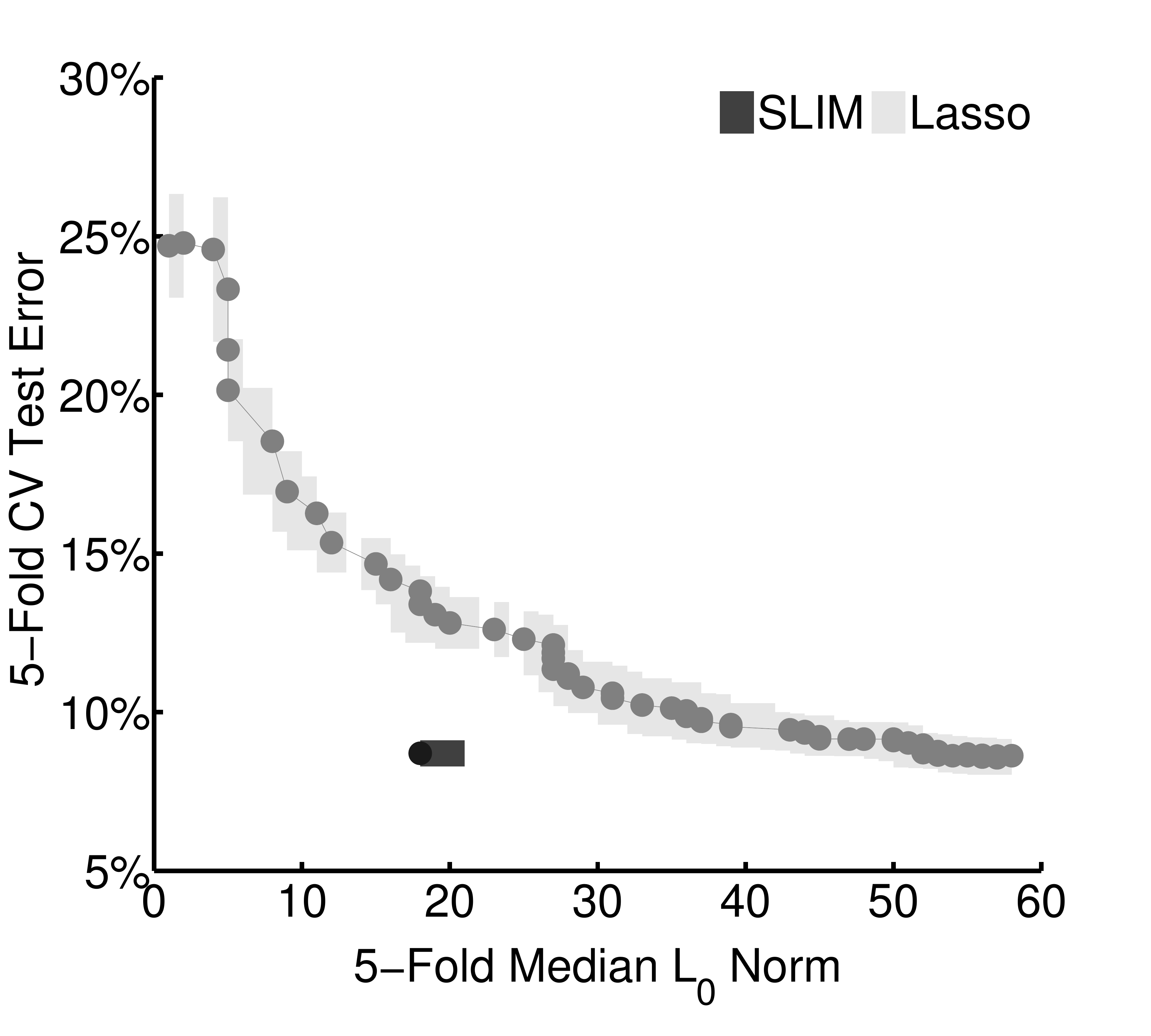}
    \caption{\spambase}
    \label{Fig::SLIMvsLasso_spambase}
\end{subfigure}
\hspace{0.1cm}
\begin{subfigure}[b]{0.475\textwidth}
    \centering
    \includegraphics[width=\textwidth]{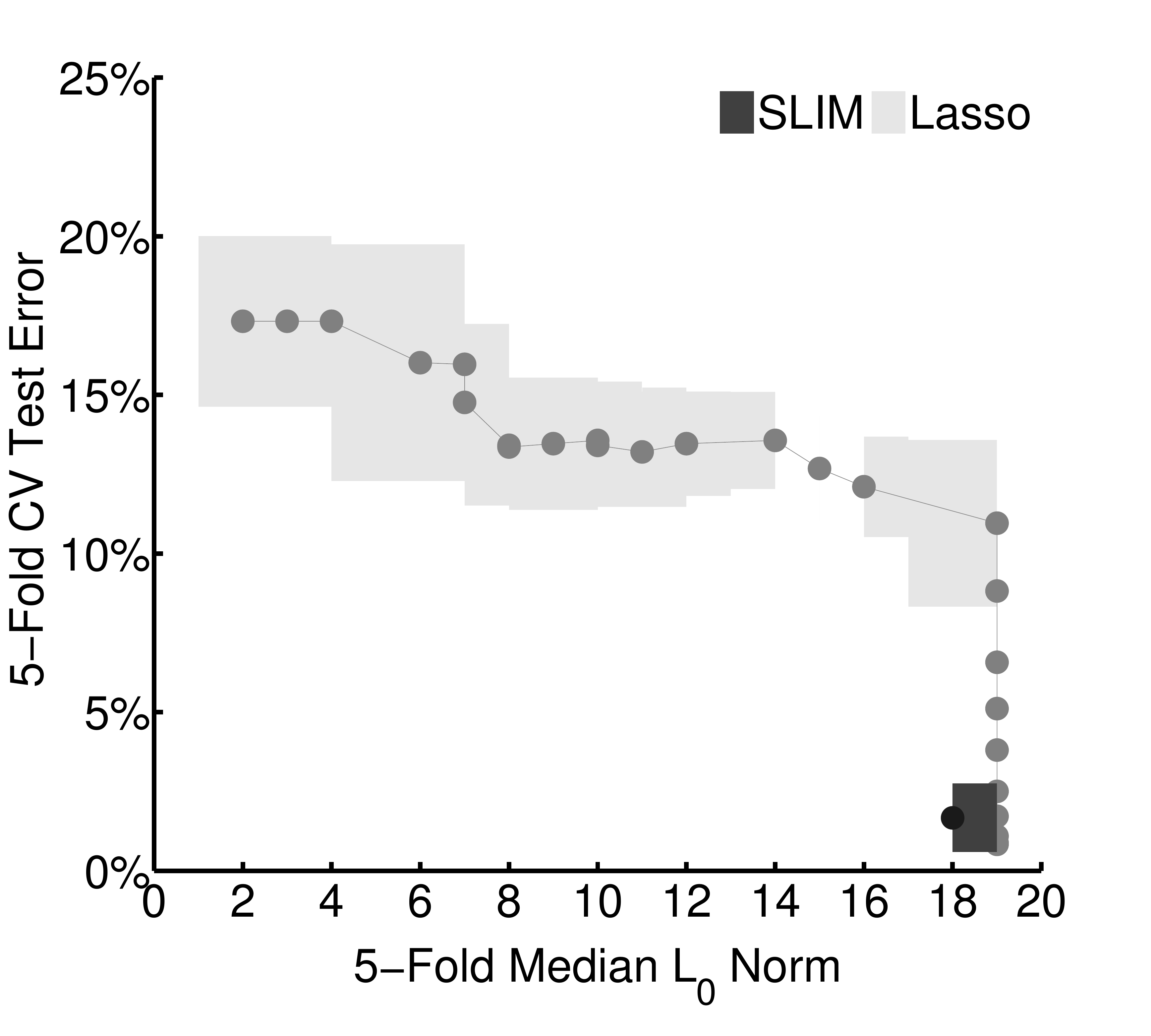}
    \caption{\tictactoe}
    \label{Fig::SLIMvsLasso_tictactoe}
\end{subfigure}
\end{figure}

LARS Lasso is a state-of-the-art method for generating sparse prediction models, which can efficiently adjust the value of the $\ell_1$-penalty to produce models at every possible level of sparsity. In Figures \ref{Fig::SLIMvsLasso_breastcancer}-\ref{Fig::SLIMvsLasso_tictactoe}, we compare the accuracy and sparsity of SLIM scoring systems to all models on the full regularization path of LARS Lasso. 
We plot LARS Lasso's performance in light gray with medium gray dots and SLIM's performance in dark gray with black dots. Our plots show that SLIM's classifiers dominated those of LARS Lasso for five out of the six datasets - even after accounting for all of the possible choices for LARS' regularization parameter. In the remaining dataset (\mammo) SLIM's performance was essentially tied with that of LARS Lasso for a particular value of its regularization parameter. This shows the effect of the approximate loss function and $\ell_1$ regularization term of LARS Lasso, which inadvertently adds strong additional regularization on the coefficient values in favor of convexity. 
\section{Conclusions}
Interpretability is not necessarily an important quality for all classification problems. Many problems in pattern recognition, for instance, require scalable methods that can produce quick and accurate predictions on a massive scale. Problems involving domain experts, however, require methods that are primarily accurate and interpretable -- especially because domain experts are unlikely to use models that they do not understand. Regardless of the problem at hand, models that are highly interpretable can not only relate variables to outcomes in a clear and convincing way, but also yield rules and insights as part of a data mining process. 

In spite of these benefits, interpretability has been difficult to address using existing classification models. One reason for this is a widely held belief that there is a trade-off between interpretability and predictive accuracy. In an article about the ``pitfalls" of prediction \citep{pitfall} in the National Institute of Justice (NIJ) journal, for instance, Greg Ridgeway states that ``there is often a tradeoff, with more interpretability coming at the expense of more predictive capacity." In this work, we have explicitly shown that this trade-off is not necessarily true.

In the same article \citep{pitfall}, Ridgeway also refers to a scoring system that the LAPD uses to identify recruits who are likely to become officers. In describing this scoring system, which uses 9 integer coefficients between 1 and 22, Ridgeway states: ``This simplicity gets at the important issue: A decent transparent model that is actually used will outperform a sophisticated system that predicts better but sits on a shelf. If the researchers had created a model that predicted well but was more complicated, the LAPD likely would have ignored it, thus defeating the whole purpose." We have shown in this paper that in many circumstances, it may be possible to have the best of both worlds: a learned model that is accurate and interpretable enough to be used in practice.

\newpage
\appendix

\section{Personalized Interpretable Linear Models}
\label{Appendix::PILM}
Given that interpretability is subjective, users may wish to produce a classification model that uses their own definition of interpretability. A Personalized Interpretable Linear Model (PILM) is a special case of \eqref{Eq::InterpretableOptimizationProblem} where users can train an accurate and interpretable classifier using an interpretability penalty function and a set of interpretability constraints that they have defined themselves.

To formulate this model as a discrete optimization problem, we first define $r = 1,\ldots, R$ interpretability sets $\Lset^1, \ldots, \Lset^R$, where
\begin{align}\label{Eq::Lsetr}
\Lset_r = \{\ell_{r,1},\ldots,\ell_{r,K_r}\} \text{ for } r = 1,\ldots, R
\end{align}
Next, we define an interpretability penalty function to penalizes coefficient $j$ by $C_r$ when $\lambda_j \in \Lset^r$,
\begin{align}\label{Eq::InterpretabilityPenalty}
\textrm{InterpretabilityPenalty}(\lambda_j) &= 
  \begin{cases}
    C_1 & \text{if} \quad \lambda_j \in \Lset_j^1\\
    & \vdots \\
    C_R & \text{if} \quad \lambda_j \in \Lset_j^R\\
  \end{cases}
\end{align}
We assume that the sets are mutually exclusively and that $\Lset_r$ is more interpretable than $\Lset^{r+1}$. When we set $0<C_0<C_1<\ldots<C_R$, optimizing the loss function and the interpretability penalty effectively regularizes for interpretability by penalizing coefficients that belong to less interpretable sets.

When this model minimizes the zero-one loss, the value of the $C_r$ represents the minimum gain in accuracy required to use a coefficient from $\Lset_j^r$. This implies that a coefficient $\lambda_j$ would only belongs to $\Lset_j^{r+1}$ instead of $\Lset_j^r$ if there exists a coefficient in $\Lset_j^{r+1}$ that yields a gain in accuracy of more than $C_{r+1}-C_{r}$. This convenient fact allows us to purposefully set the values of $C_r$ a priori. 

%
%
Once we have chosen an appropriate set of interpretability constraints and designed a suitable interpretability penalty function, we can produce a classifier that balances accuracy and interpretability by solving the following mixed-integer program:
\begin{equationarray}{crcl>{\hspace{0.1cm}}ll}
\min_{\bm{z},\lambdab} &\frac{1}{N}\sum_{i=1}^{N} z_i & + &  \sum_{j=1}^P I_j  \notag \\
\textrm{s.t.} & M_iz_i & \geq & \gamma -\sum_{j=1}^P y_ix_{ij}\lambda_j & \mprange{i}{1}{N} & \mpdes{0-1 loss} \label{Con::GIMLoss} \\
& & & & & \notag \\
&\lambda_j & = & \sum_{r=1}^R \sum_{k=1}^{K_r} l_{r,k} u_{j,k,r}     &\mprange{j}{1}{P} & \mpdes{coefficient values} \label{Con::GIMIntCon} \\
& & & & & \notag \\
& I_j & = &\sum_{r=1}^R C_r s_{j,r} & \mprange{j}{1}{P} & \mpdes{int. penalty} \label{Con::GIMIntPenalty} \\
& s_{j,r} & = & \sum_{k=1}^{K_r} u_{j,r,k} & \mprange{j}{1}{P} \quad \mprange{r}{1}{R} & \mpdes{int. level indicator}\label{Con::GIMIntPenaltySet} \\
& \sum_{r=1}^R s_{j,r} & = & 1 & \mprange{j}{1}{P} & \mpdes{restrict coef. to 1 int. level} \label{Con::GIMIntPenaltySet2} \\
& z_i & \in & \B &  \mprange{i}{1}{N} & \mpdes{0-1 loss indicators} \notag  \\
& s_{j,r} & \in & \B &  \mprange{j}{1}{P} \quad  \mprange{r}{1}{R} & \mpdes{int. level indicators} \notag \\ 
& u_{j,r,k} & \in & \B &  \mprange{j}{1}{P} \quad \mprange{r}{1}{R} \quad \mprange{k}{1}{K_r} & \mpdes{coef. value indicators} \notag 
\end{equationarray}

Here:

\begin{itemize}

\item The variables $z_i = \indic{y_i \mathbf{x}_i^T \lambdab \leq 0}$ are binary indicator variables that indicate a misclassification. Each $z_i$ is set to $1$ if the classifier $\lambdab$ makes a mistake on example $i$ though the constraints \eqref{Con::GIMLoss}. These constraints depend on user-defined scalar parameters $\gamma$ and $M_i$. We typically set $\gamma = 0.1$ and $M_i = \gamma - \max{\sum_{j=1}^P y_ix_{ij}}$. 
\item The variables $I_j$ are auxiliary variables to represent the interpretability of each coefficient $\lambda_j$. The value of $I_j$ is set as $C_r$ if and only if $\lambda_j$ belongs to the interpretability set $\Lset_j^r$  in constraint  \ref{Con::GIMIntPenalty}. This depends on constraints \ref{Con::GIMIntPenaltySet} and \ref{Con::GIMIntPenaltySet2} which force the binary variables $s_{j,r}$ to 1 if and only if $\lambda_j \in \Lset_j^r$.

\item Constraints \eqref{Con::GIMIntCon} restrict $\lambda_j$ to one of the values of $\Lset_r$ using a one-of-K formulation. The $u_{j,k,r}$ are binary variables that are equal to 1 if $\lambda_j$ is equal to $l_{k,r}$. We ensure that $\lambda_j$ can only take a single value from a single interpretability by limiting the sum of $u_{j,k,r}$ to 1 using constraints \eqref{Con::GIMIntPenalty} and \eqref{Con::GIMIntValueSet}.

\end{itemize}


\clearpage
\section{SLIM for Imbalanced Datasets}
\label{Appendix::SLIMImbalancedModel}
Many scoring systems are used to detect rare events, such as a heart attack or a violent crime. In these cases, training a classifier that maximizes the classification accuracy often results in degeneracy (e.g. if the probability of heart attack is 1\%, any classifier that never predicts a heart attack is 99\% accurate). Faced with such imbalanced data, we can adjust SLIM's objective function to optimize a user-specified balance between sensitivity and specificity as follows:
\begin{align*}\centering
\min_{\lambdab \in \Lset} \, \frac{\wplus}{N}\sum_{i\in\mathcal{I}^+} \indic{ \xb_i^T \lambdab \leq 0} + \frac{\wminus}{N} \sum_{i\in\mathcal{I}^-} \indic{ \xb_i^T \lambdab \leq 0} + C_0\vnorm{\lambdab}_0 + C_1\vnorm{\lambdab}_1 .
\end{align*}
Here, $\mathcal{I}^+ = \{i:y_i=+1\}$ is the set of examples with positive labels and $\mathcal{I}^- = \{i:y_i=-1\}$ is the set of examples with negative labels. The parameters $\wplus$ and $\wminus$ are user-defined scalars that balance the sensitivity and specificity of the SLIM classifier. Setting the values of these weights to $\wplus=\wminus=1$ reproduces the standard version of SLIM that we present in Section \ref{Sec::SLIMBalancedModel}. By default, we set the values of the weights to $\wplus = \frac{N}{2|\mathcal{I}^+|}$ and $\wminus \frac{N}{2|\mathcal{I}^-|}$ in order to produce a classifier that is just as accurate on the positive examples as it is on the negative examples. 

This objective can also be optimized using an MIP with the same number of variables as constraints as the standard version of SLIM that we present in Section \ref{Sec::SLIMBalancedModel}:
\begin{equationarray}{crcl>{\hspace{1cm}}l>{\hspace{1cm}}r}
\min_{\bm{z},\lambdab} &\frac{\wplus}{N} \sum_{i\in\mathcal{I}^+} z_i & + & \frac{\wminus}{N} \sum_{i\in\mathcal{I}^-} z_i + \sum_{j=1}^P I_j & \notag \\
\textrm{s.t.}          & M_i z_i                  & \geq & \gamma -\sum_{j=1}^P x_{i,j}\lambda_j          &i \in \mathcal{I}^+ & \mpdes{0-1 loss for $i:y_i=+1$} \label{Con::ImbSLIMLossPos} \\
                       & M_i z_i                  & \geq & \gamma +\sum_{j=1}^P x_{i,j}\lambda_j          &i \in \mathcal{I}^- & \mpdes{0-1 loss for $i:y_i=-1$} \label{Con::ImbSLIMLossNeg} \\
& & & & & \notag \\
& \lambdab & \in & \Lset & & \mpdes{coefficient values} \notag \\ 
& & & & & \notag \\
& I_j & = & C_0\alpha_j + C_1\beta_j &\mprange{j}{1}{P}& \mpdes{int. penalty} \notag \\ 
&\Lambda_j\alpha_j    & \geq & \lambda_j   &\mprange{j}{1}{P} & \mpdes{$\ell_0$-norm \#2} \notag \\ 
&\Lambda_j\alpha_j    & \geq & -\lambda_j   &\mprange{j}{1}{P} & \mpdes{$\ell_0$-norm \#2} \notag \\ 
&\beta_j              & \geq & \lambda_j   &\mprange{j}{1}{P} & \mpdes{$\ell_1$-norm \#1} \notag \\ 
&\beta_j              & \geq & -\lambda_j   &\mprange{j}{1}{P} & \mpdes{$\ell_1$-norm \#2} \notag \\ 
& & & & & \notag \\
& z_i & \in & \B &  \mprange{i}{1}{N} & \mpdes{0-1 loss indicators} \notag  \\
& I_j  & \in & \R_+  & \mprange{j}{1}{P} & \mpdes{int. penalty values} \notag \\
& \alpha_j  & \in & \B  & \mprange{j}{1}{P} & \mpdes{$\ell_0$ indicators} \notag \\
& \beta_j    & \in & \R_+ & \mprange{j}{1}{P} & \mpdes{abs. value variables} \notag
\end{equationarray}

In this formulation, constraints \eqref{Con::ImbSLIMLossPos} ensure that  $z_i = \indic{\hat{y}_i\neq 1}$ for examples where $y_i=+1$, and constraints (\ref{Con::ImbSLIMLossNeg}) ensure that $z_i = \indic{\hat{y}_i\neq -1}$ for examples where $y_i=-1$. The remaining variables, constraints and parameters are analogous to those from the formulation in Section \ref{Sec::SLIMBalancedModel}.

\newpage
\section{Table of Results on Sparsity vs. Accuracy}
\label{Appendix::TableOfExperimentalResults}
We summarize the results from our numerical experiments in Section \ref{Sec::NumericalExperiments} in Tables  \ref{Table::ComparisonAccuracy} and \ref{Table::ComparisonSparsity}. The results in Table\ref{Table::ComparisonAccuracy} reflect the performance of the baseline algorithms when we have set free parameters so as to minimize the mean 5-fold cross-validation (CV) error. In Table \ref{Table::ComparisonSparsity}, the last 4 columns report results corresponding to the the sparsest model that was within one standard deviation of the accuracy of the model produced in Table \ref{Table::ComparisonAccuracy} (the remaining of the columns were reproduced from Table \ref{Table::ComparisonAccuracy} to allow easier comparison between methods).\\

\noindent Note that both Tables \ref{Table::ComparisonAccuracy} and \ref{Table::ComparisonSparsity} bundle the following experimental results for a given dataset:
\begin{itemize}[leftmargin=0.45cm,topsep=4pt,parsep=2pt]
\item{\textbf{test error}}, corresponding to the 5-fold CV mean and standard deviation of the test error;
\item{\textbf{train error}}, corresponding to the 5-fold CV mean and standard deviation of the train error;
\item{\textbf{model size}}, corresponding to the 5-fold CV median model size;
\item{\textbf{model range}}, corresponding to the interval between the 5-fold CV minimum model size and the 5-fold CV maximum model size.
\end{itemize}

\noindent As a reminder, model size corresponds to the number of coefficients for SLIM, LR, Lasso, Ridge and EN, the number of leaves for C5.0T and CART, and the number of rules for C5.0R. For RF and SVM, we have set the model size to the number of features in each dataset as the statistic is meaningless for these methods.

\begin{table}[htbp]
 \footnotesize
  \centering
  \caption{Accuracy vs. Sparsity for All Methods (Emphasis on Accuracy).}
  \vspace{0.2cm}
   \resizebox{6.5cm}{!} {\centering
   \rotatebox{90}{
    \begin{tabular}{llcccccccccc}
    \toprule
     \textbf{Dataset} & \textbf{Metric} & \textbf{LR} & \textbf{CART} & \textbf{RF} & \textbf{SVM} & \textbf{C50T} & \textbf{C50R} & \textbf{Lasso} & \textbf{Ridge} & \textbf{EN} & \textbf{SLIM} \\
    \midrule
\multirow{4}{*}{\breastcancer} & test error & 3.7 $\pm$ 0.9\% & 5.9 $\pm$ 1.5\% & 2.7 $\pm$ 0.9\% & 2.9 $\pm$ 1.4\% & 5.3 $\pm$ 2.0\% & 4.7 $\pm$ 1.2\% & 3.2 $\pm$ 0.3\% & 3.2 $\pm$ 0.3\% & 3.2 $\pm$ 0.3\% & 3.7 $\pm$ 1.7\% \\
          & train error & 2.9 $\pm$ 0.3\% & 4.0 $\pm$ 0.6\% & 3.0 $\pm$ 0.4\% & 2.3 $\pm$ 0.2\% & 2.6 $\pm$ 0.4\% & 2.6 $\pm$ 0.5\% & 2.9 $\pm$ 0.3\% & 3.0 $\pm$ 0.2\% & 2.9 $\pm$ 0.3\% & 3.1 $\pm$ 0.4\% \\
          & model size & 9     & 4     & 10    & 10    & 8     & 6     & 9     & 10    & 10    & 3 \\
          & model range & 9 - 9 & 3 - 7 & 10 - 10 & 10 - 10 & 6 - 10 & 4 - 8 & 9 - 10 & 10 - 10 & 10 - 10 & 3 - 3 \\\hline
\multirow{4}{*}{\haberman} & test error & 26.5 $\pm$ 7.5\% & 26.8 $\pm$ 8.9\% & 28.1 $\pm$ 7.9\% & 26.2 $\pm$ 8.5\% & 27.8 $\pm$ 6.7\% & 27.8 $\pm$ 6.7\% & 25.8 $\pm$ 2.6\% & 26.1 $\pm$ 2.7\% & 25.8 $\pm$ 2.6\% & 23.2 $\pm$ 6.5\% \\
          & train error & 25.2 $\pm$ 1.9\% & 20.4 $\pm$ 1.8\% & 27.9 $\pm$ 2.3\% & 19.7 $\pm$ 2.0\% & 23.7 $\pm$ 2.1\% & 23.7 $\pm$ 2.1\% & 26.7 $\pm$ 1.6\% & 26.1 $\pm$ 1.5\% & 25.7 $\pm$ 1.7\% & 21.6 $\pm$ 1.9\% \\
          & model size & 3     & 6     & 4     & 4     & 3     & 3     & 2     & 4     & 4     & 3 \\
          & model range & 3 - 3 & 4 - 7 & 4 - 4 & 4 - 4 & 1 - 3 & 0 - 3 & 2 - 3 & 4 - 4 & 4 - 4 & 3 - 3 \\\hline
\multirow{4}{*}{\internetad} & test error & 8.5 $\pm$ 1.4\% & 4.5 $\pm$ 1.4\% & 2.5 $\pm$ 0.8\% & 3.7 $\pm$ 1.0\% & 3.9 $\pm$ 0.9\% & 4.1 $\pm$ 0.9\% & 2.7 $\pm$ 0.4\% & 5.6 $\pm$ 0.6\% & 2.7 $\pm$ 0.4\% & 3.6 $\pm$ 0.8\% \\
          & train error & 0.5 $\pm$ 0.2\% & 3.4 $\pm$ 0.1\% & 2.5 $\pm$ 0.2\% & 0.1 $\pm$ 0.0\% & 2.9 $\pm$ 0.5\% & 3.2 $\pm$ 0.4\% & 1.2 $\pm$ 0.2\% & 5.1 $\pm$ 0.4\% & 0.5 $\pm$ 0.2\% & 2.8 $\pm$ 0.3\% \\
          & model size & 616   & 7     & 1431  & 1431  & 10    & 5     & 118   & 1425  & 560   & 14 \\
          & model range & 606 - 621 & 6 - 7 & 1431 - 1431 & 1431 - 1431 & 8 - 20 & 4 - 8 & 103 - 128 & 1410 - 1428 & 443 - 588 & 14 - 14 \\\hline
\multirow{4}{*}{\mammo}  & test error & 19.2 $\pm$ 3.9\% & 21.4 $\pm$ 3.5\% & 20.2 $\pm$ 4.5\% & 21.0 $\pm$ 3.8\% & 19.8 $\pm$ 3.7\% & 20.1 $\pm$ 4.3\% & 19.0 $\pm$ 1.7\% & 19.5 $\pm$ 1.7\% & 21.4 $\pm$ 1.2\% & 20.1 $\pm$ 2.7\% \\
          & train error & 19.0 $\pm$ 1.1\% & 19.5 $\pm$ 1.0\% & 20.4 $\pm$ 1.1\% & 19.2 $\pm$ 1.1\% & 18.9 $\pm$ 1.2\% & 18.9 $\pm$ 1.2\% & 19.3 $\pm$ 1.1\% & 25.7 $\pm$ 0.7\% & 20.7 $\pm$ 0.8\% & 17.8 $\pm$ 1.0\% \\
          & model size & 9     & 5     & 12     & 12     & 6     & 5     & 6   & 12     & 10     & 7 \\
          & model range & 9 - 9 & 3 - 6 & 12 - 12 & 12 - 12 & 5 - 13 & 3 - 10 & 6 - 7 & 12 - 12 & 9 - 12 & 6 - 8 \\\hline
\multirow{4}{*}{\spambase}  & test error & 7.3 $\pm$ 1.0\% & 10.6 $\pm$ 1.3\% & 4.8 $\pm$ 0.3\% & 6.5 $\pm$ 0.4\% & 7.7 $\pm$ 0.4\% & 6.8 $\pm$ 0.4\% & 7.1 $\pm$ 0.5\% & 8.8 $\pm$ 0.4\% & 7.1 $\pm$ 0.5\% & 7.4 $\pm$ 0.8\% \\
          & train error & 6.9 $\pm$ 0.3\% & 9.8 $\pm$ 0.3\% & 5.0 $\pm$ 0.0\% & 3.3 $\pm$ 0.1\% & 4.3 $\pm$ 0.2\% & 4.6 $\pm$ 0.2\% & 6.8 $\pm$ 0.3\% & 8.5 $\pm$ 0.2\% & 6.9 $\pm$ 0.4\% & 6.6 $\pm$ 0.6\% \\
          & model size & 57    & 7     & 58    & 58    & 63    & 26    & 57    & 58    & 57    & 18 \\
          & model range & 57 - 57 & 6 - 9 & 58 - 58 & 58 - 58 & 57 - 73 & 23 - 27 & 55 - 58 & 58 - 58 & 55 - 58 & 18 - 21 \\\hline
\multirow{4}{*}{\tictactoe}  & test error & 2.7 $\pm$ 1.1\% & 11.7 $\pm$ 4.1\% & 1.6 $\pm$ 0.8\% & 0.7 $\pm$ 0.6\% & 7.5 $\pm$ 2.0\% & 2.6 $\pm$ 2.1\% & 1.7 $\pm$ 0.3\% & 16.4 $\pm$ 1.0\% & 1.7 $\pm$ 0.3\% & 3.3 $\pm$ 2.2\% \\
          & train error & 2.3 $\pm$ 0.8\% & 6.8 $\pm$ 2.1\% & 2.4 $\pm$ 0.3\% & 0.0 $\pm$ 0.0\% & 2.6 $\pm$ 0.6\% & 0.7 $\pm$ 0.1\% & 1.6 $\pm$ 0.1\% & 15.1 $\pm$ 0.5\% & 1.7 $\pm$ 0.2\% & 2.1 $\pm$ 1.8\% \\
          & model size & 18    & 21    & 28    & 28    & 39    & 19    & 19    & 28    & 19    & 18 \\
          & model range & 18 - 18 & 21 - 23 & 28 - 28 & 28 - 28 & 28 - 42 & 15 - 26 & 19 - 19 & 28 - 28 & 19 - 19 & 18 - 19 \\
    \bottomrule
    \end{tabular}%
    }}
    \label{Table::ComparisonAccuracy}%
\end{table} 


\begin{table}[htbp]
 \footnotesize
  \centering
  \caption{Accuracy vs. Sparsity for All Methods (Emphasis on Accuracy).}
  \vspace{0.2cm}
   \resizebox{6.5cm}{!} {\centering
   \rotatebox{90}{
    \begin{tabular}{llcccccccccc}
    \toprule
     \textbf{Dataset} & \textbf{Metric} & \textbf{LR} & \textbf{CART} & \textbf{RF} & \textbf{SVM} & \textbf{C50T} & \textbf{C50R} & \textbf{Lasso} & \textbf{Ridge} & \textbf{EN} & \textbf{SLIM} \\
    \midrule
\multirow{4}{*}{\breastcancer} & test error & 3.7 $\pm$ 0.9\% & 5.9 $\pm$ 1.5\% & 2.7 $\pm$ 0.9\% & 2.9 $\pm$ 1.4\% & 5.3 $\pm$ 2.0\% & 4.7 $\pm$ 1.2\% & 3.5 $\pm$ 0.4\% & 3.5 $\pm$ 0.6\% & 3.5 $\pm$ 0.4\% & 4.8 $\pm$ 1.6\% \\
          & train error & 2.9 $\pm$ 0.3\% & 4.0 $\pm$ 0.6\% & 3.0 $\pm$ 0.4\% & 2.3 $\pm$ 0.2\% & 2.6 $\pm$ 0.4\% & 2.6 $\pm$ 0.5\% & 3.1 $\pm$ 0.4\% & 3.4 $\pm$ 0.3\% & 2.8 $\pm$ 0.2\% & 4.3 $\pm$ 0.7\% \\
          & model size & 9     & 4     & 10    & 10    & 8     & 6     & 10    & 10    & 10    & 3 \\
          & model range & 9 - 9 & 3 - 7 & 10 - 10 & 10 - 10 & 6 - 10 & 4 - 8 & 10 - 10 & 10 - 10 & 10 - 10 & 3 - 3 \\\hline
\multirow{4}{*}{ \haberman} & test error & 26.5 $\pm$ 7.5\% & 26.8 $\pm$ 8.9\% & 28.1 $\pm$ 7.9\% & 26.2 $\pm$ 8.5\% & 27.8 $\pm$ 6.7\% & 27.8 $\pm$ 6.7\% & 26.5 $\pm$ 2.4\% & 26.5 $\pm$ 2.4\% & 26.5 $\pm$ 2.4\% & 26.5 $\pm$ 4.8\% \\
          & train error & 25.2 $\pm$ 1.9\% & 20.4 $\pm$ 1.8\% & 27.9 $\pm$ 2.3\% & 19.7 $\pm$ 2.0\% & 23.7 $\pm$ 2.1\% & 23.7 $\pm$ 2.1\% & 26.5 $\pm$ 1.3\% & 26.5 $\pm$ 1.3\% & 26.9 $\pm$ 1.4\% & 26.5 $\pm$ 1.2\% \\
          & model size & 3     & 6     & 4     & 4     & 3     & 3     & 2     & 4     & 4     & 1 \\
          & model range & 3 - 3 & 4 - 7 & 4 - 4 & 4 - 4 & 1 - 3 & 0 - 3 & 1 - 2 & 4 - 4 & 4 - 4 & 1 - 1 \\\hline
\multirow{4}{*}{ \internetad} & test error & 8.5 $\pm$ 1.4\% & 4.5 $\pm$ 1.4\% & 2.5 $\pm$ 0.8\% & 3.7 $\pm$ 1.0\% & 3.9 $\pm$ 0.9\% & 4.1 $\pm$ 0.9\% & 3.1 $\pm$ 0.6\% & 6.1 $\pm$ 0.4\% & 3.1 $\pm$ 0.6\% & 3.6 $\pm$ 0.8\% \\
          & train error & 0.5 $\pm$ 0.2\% & 3.4 $\pm$ 0.1\% & 2.5 $\pm$ 0.2\% & 0.1 $\pm$ 0.0\% & 2.9 $\pm$ 0.5\% & 3.2 $\pm$ 0.4\% & 2.4 $\pm$ 0.2\% & 5.7 $\pm$ 0.4\% & 0.7 $\pm$ 0.2\% & 2.8 $\pm$ 0.3\% \\
          & model size & 616   & 7     & 1431  & 1431  & 10    & 5     & 62    & 1425  & 473   & 14 \\
          & model range & 606 - 621 & 6 - 7 & 1431 - 1431 & 1431 - 1431 & 8 - 20 & 4 - 8 & 55 - 64 & 1410 - 1428 & 371 - 502 & 14 - 14 \\\hline
\multirow{4}{*}{ \mammo}  & test error & 19.2 $\pm$ 3.9\% & 21.4 $\pm$ 3.5\% & 20.2 $\pm$ 4.5\% & 21.0 $\pm$ 3.8\% & 19.8 $\pm$ 3.7\% & 20.1 $\pm$ 4.3\% & 20.6 $\pm$ 1.7\% & 21.0 $\pm$ 1.1\% & 22.6 $\pm$ 1.9\% & 23.6 $\pm$ 3.7\% \\
          & train error & 19.0 $\pm$ 1.1\% & 19.5 $\pm$ 1.0\% & 20.4 $\pm$ 1.1\% & 19.2 $\pm$ 1.1\% & 18.9 $\pm$ 1.2\% & 18.9 $\pm$ 1.2\% & 20.3 $\pm$ 1.2\% & 20.8 $\pm$ 0.8\% & 20.9 $\pm$ 0.7\% & 21.6 $\pm$ 1.2\% \\
          & model size & 9     & 5     & 12     & 12     & 6     & 5     & 4   & 12     & 10     & 2 \\
          & model range & 9 - 9 & 3 - 6 & 12 - 12 & 12 - 12 & 5 - 13 & 3 - 10 & 4 - 4 & 12 - 12 & 9 - 10 & 2 - 3 \\\hline
\multirow{4}{*}{ \spambase} & test error & 7.3 $\pm$ 1.0\% & 10.6 $\pm$ 1.3\% & 4.8 $\pm$ 0.3\% & 6.5 $\pm$ 0.4\% & 7.7 $\pm$ 0.4\% & 6.8 $\pm$ 0.4\% & 7.6 $\pm$ 0.5\% & 9.1 $\pm$ 0.4\% & 7.6 $\pm$ 0.5\% & 7.4 $\pm$ 0.8\% \\
          & train error & 6.9 $\pm$ 0.3\% & 9.8 $\pm$ 0.3\% & 5.0 $\pm$ 0.0\% & 3.3 $\pm$ 0.1\% & 4.3 $\pm$ 0.2\% & 4.6 $\pm$ 0.2\% & 7.1 $\pm$ 0.1\% & 8.9 $\pm$ 0.3\% & 7.2 $\pm$ 0.1\% & 6.6 $\pm$ 0.6\% \\
          & model size & 57    & 7     & 58    & 58    & 63    & 26    & 52    & 58    & 52    & 18 \\
          & model range & 57 - 57 & 6 - 9 & 58 - 58 & 58 - 58 & 57 - 73 & 23 - 27 & 50 - 54 & 58 - 58 & 51 - 54 & 18 - 21 \\\hline
\multirow{4}{*}{ \tictactoe} & test error & 2.7 $\pm$ 1.1\% & 11.7 $\pm$ 4.1\% & 1.6 $\pm$ 0.8\% & 0.7 $\pm$ 0.6\% & 7.5 $\pm$ 2.0\% & 2.6 $\pm$ 2.1\% & 1.7 $\pm$ 0.3\% & 16.8 $\pm$ 0.9\% & 1.7 $\pm$ 0.3\% & 3.3 $\pm$ 2.2\% \\
          & train error & 2.3 $\pm$ 0.8\% & 6.8 $\pm$ 2.1\% & 2.4 $\pm$ 0.3\% & 0.0 $\pm$ 0.0\% & 2.6 $\pm$ 0.6\% & 0.7 $\pm$ 0.1\% & 1.6 $\pm$ 0.1\% & 15.9 $\pm$ 0.8\% & 1.7 $\pm$ 0.2\% & 2.1 $\pm$ 1.8\% \\
          & model size & 18    & 21    & 28    & 28    & 39    & 19    & 19    & 28    & 19    & 18 \\
          & model range & 18 - 18 & 21 - 23 & 28 - 28 & 28 - 28 & 28 - 42 & 15 - 26 & 19 - 19 & 28 - 28 & 19 - 19 & 18 - 19 \\
    \bottomrule
    \end{tabular}%
    }
    }
    \label{Table::ComparisonSparsity}%
\end{table}

\clearpage
\section{Computational Performance of SLIM}
\label{Appendix::ComputationalPerformance}
\begin{figure}[htbp]
\centering
\begin{subfigure}[b]{0.33\textwidth}
    \centering
        \includegraphics[width=\textwidth,trim=5 20 5 50, clip]{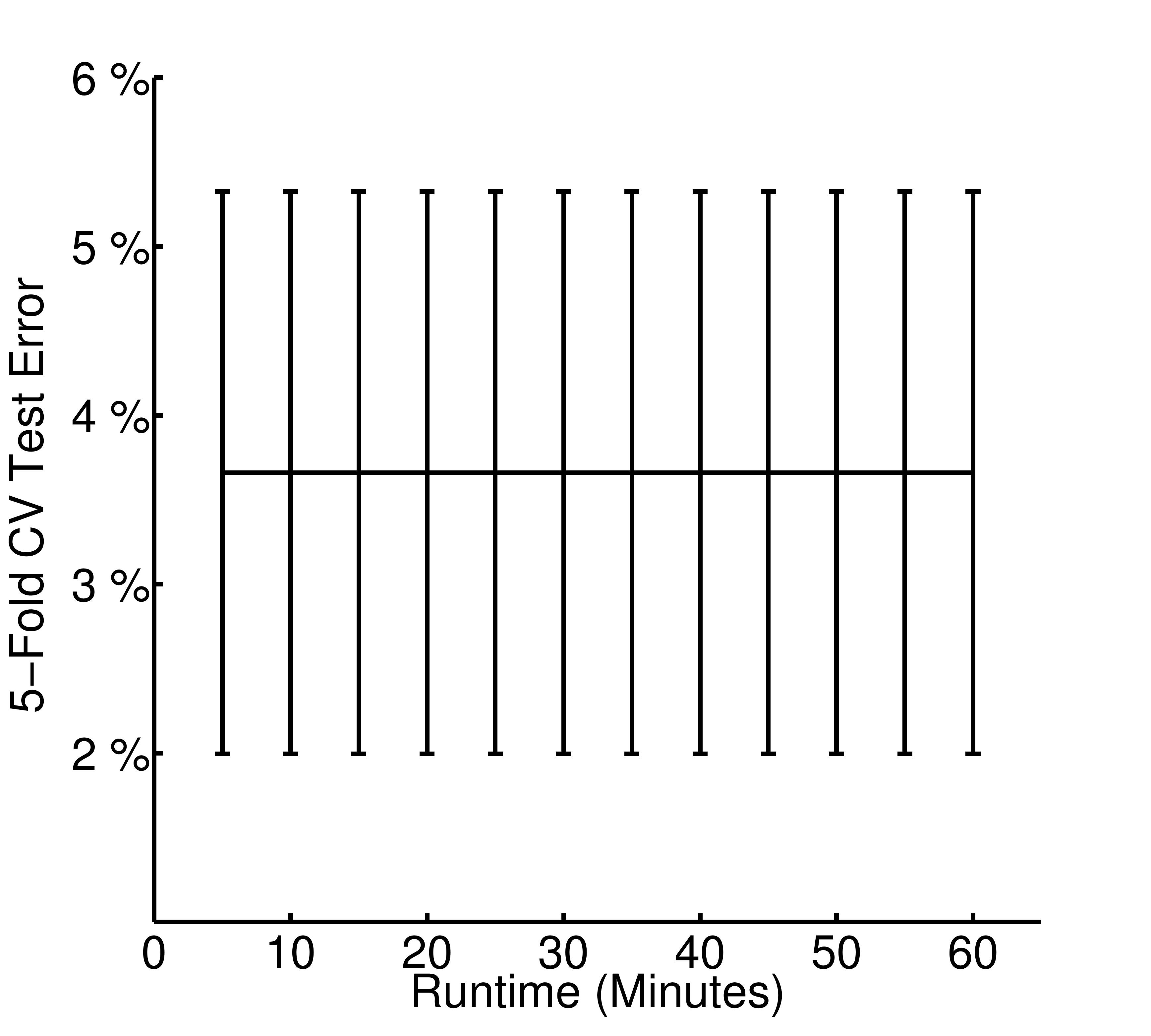}
    \label{S6_RUNTIME_TEST_ERRS_breastcancer}
\end{subfigure}
\hspace{0.1cm}
\begin{subfigure}[b]{0.33\textwidth}
    \centering
    \includegraphics[width=\textwidth,trim=5 20 5 50, clip]{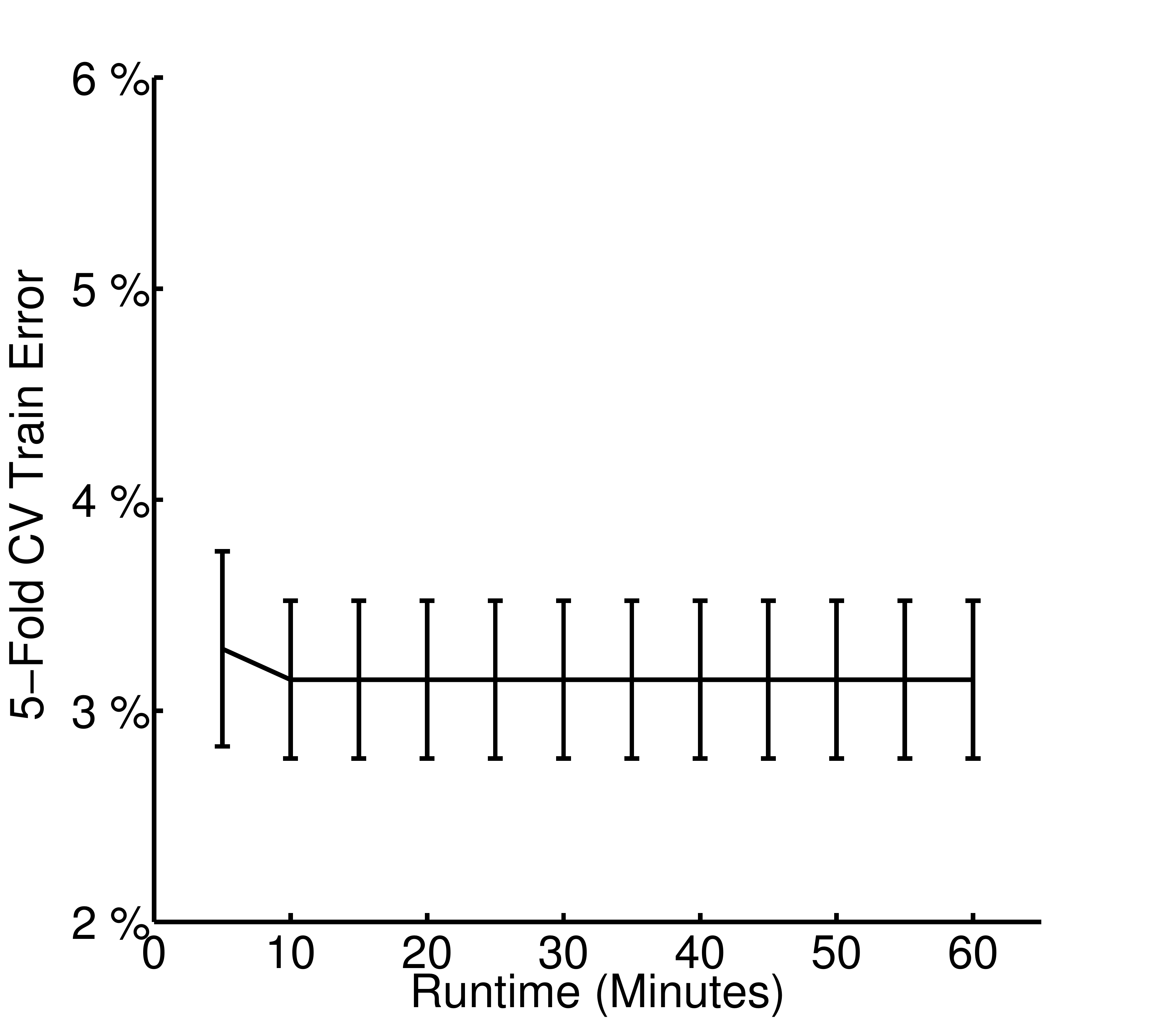}
    \label{S6_RUNTIME_TRAIN_ERRS_breastcancer}
\end{subfigure}

\begin{subfigure}[b]{0.33\textwidth}
    \centering
    \includegraphics[width=\textwidth,trim=5 20 5 50, clip]{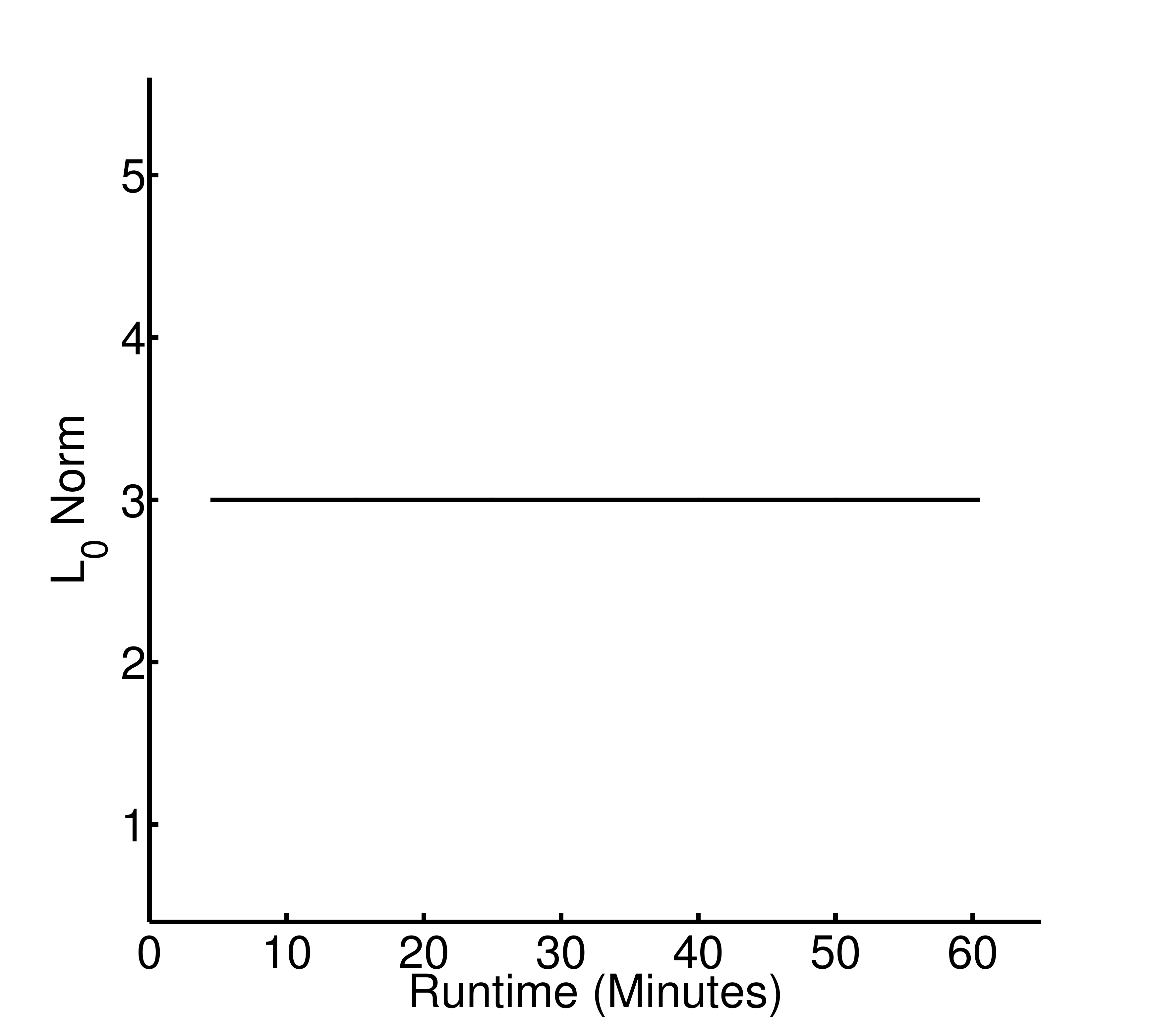}
    \label{S6_RUNTIME_LO_NORMS_breastcancer}
\end{subfigure}
\hspace{0.1cm}
\begin{subfigure}[b]{0.33\textwidth}
    \centering
    \includegraphics[width=\textwidth,trim=5 20 5 50, clip]{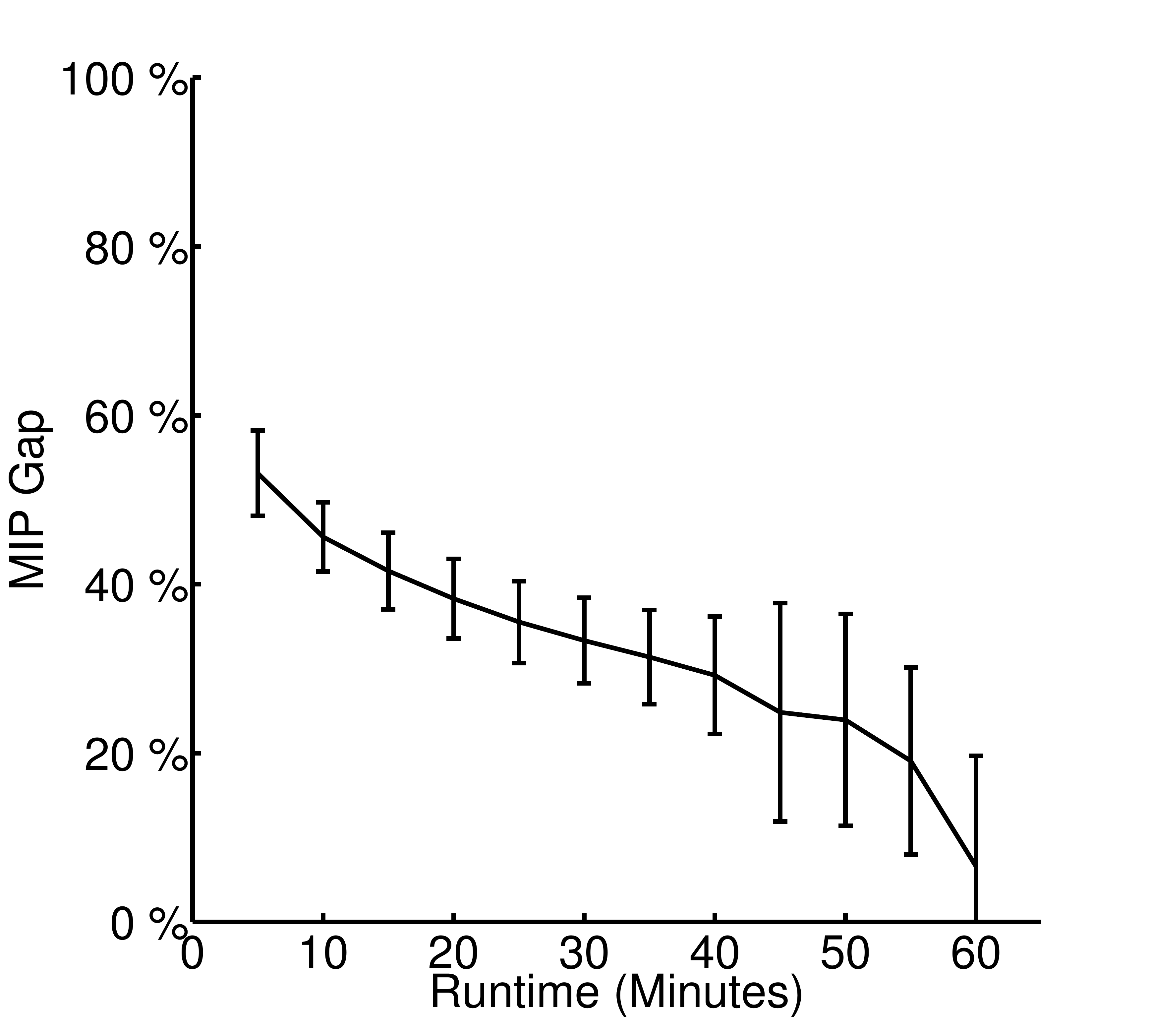}
    \label{S6_RUNTIME_MIPGAPS_breastcancer}
\end{subfigure}
\caption{Computational performance over time for \breastcancer.}
\label{Fig::CompPlots_breastcancer}
\end{figure}

\begin{figure}[htbp]
\centering
\begin{subfigure}[b]{0.30\textwidth}
    \centering
    \includegraphics[width=\textwidth,trim=5 20 5 50, clip]{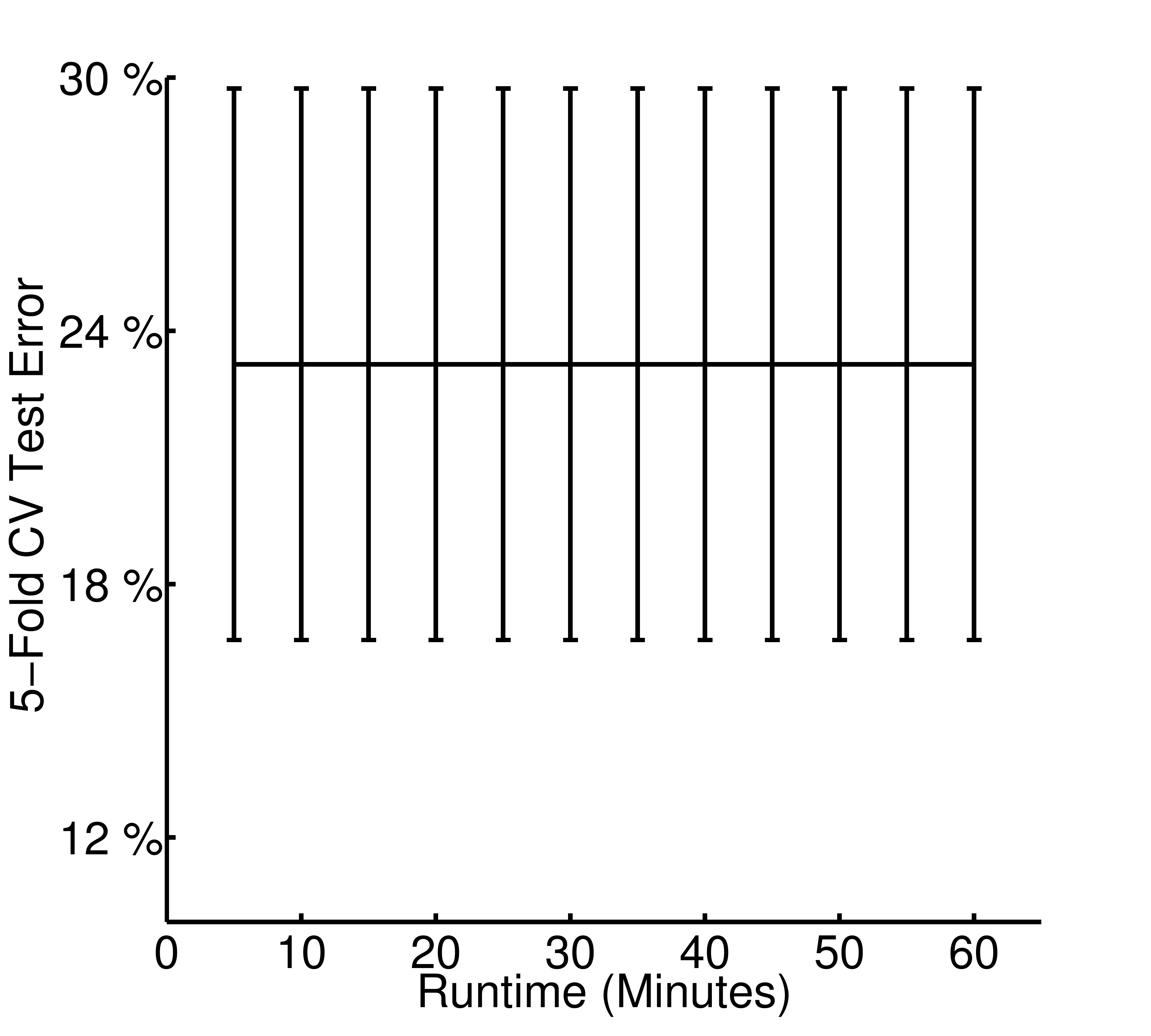}
    \label{S6_RUNTIME_TEST_ERRS_haberman}
\end{subfigure}
\hspace{0.1cm}
\begin{subfigure}[b]{0.30\textwidth}
    \centering
    \includegraphics[width=\textwidth,trim=5 20 5 50, clip]{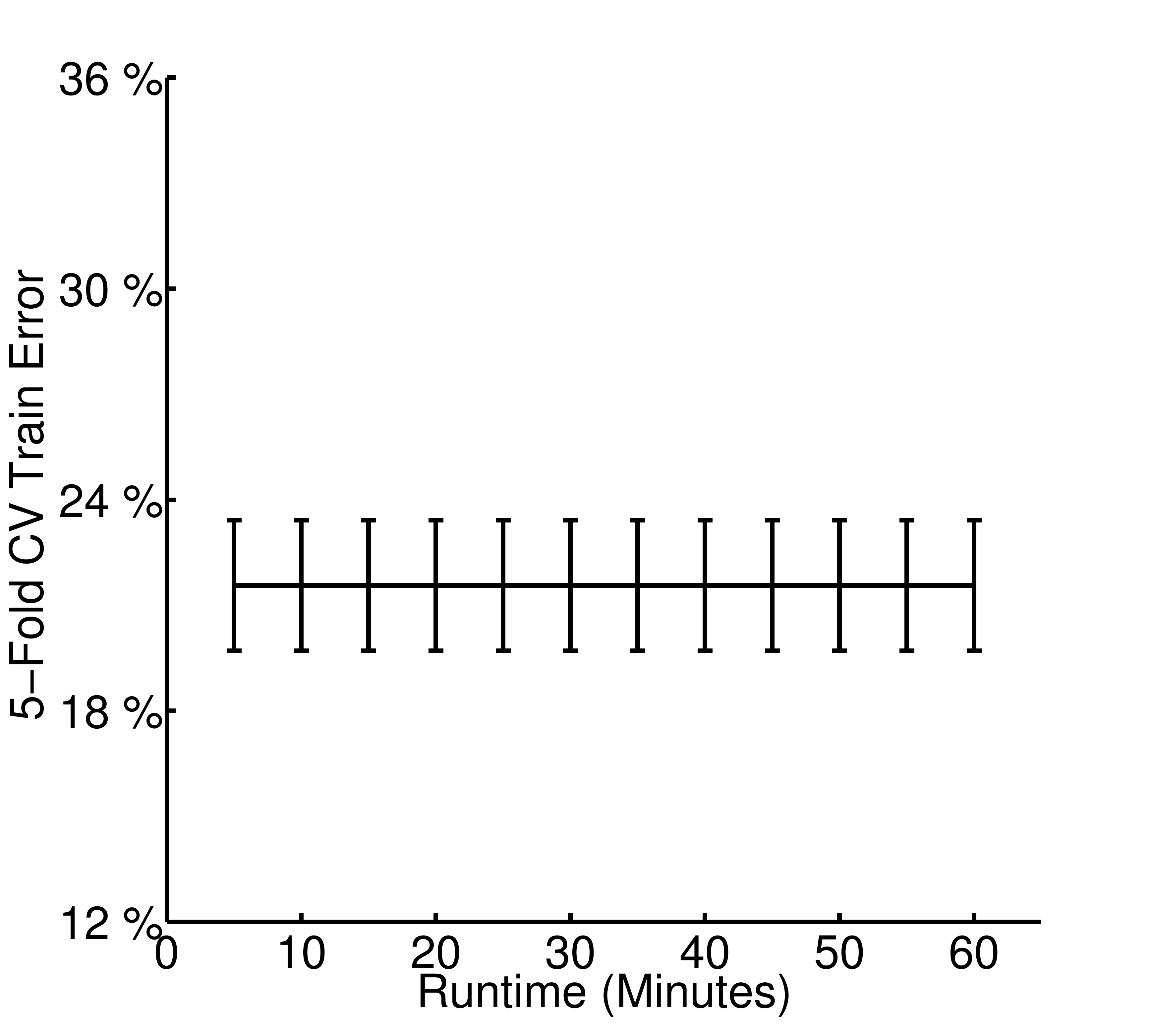}
    \label{S6_RUNTIME_TRAIN_ERRS_haberman}
\end{subfigure}

\begin{subfigure}[b]{0.30\textwidth}
    \centering
    \includegraphics[width=\textwidth,trim=5 20 5 50, clip]{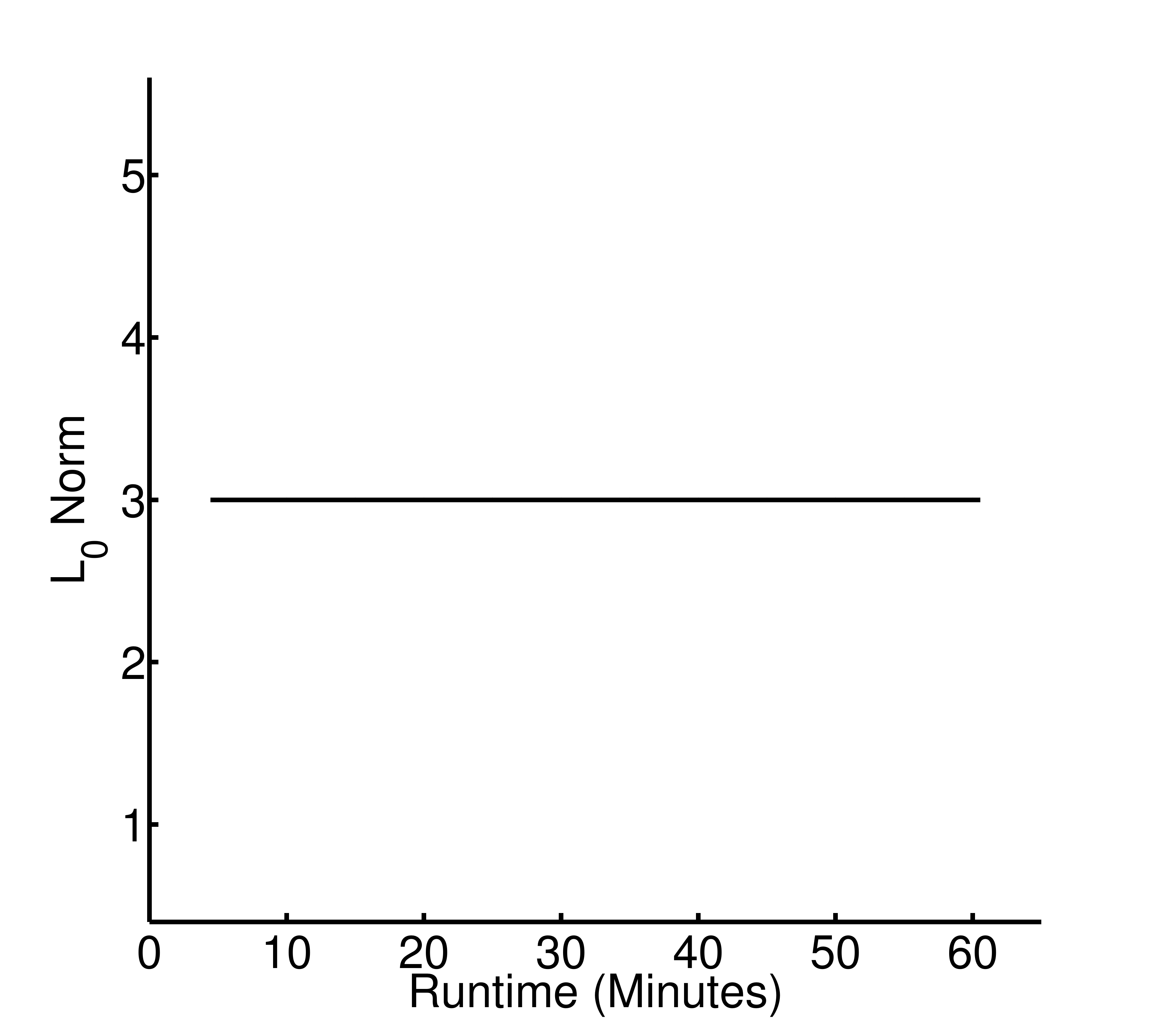}
    \label{S6_RUNTIME_LO_NORMS_haberman}
\end{subfigure}
\hspace{0.1cm}
\begin{subfigure}[b]{0.30\textwidth}
    \centering
    \includegraphics[width=\textwidth,trim=5 20 5 50, clip]{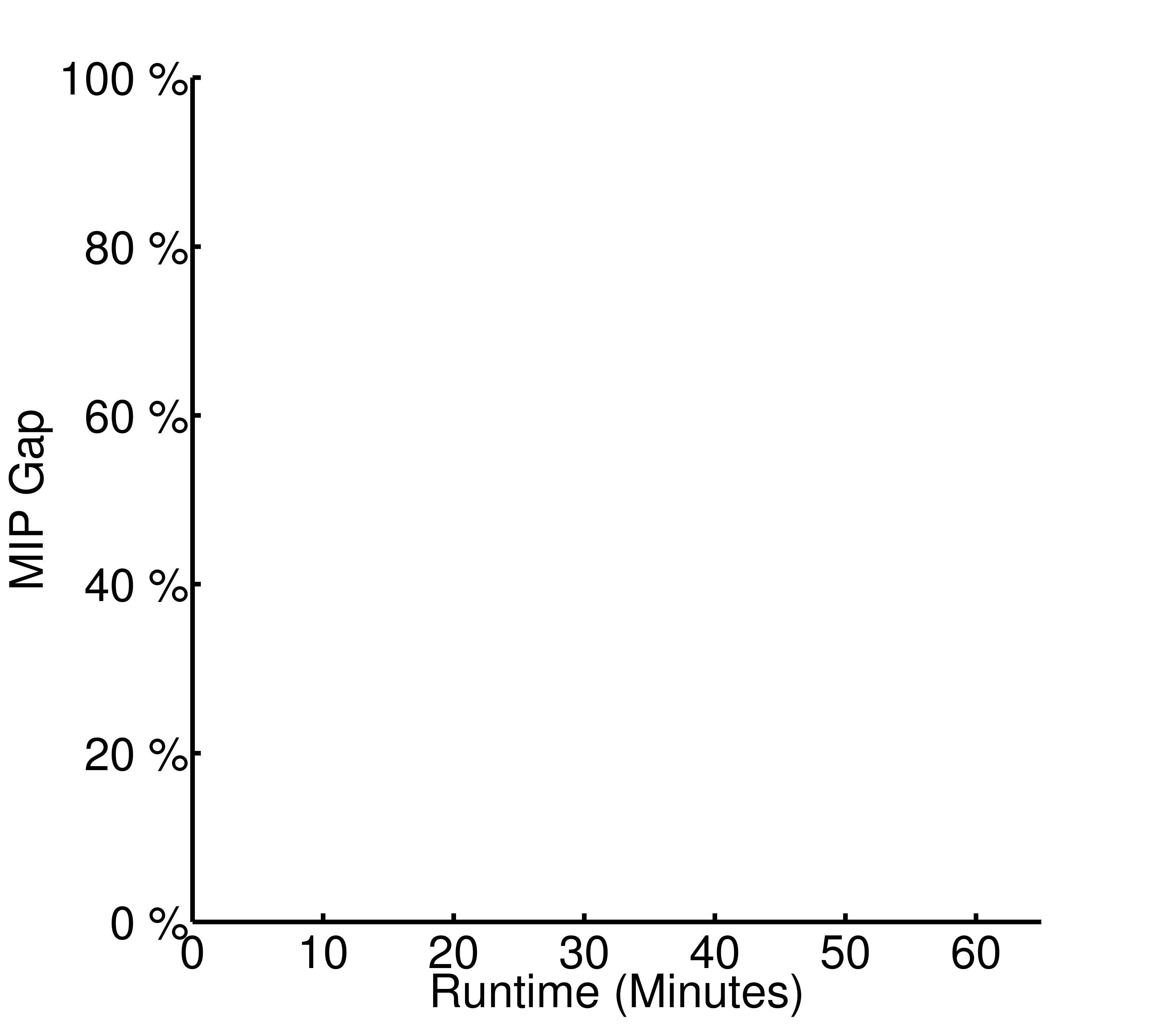}
    \label{S6_RUNTIME_MIPGAPS_haberman}
\end{subfigure}
\caption{Computational performance over time for \haberman.}
\label{Fig::CompPlots_haberman}
\end{figure}

\begin{figure}[htbp]
\centering
\begin{subfigure}[b]{0.30\textwidth}
    \centering
    \includegraphics[width=\textwidth,trim=5 20 5 50, clip]{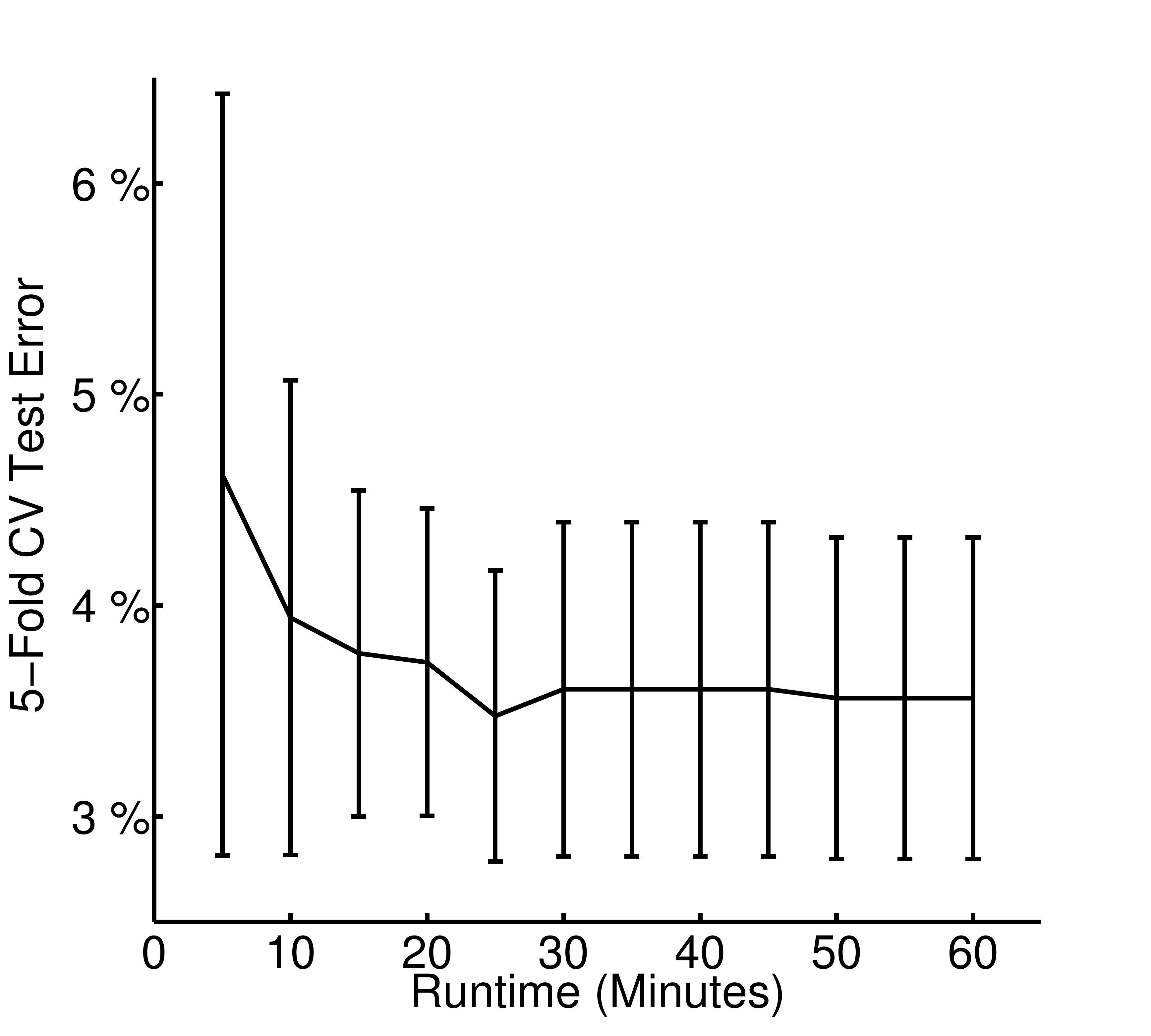}
    \label{S6_RUNTIME_TEST_ERRS_internetad}
\end{subfigure}
\hspace{0.1cm}
\begin{subfigure}[b]{0.30\textwidth}
    \centering
    \includegraphics[width=\textwidth,trim=5 20 5 50, clip]{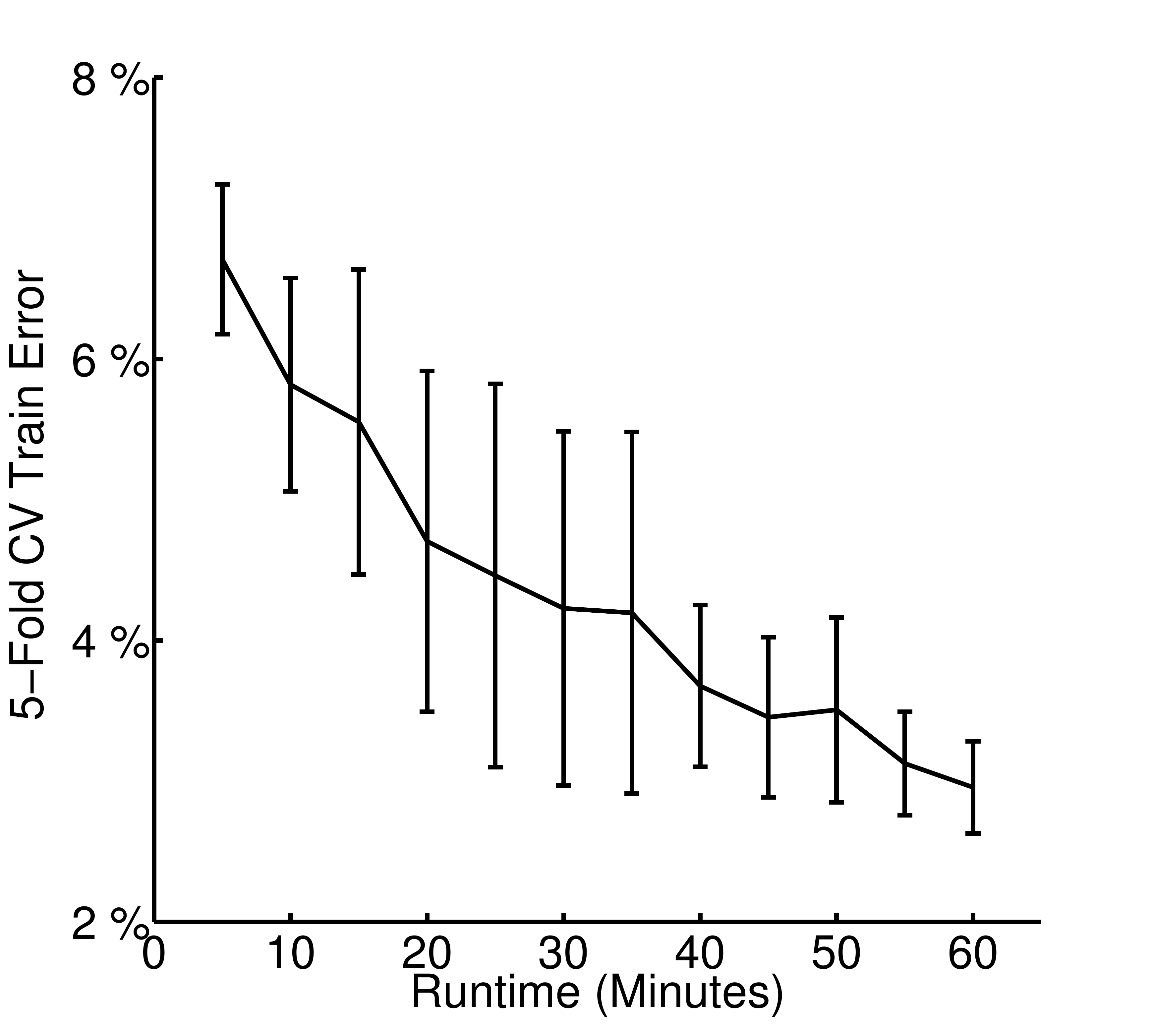}
    \label{S6_RUNTIME_TRAIN_ERRS_internetad}
\end{subfigure}

\begin{subfigure}[b]{0.30\textwidth}
    \centering
    \includegraphics[width=\textwidth,trim=5 20 5 50, clip]{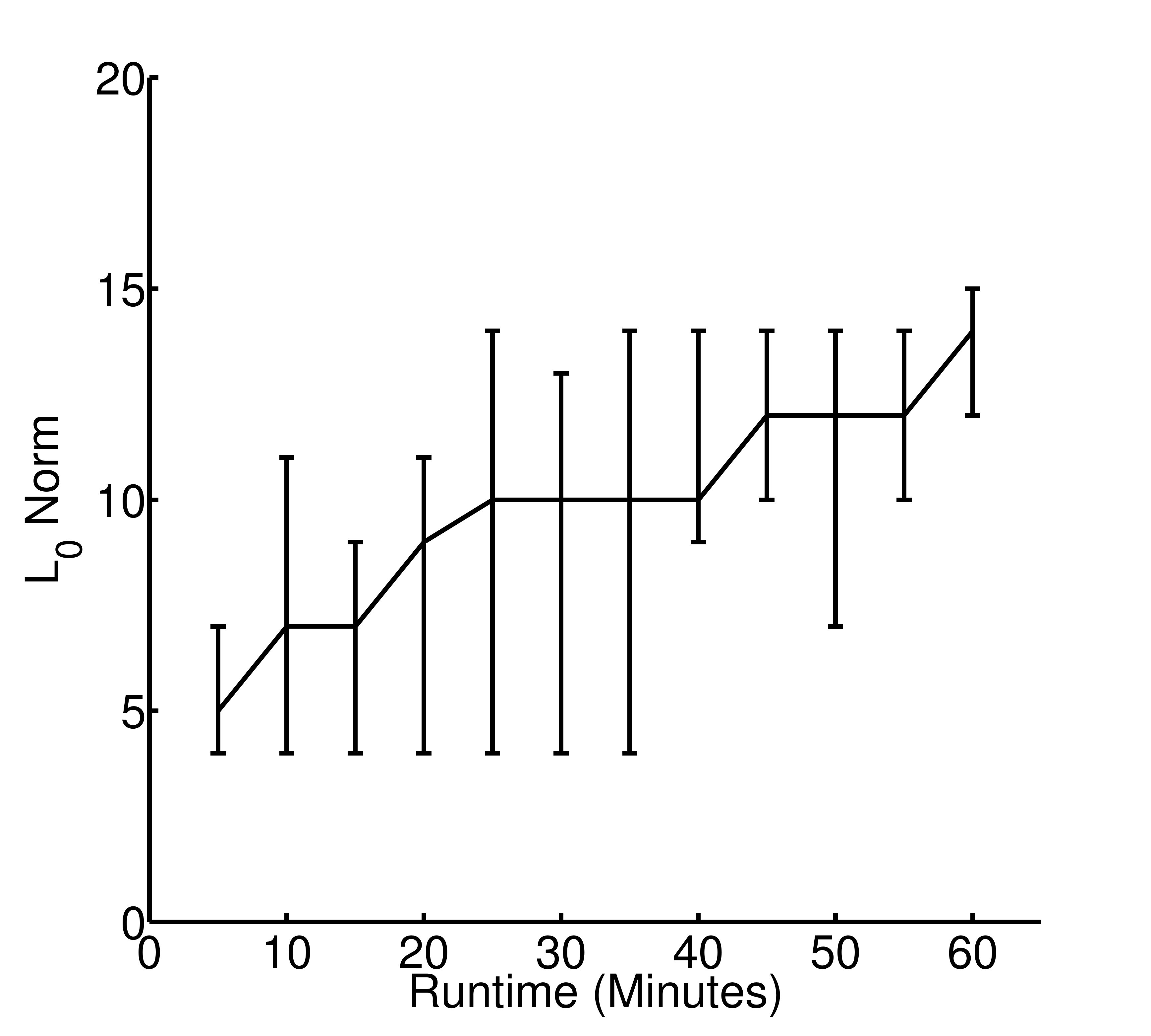}
    \label{S6_RUNTIME_LO_NORMS_internetad}
\end{subfigure}
\begin{subfigure}[b]{0.30\textwidth}
    \centering
    \includegraphics[width=\textwidth,trim=5 20 5 50, clip]{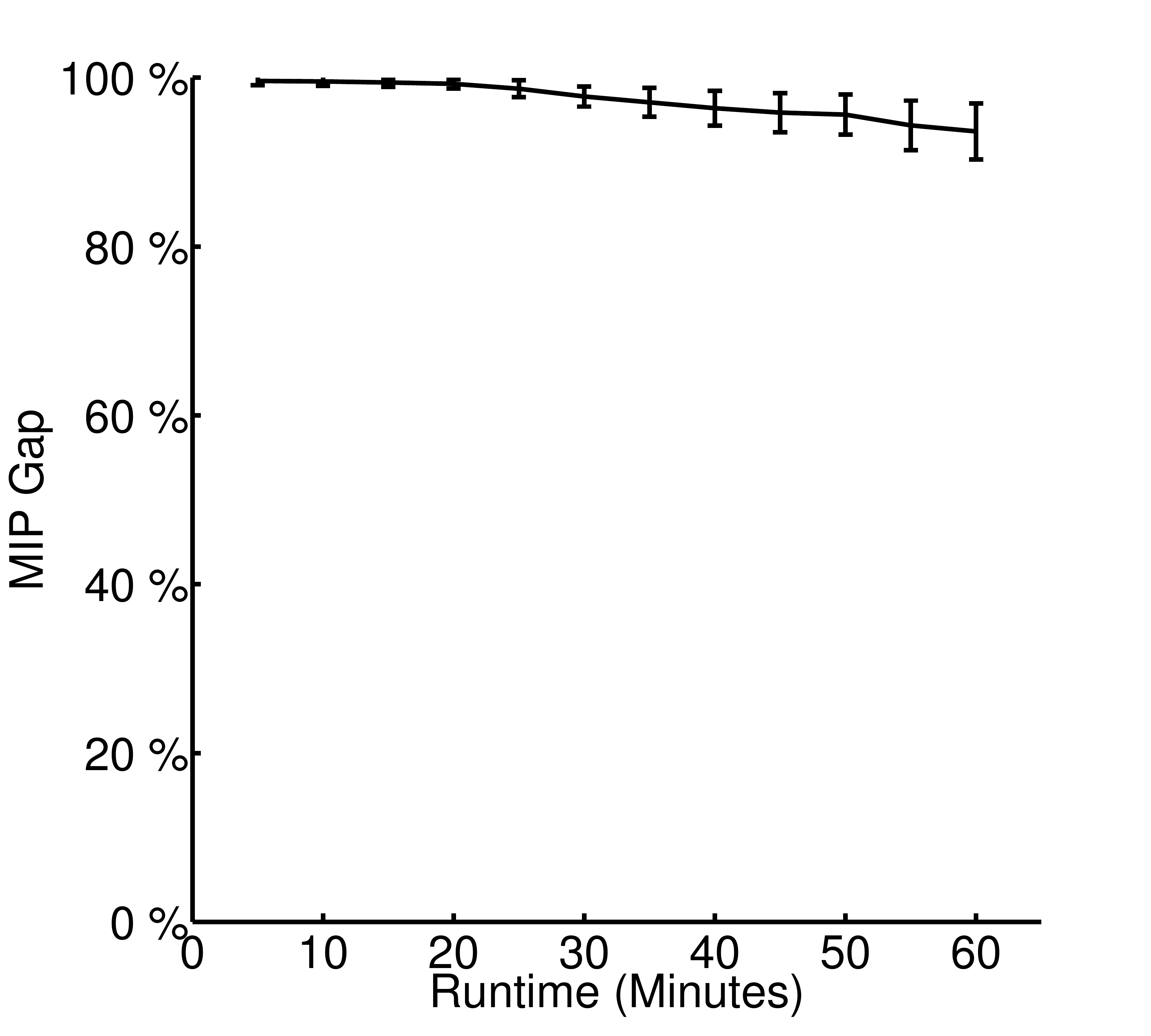}
    \label{S6_RUNTIME_MIPGAPS_internetad}
\end{subfigure}
\caption{Computational performance over time for \internetad.}
\label{Fig::CompPlots_internetad}
\end{figure}

\begin{figure}[htbp]
\centering
\begin{subfigure}[b]{0.30\textwidth}
    \centering
    \includegraphics[width=\textwidth,trim=5 20 5 50, clip]{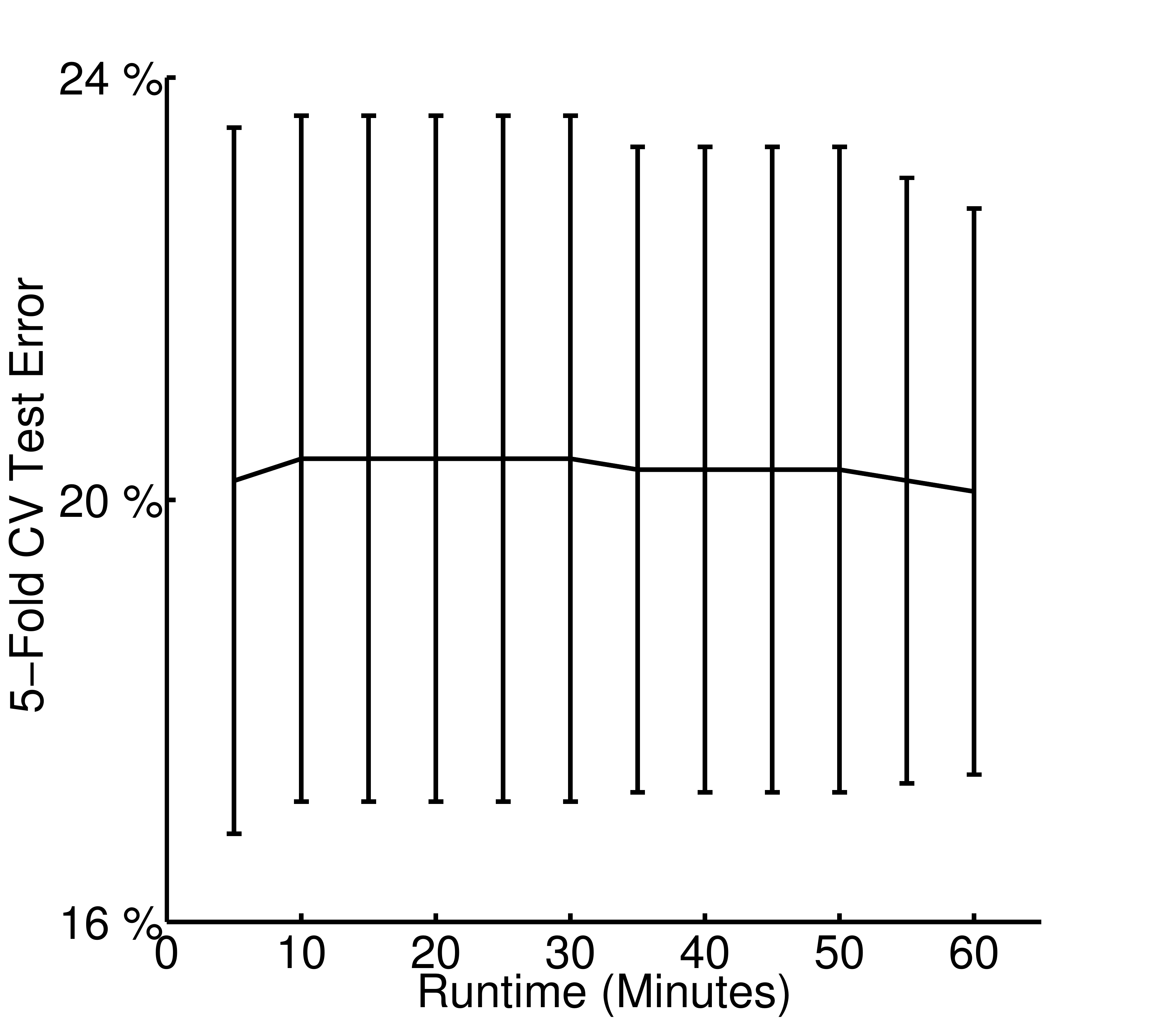}
    \label{S6_RUNTIME_TEST_ERRS_mammo}
\end{subfigure}
\hspace{0.1cm}
\begin{subfigure}[b]{0.30\textwidth}
    \centering
    \includegraphics[width=\textwidth,trim=5 20 5 50, clip]{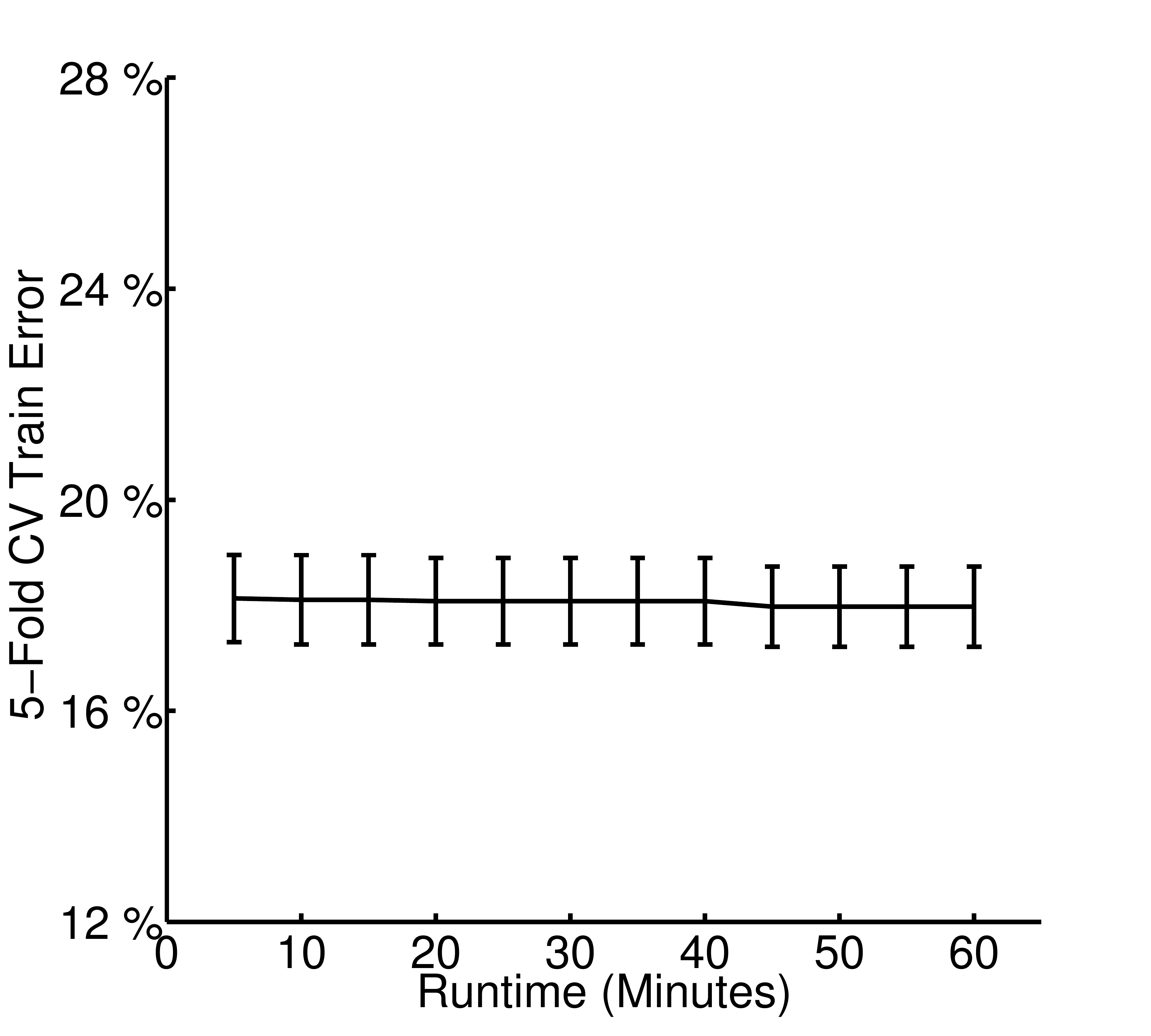}
    \label{S6_RUNTIME_TRAIN_ERRS_mammo}
\end{subfigure}

\begin{subfigure}[b]{0.30\textwidth}
    \centering
    \includegraphics[width=\textwidth,trim=5 20 5 50, clip]{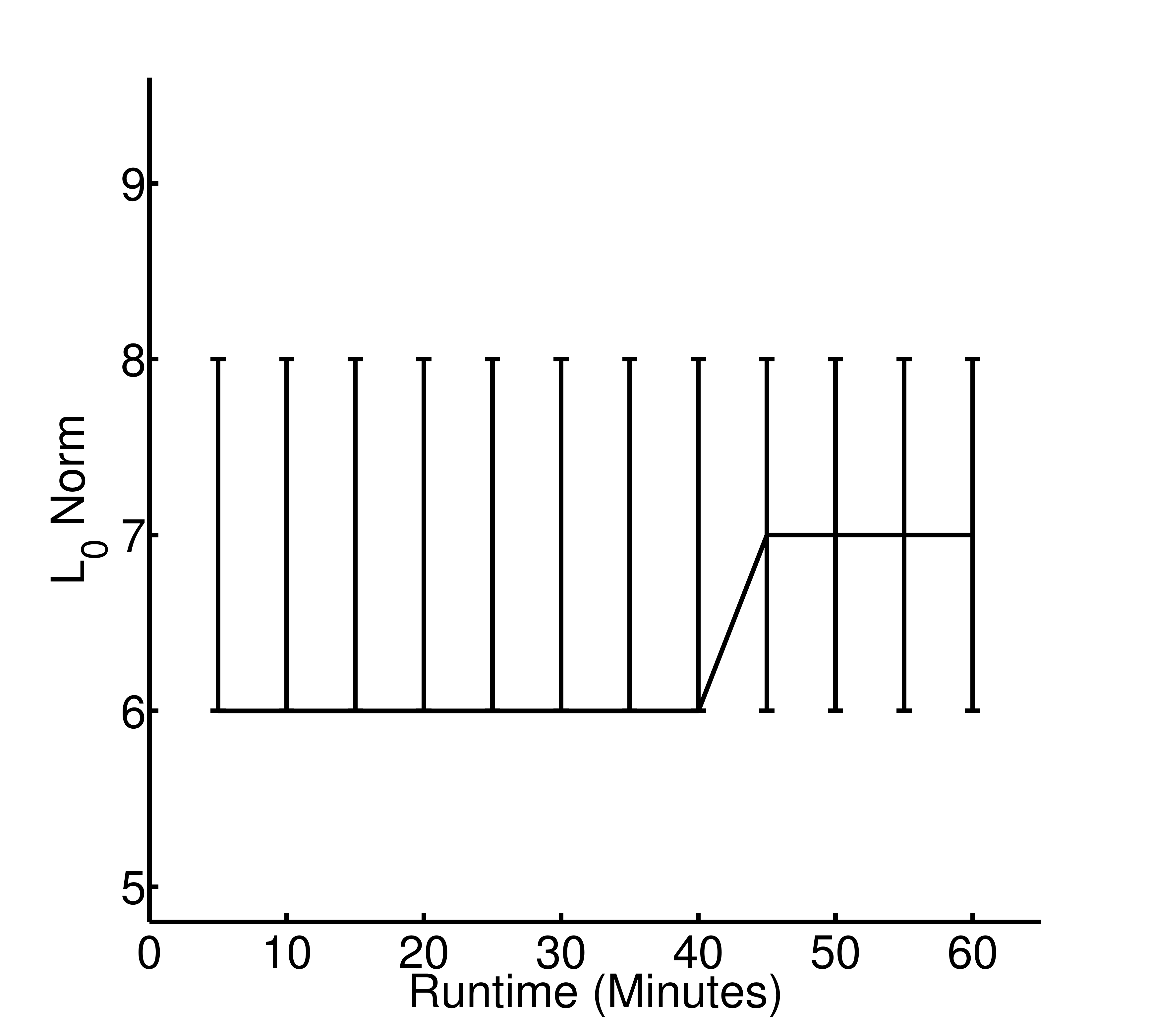}
    \label{S6_RUNTIME_LO_NORMS_mammo}
\end{subfigure}
\hspace{0.1cm}
\begin{subfigure}[b]{0.30\textwidth}
    \centering
    \includegraphics[width=\textwidth,trim=5 20 5 50, clip]{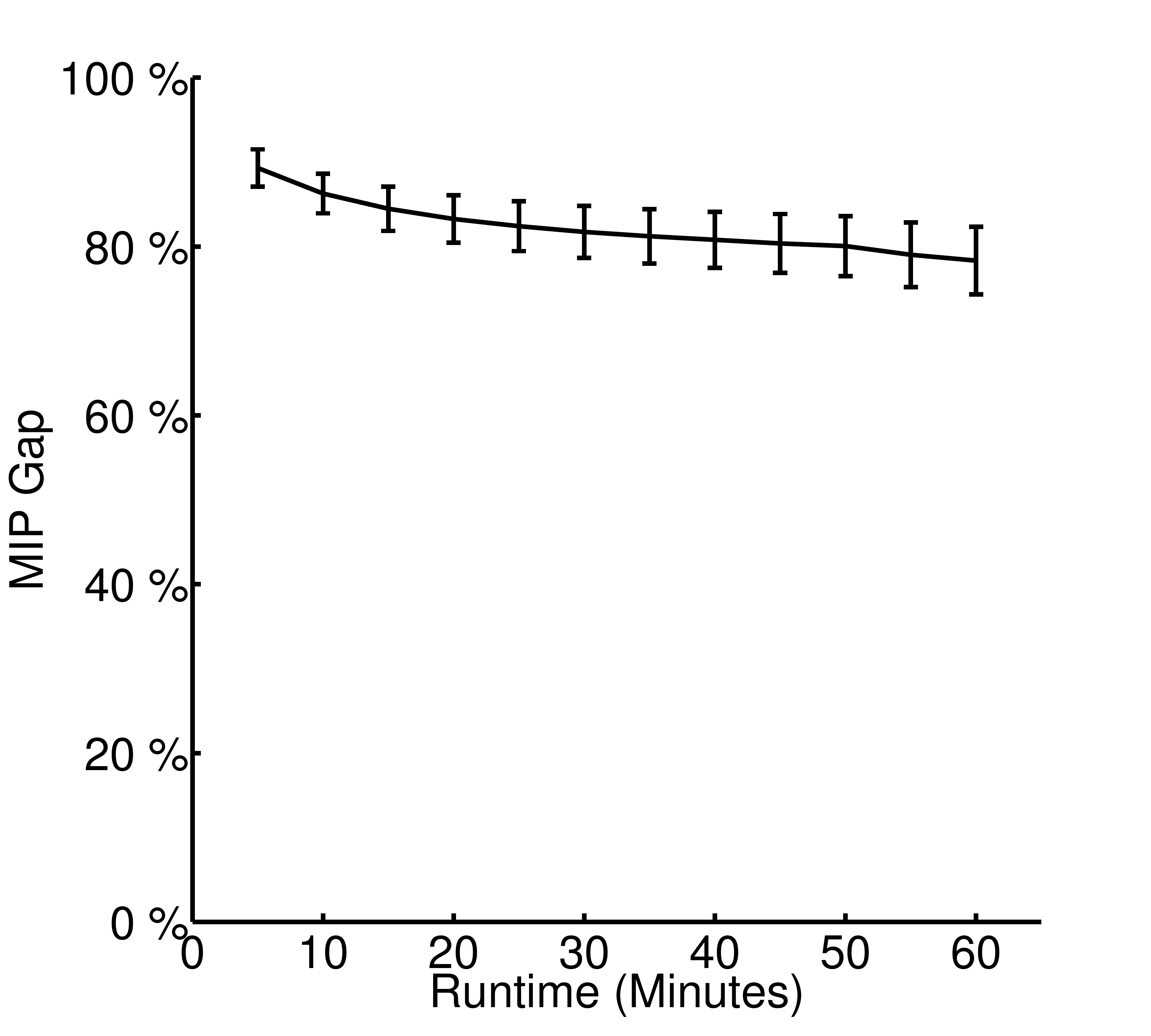}
    \label{S6_RUNTIME_MIPGAPS_mammo}
\end{subfigure}
\caption{Computational performance over time for \mammo.}
\label{Fig::CompPlots_mammo}
\end{figure}

\begin{figure}[htbp]
\centering
\begin{subfigure}[b]{0.30\textwidth}
    \centering
    \includegraphics[width=\textwidth,trim=5 20 5 50, clip]{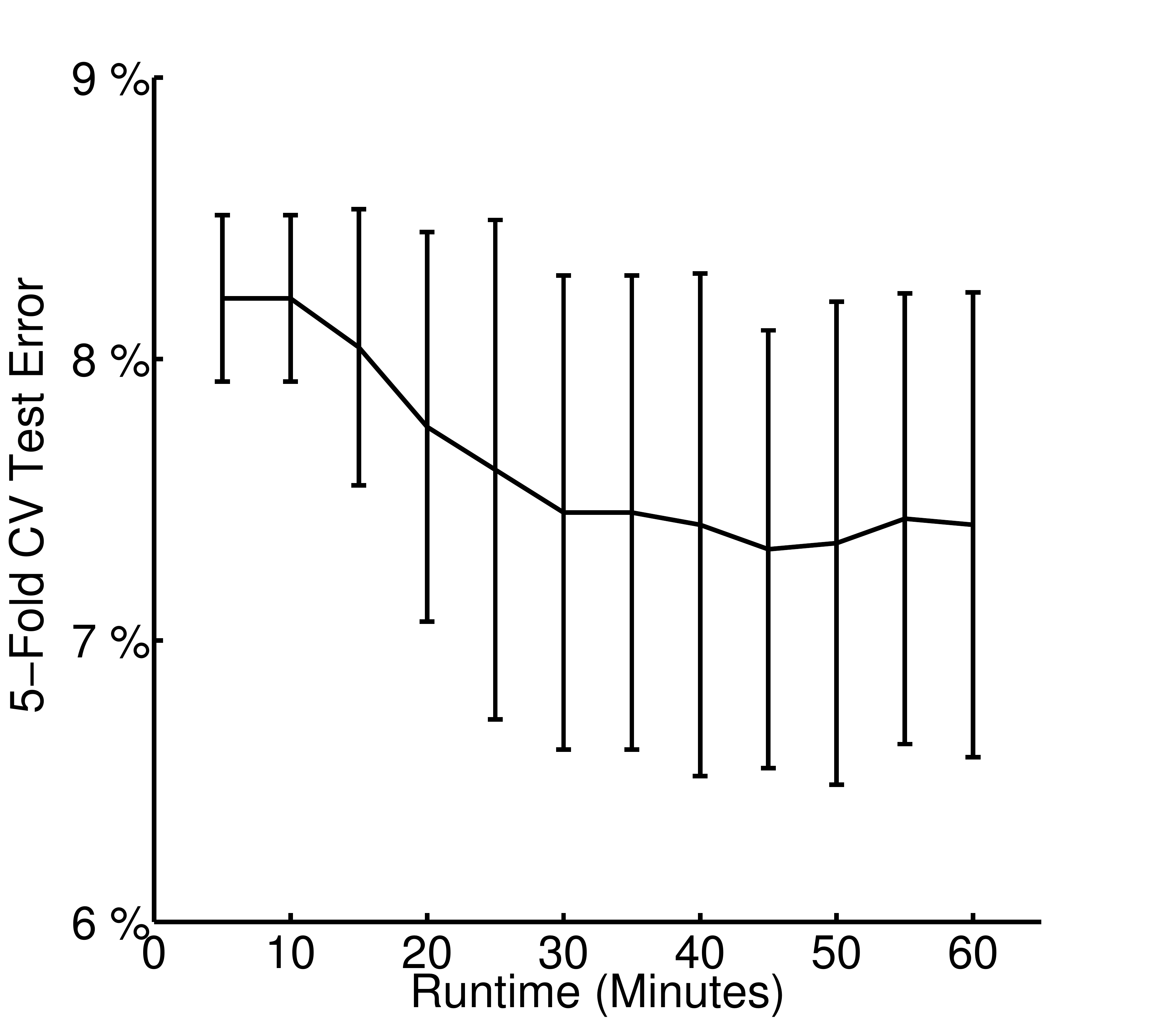}
    \label{S6_RUNTIME_TEST_ERRS_spambase}
\end{subfigure}
\hspace{0.1cm}
\begin{subfigure}[b]{0.30\textwidth}
    \centering
    \includegraphics[width=\textwidth,trim=5 20 5 50, clip]{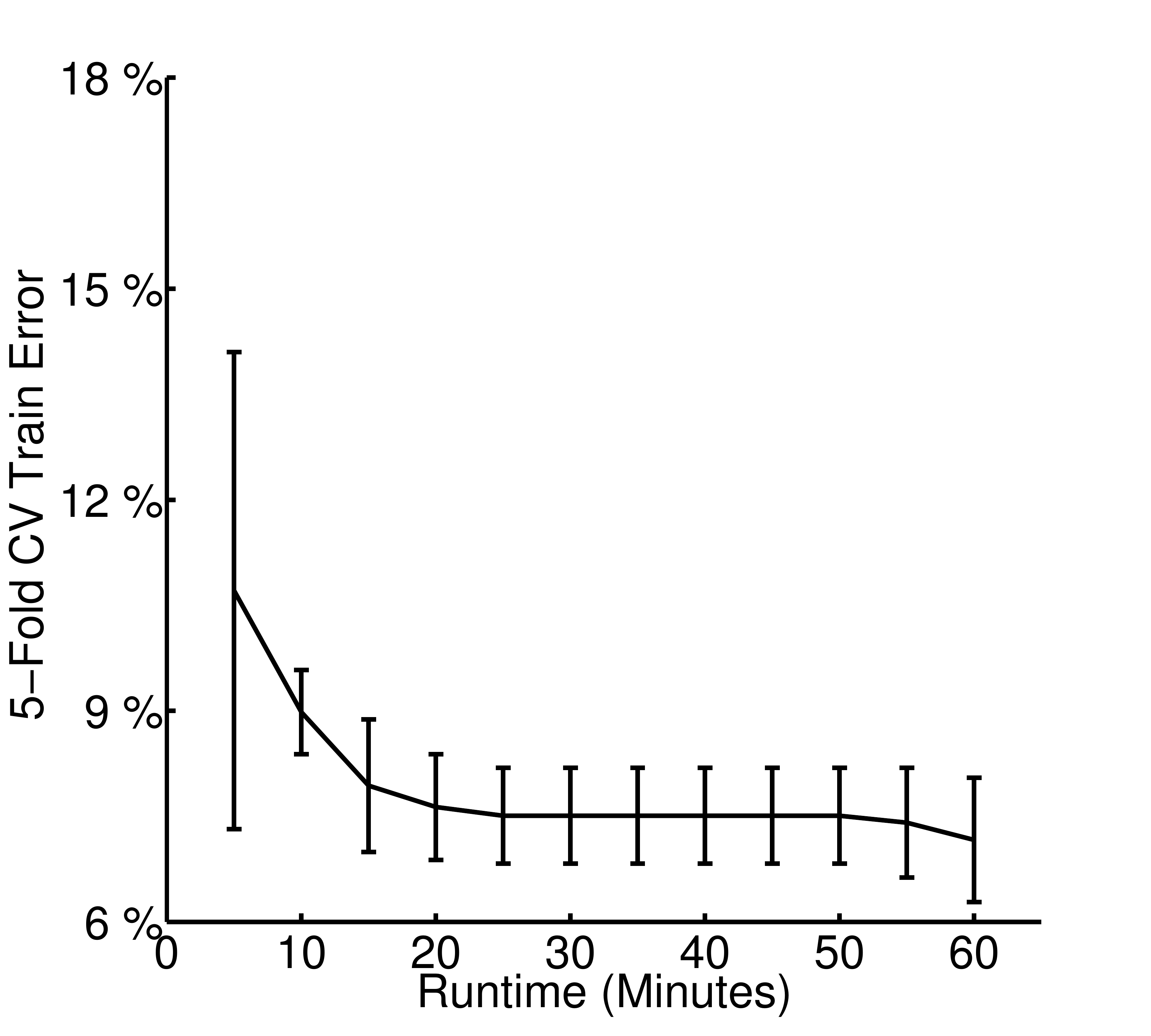}
    \label{S6_RUNTIME_TRAIN_ERRS_spambase}
\end{subfigure}

\begin{subfigure}[b]{0.30\textwidth}
    \centering
    \includegraphics[width=\textwidth,trim=5 20 5 50, clip]{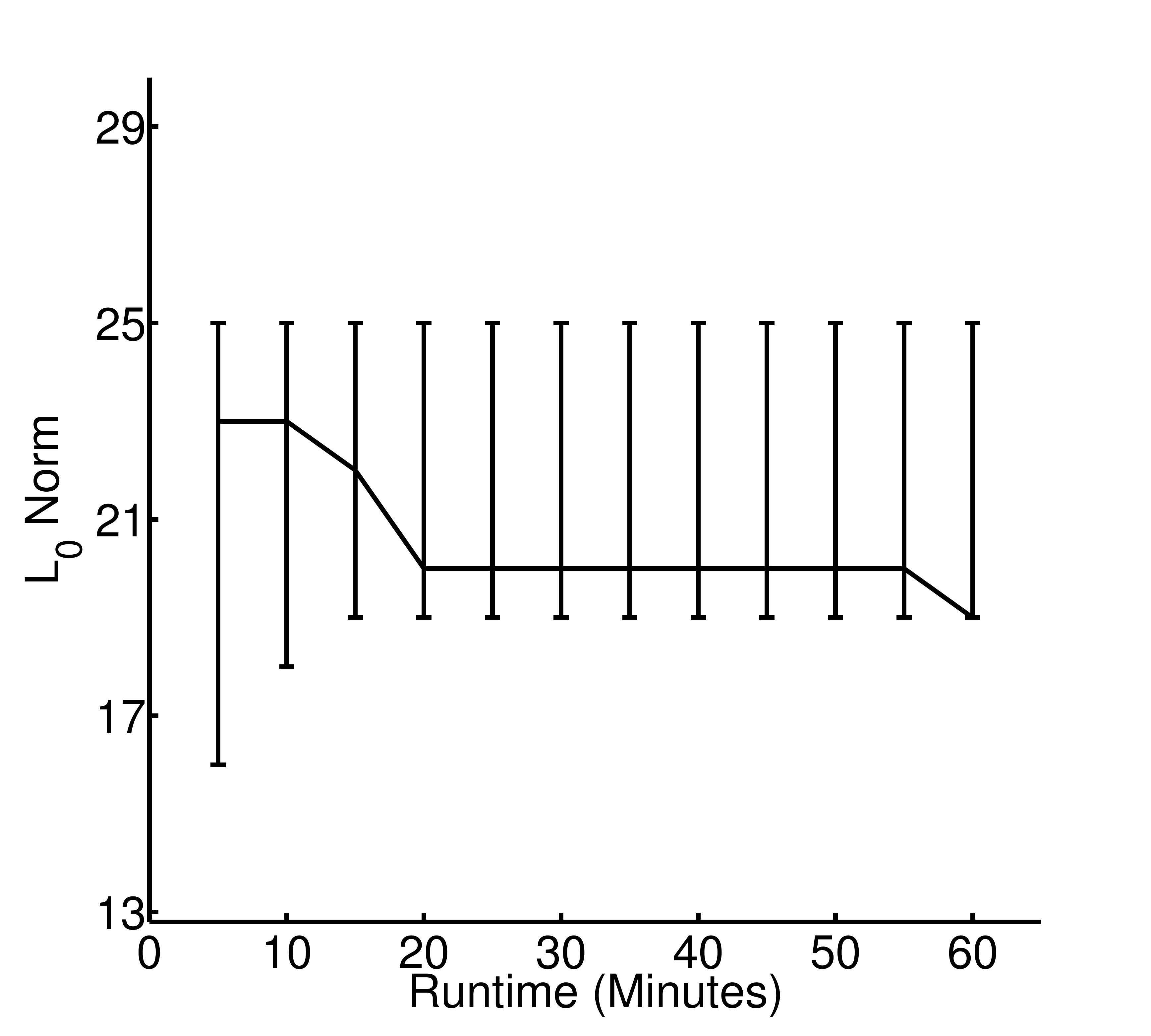}
    \label{S6_RUNTIME_LO_NORMS_spambase}
\end{subfigure}
\hspace{0.1cm}
\begin{subfigure}[b]{0.30\textwidth}
    \centering
    \includegraphics[width=\textwidth,trim=5 20 5 50, clip]{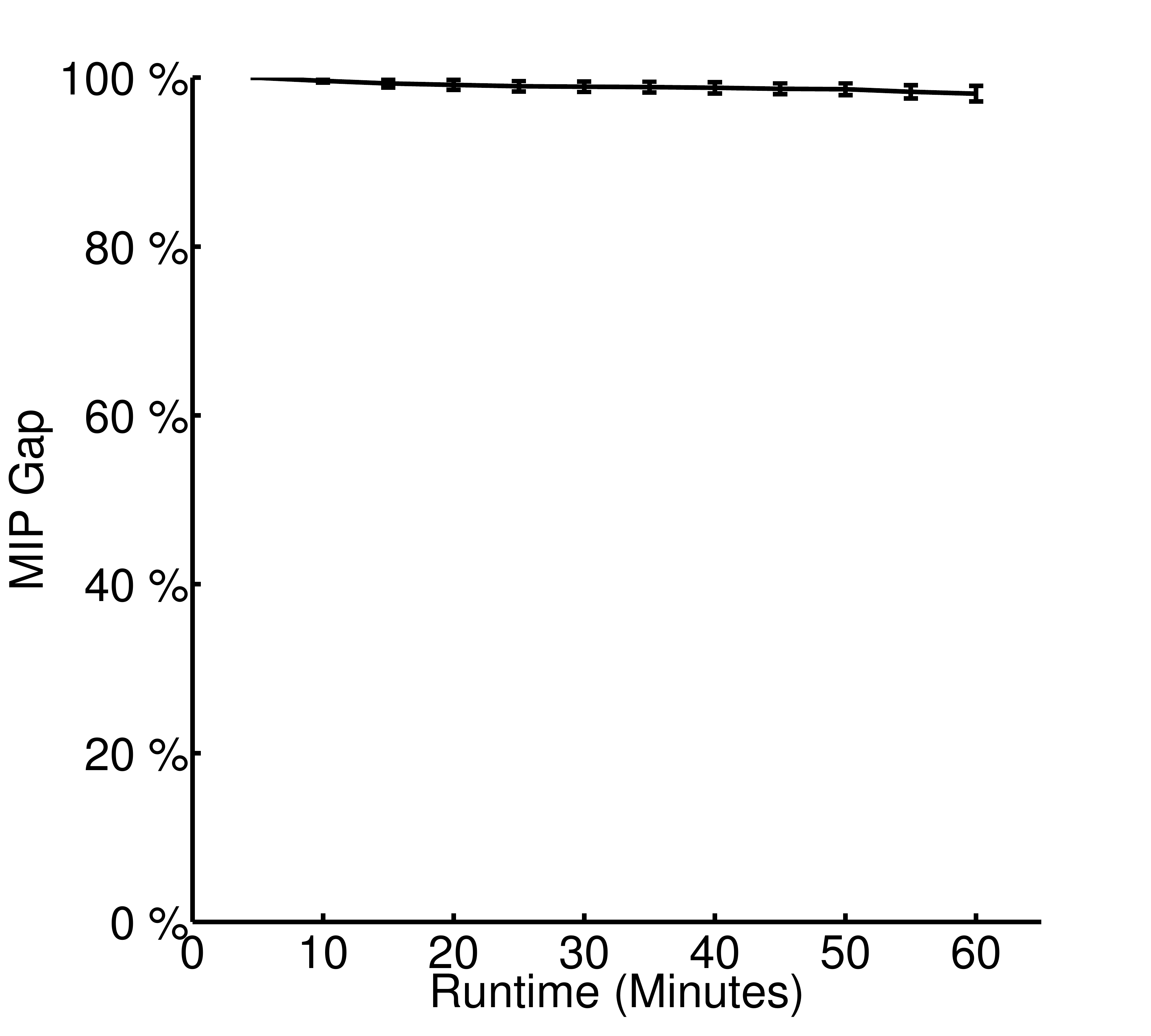}
    \label{S6_RUNTIME_MIPGAPS_spambase}
\end{subfigure}
\caption{Computational performance over time for \spambase.}
\label{Fig::CompPlots_spambase}
\end{figure}

\begin{figure}[htbp]
\centering
\begin{subfigure}[b]{0.30\textwidth}
    \centering
    \includegraphics[width=\textwidth,trim=5 20 5 50, clip]{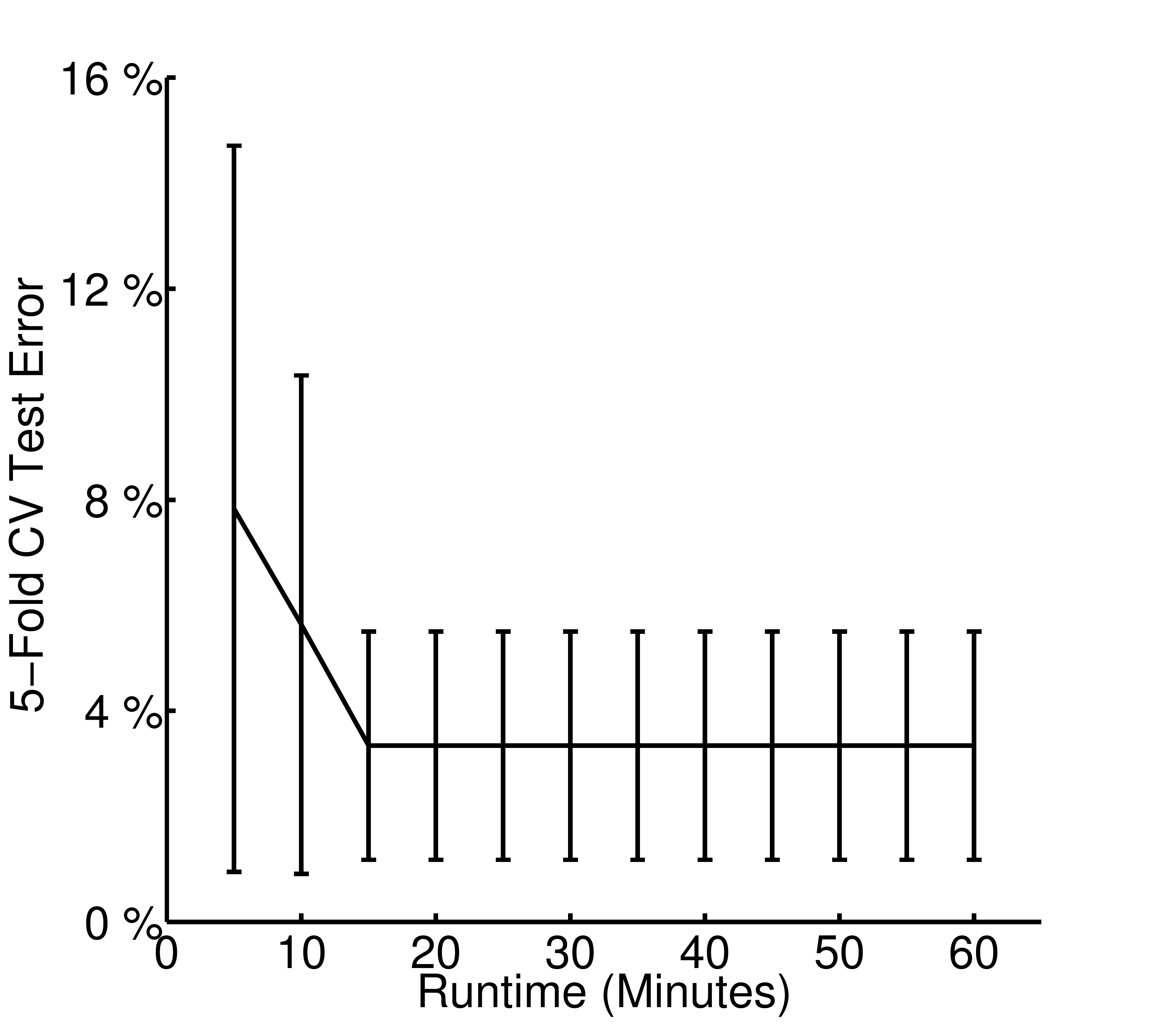}
    \label{S6_RUNTIME_TEST_ERRS_tictactoe}
\end{subfigure}
\hspace{0.1cm}
\begin{subfigure}[b]{0.30\textwidth}
    \centering
    \includegraphics[width=\textwidth,trim=5 20 5 50, clip]{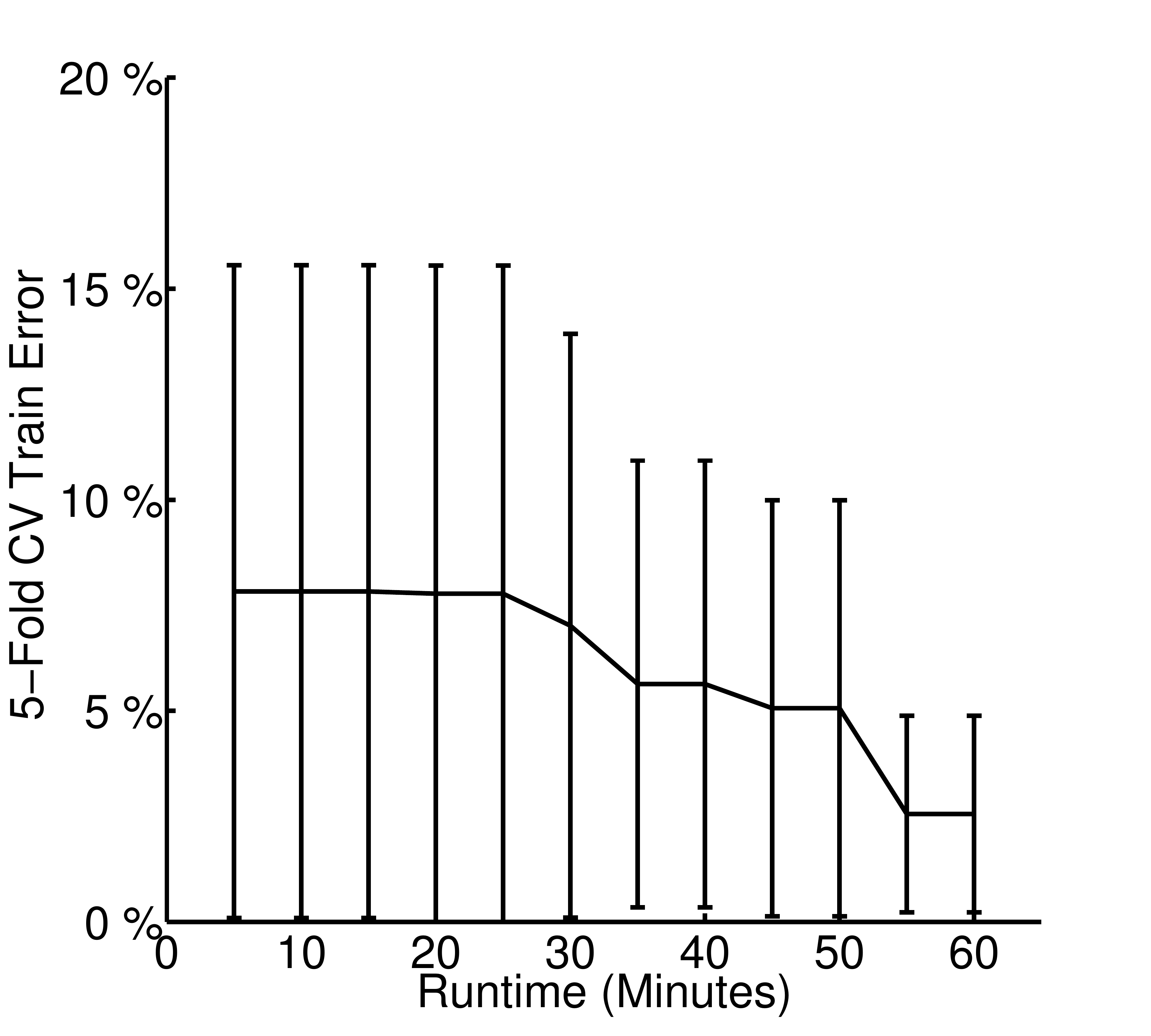}
    \label{S6_RUNTIME_TRAIN_ERRS_tictactoe}
\end{subfigure}

\begin{subfigure}[b]{0.30\textwidth}
    \centering
    \includegraphics[width=\textwidth,trim=5 20 5 50, clip]{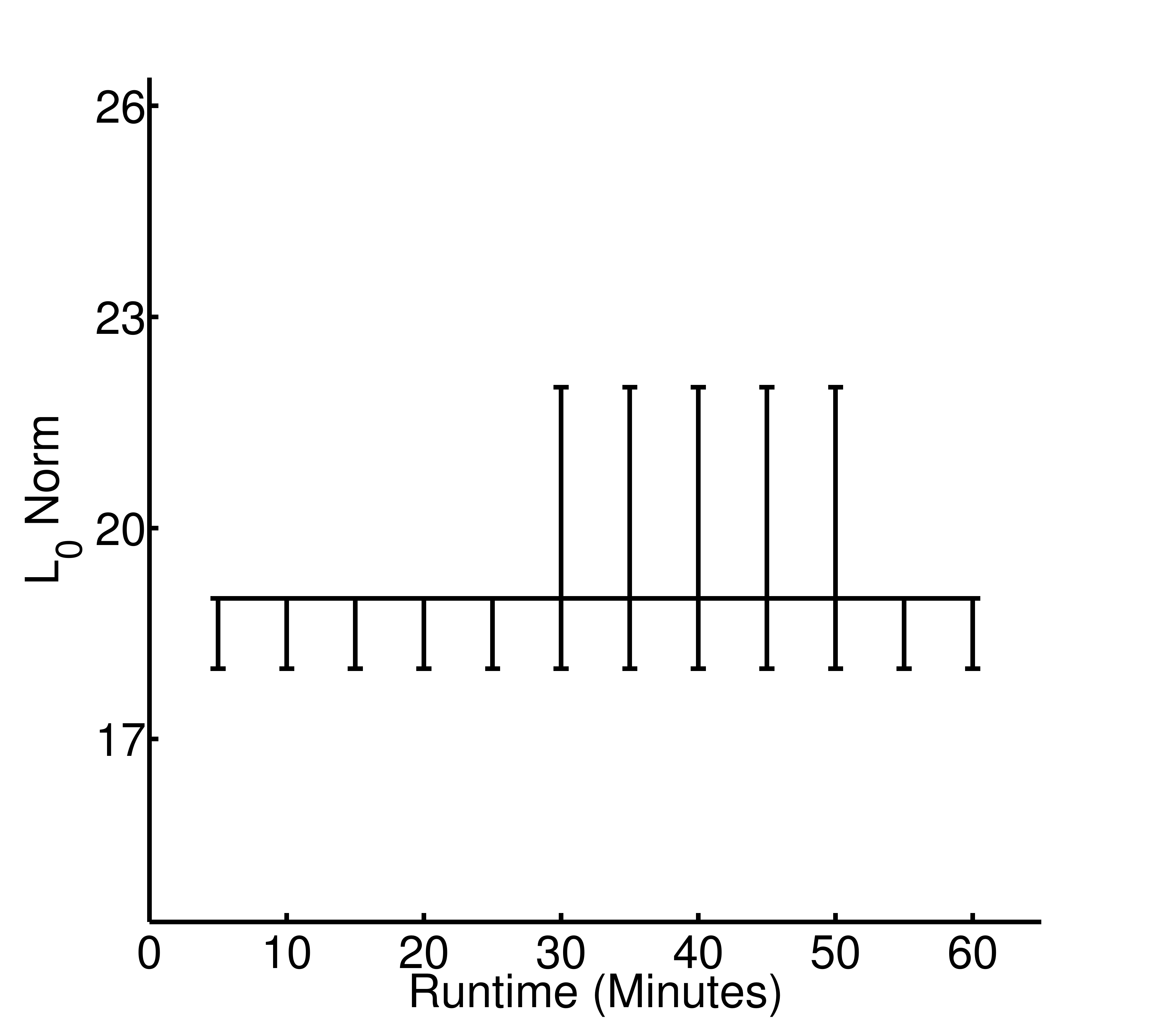}
    \label{S6_RUNTIME_LO_NORMS_tictactoe}
\end{subfigure}
\hspace{0.1cm}
\begin{subfigure}[b]{0.30\textwidth}
    \centering
    \includegraphics[width=\textwidth,trim=5 20 5 50, clip]{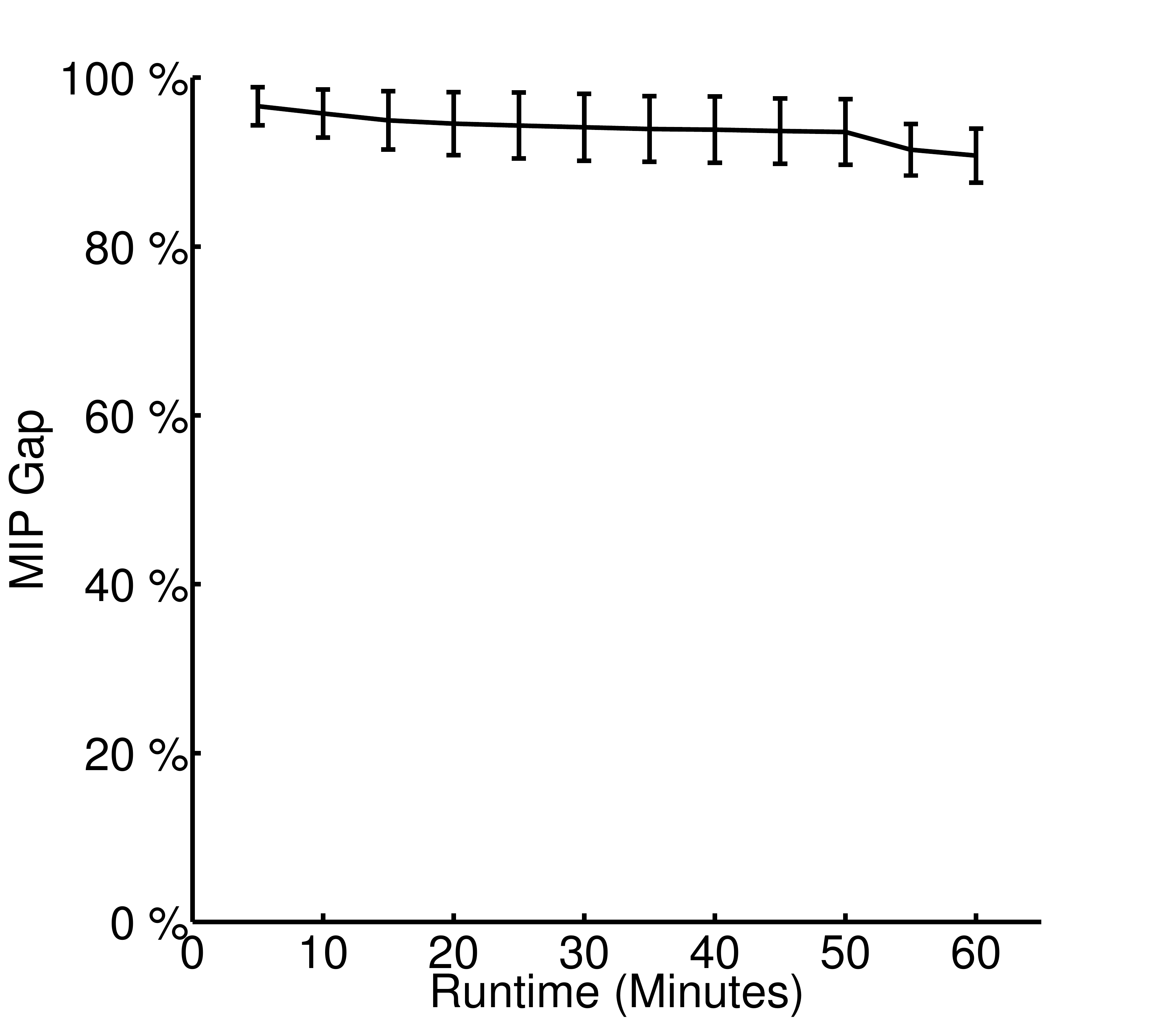}
    \label{S6_RUNTIME_MIPGAPS_tictactoe}
\end{subfigure}
\caption{Computational performance over time for \tictactoe.}
\label{Fig::CompPlots_tictactoe}
\end{figure}

Figures \ref{Fig::CompPlots_breastcancer} to \ref{Fig::CompPlots_tictactoe} illustrate the computational performance of SLIM on the datasets from Section \ref{Sec::NumericalExperiments} by showing how the scoring systems produced by the MIP formulation in Section \ref{Sec::SLIMBalancedModel} change with time. In particular, we track how the 5-fold CV test error, the 5-Fold CV training error, the $\ell_0$-norm and the MIP gap change over time. In many datasets, we can see that SLIM produces scoring systems whose key properties (i.e. the test error, training error and $\ell_0$-norm) stabilize over time. Even so, the MIP gap may remain large - especially for larger datasets such as \internetad\, and  \spambase\, (see Figures \ref{Fig::CompPlots_internetad}  and \ref{Fig::CompPlots_spambase}, respectively). This highlights the fact that current MIP solvers can often quickly find an optimal or near-optimal solution, but require a longer time to prove optimality. Note that when SLIM is used on small datasets, MIP solvers not only produce an optimal solution, but also provide a certificate of optimality; this is the case with the \haberman\, dataset (see Figure \ref{Fig::CompPlots_haberman}) where the MIP gap decreases to 0\% almost immediately.


\newpage
\clearpage


\bibliographystyle{plain}
\bibliography{SLIMforInterpretableClassification}   

\begin{thebibliography}{10}
\providecommand{\url}[1]{{#1}}
\providecommand{\urlprefix}{URL }
\expandafter\ifx\csname urlstyle\endcsname\relax
  \providecommand{\doi}[1]{DOI~\discretionary{}{}{}#1}\else
  \providecommand{\doi}{DOI~\discretionary{}{}{}\begingroup
  \urlstyle{rm}\Url}\fi

\bibitem{akaike1998information}
Akaike, H.: Information theory and an extension of the maximum likelihood
  principle.
\newblock In: Selected Papers of Hirotugu Akaike, pp. 199--213. Springer (1998)

\bibitem{andrade2009handbook}
Andrade, J.T.: Handbook of violence risk assessment and treatment: New
  approaches for mental health professionals.
\newblock Springer Publishing Company (2009)

\bibitem{antman2000timi}
Antman, E.M., Cohen, M., Bernink, P.J., McCabe, C.H., Horacek, T., Papuchis,
  G., Mautner, B., Corbalan, R., Radley, D., Braunwald, E.: The {TIMI} risk
  score for unstable angina/non--{ST} elevation {MI}.
\newblock The Journal of the American Medical Association \textbf{284}(7),
  835--842 (2000)

\bibitem{Bache+Lichman:2013}
Bache, K., Lichman, M.: {UCI} machine learning repository (2013).
\newblock \urlprefix\url{http://archive.ics.uci.edu/ml}

\bibitem{balakrishnan2008algorithms}
Balakrishnan, S., Madigan, D.: Algorithms for sparse linear classifiers in the
  massive data setting.
\newblock The Journal of Machine Learning Research \textbf{9}, 313--337 (2008)

\bibitem{bi2003dimensionality}
Bi, J., Bennett, K., Embrechts, M., Breneman, C., Song, M.: Dimensionality
  reduction via sparse support vector machines.
\newblock The Journal of Machine Learning Research \textbf{3}, 1229--1243
  (2003)

\bibitem{bien2011prototype}
Bien, J., Tibshirani, R.: Prototype selection for interpretable classification.
\newblock The Annals of Applied Statistics \textbf{5}(4), 2403--2424 (2011)

\bibitem{bone1992american}
Bone, R., Balk, R., Cerra, F., Dellinger, R., Fein, A., Knaus, W., Schein, R.,
  Sibbald, W., Abrams, J., Bernard, G., et~al.: American college of chest
  physicians/society of critical care medicine consensus conference:
  Definitions for sepsis and organ failure and guidelines for the use of
  innovative therapies in sepsis.
\newblock Critical Care Medicine \textbf{20}(6), 864--874 (1992)

\bibitem{bradley1999mathematical}
Bradley, P.S., Fayyad, U.M., Mangasarian, O.L.: Mathematical programming for
  data mining: formulations and challenges.
\newblock INFORMS Journal on Computing \textbf{11}(3), 217--238 (1999)

\bibitem{bratko1997machine}
Bratko, I.: Machine learning: Between accuracy and interpretability.
\newblock Courses and Lectures-International Centre for Mechanical Sciences pp.
  163--178 (1997)

\bibitem{breimanRF}
Breiman, L.: Random forests.
\newblock Mach. Learn. \textbf{45}(1), 5--32 (2001).
\newblock \doi{10.1023/A:1010933404324}.
\newblock \urlprefix\url{http://dx.doi.org/10.1023/A:1010933404324}

\bibitem{carrizosa2010binarized}
Carrizosa, E., Mart{\'\i}n-Barrag{\'a}n, B., Morales, D.R.: Binarized support
  vector machines.
\newblock INFORMS Journal on Computing \textbf{22}(1), 154--167 (2010)

\bibitem{carrizosa2011detecting}
Carrizosa, E., Mart{\'\i}n-Barrag{\'a}n, B., Morales, D.R.: Detecting relevant
  variables and interactions in supervised classification.
\newblock European Journal of Operational Research \textbf{213}(1), 260--269
  (2011)

\bibitem{carrizosaDILSVM13}
Carrizosa, E., Nogales-G\'omez, A., Romero~Morales, D.: Strongly agree or
  strongly disagree?: Rating features in support vector machines.
\newblock Tech. rep., Sa{\"\i}d Business School, University of Oxford, UK
  (2013)

\bibitem{carrizosa2013supervised}
Carrizosa, E., Romero~Morales, D.: Supervised classification and mathematical
  optimization.
\newblock Computers \& Operations Research \textbf{40}(1), 150--165 (2013)

\bibitem{abs2002marine}
Consulting, A.: Marine Safety: Tools for Risk-Based Decision Making.
\newblock Rowman \& Littlefield (2002)

\bibitem{cusick2010crime}
Cusick, G.R., Courtney, M.E., Havlicek, J., Hess, N.: Crime during the
  Transition to Adulthood: How Youth Fare as They Leave Out-of-Home Care.
\newblock National Institute of Justice, Office of Justice Programs, US
  Department of Justice (2010)

\bibitem{dawes1979robust}
Dawes, R.M.: The robust beauty of improper linear models in decision making.
\newblock American psychologist \textbf{34}(7), 571--582 (1979)

\bibitem{efron2004least}
Efron, B., Hastie, T., Johnstone, I., Tibshirani, R.: Least angle regression.
\newblock The Annals of Statistics \textbf{32}(2), 407--499 (2004)

\bibitem{elter2007prediction}
Elter, M., Schulz-Wendtland, R., Wittenberg, T.: The prediction of breast
  cancer biopsy outcomes using two cad approaches that both emphasize an
  intelligible decision process.
\newblock Medical Physics \textbf{34}, 4164 (2007)

\bibitem{business}
Flanigan, S., Morse, R.: Methodology: Best business schools rankings (2013).
\newblock U.S. News \& Wolrd Report

\bibitem{freund1997decision}
Freund, Y., Schapire, R.E.: A decision-theoretic generalization of on-line
  learning and an application to boosting.
\newblock Journal of computer and system sciences \textbf{55}(1), 119--139
  (1997)

\bibitem{friedman2010regularization}
Friedman, J., Hastie, T., Tibshirani, R.: Regularization paths for generalized
  linear models via coordinate descent.
\newblock Journal of statistical software \textbf{33}(1), 1 (2010)

\bibitem{friedman2010glmnet}
Friedman, J., Hastie, T., Tibshirani, R.: Regularization paths for generalized
  linear models via coordinate descent.
\newblock Journal of Statistical Software \textbf{33}(1), 1--22 (2010).
\newblock \urlprefix\url{http://www.jstatsoft.org/v33/i01/}

\bibitem{gage2001validation}
Gage, B.F., Waterman, A.D., Shannon, W., Boechler, M., Rich, M.W., Radford,
  M.J.: Validation of clinical classification schemes for predicting stroke.
\newblock The journal of the American Medical Association \textbf{285}(22),
  2864--2870 (2001)

\bibitem{giacobello2012sparse}
Giacobello, D., Christensen, M.G., Murthi, M.N., Jensen, S.H., Moonen, M.:
  Sparse linear prediction and its applications to speech processing.
\newblock IEEE Transactions on Audio, Speech, and Language Processing
  \textbf{20}(5), 1644--1657 (2012)

\bibitem{GoldbergEc2012}
Goldberg, N., Eckstein, J.: Sparse weighted voting classifier selection and its
  linear programming relaxations.
\newblock Information Processing Letters \textbf{112}, 481--486 (2012)

\bibitem{greenshtein2006best}
Greenshtein, E.: Best subset selection, persistence in high-dimensional
  statistical learning and optimization under l1 constraint.
\newblock The Annals of Statistics \textbf{34}(5), 2367--2386 (2006)

\bibitem{guyon2003introduction}
Guyon, I., Elisseeff, A.: An introduction to variable and feature selection.
\newblock The Journal of Machine Learning Research \textbf{3}, 1157--1182
  (2003)

\bibitem{hastie2005entire}
Hastie, T., Rosset, S., Tibshirani, R., Zhu, J.: The entire regularization path
  for the support vector machine.
\newblock Journal of Machine Learning Research \textbf{5}(2), 1391 (2005)

\bibitem{hesterberg2008least}
Hesterberg, T., Choi, N.H., Meier, L., Fraley, C.: Least angle and
  Ã¢ÂÂ1 penalized regression: A review.
\newblock Statistics Surveys \textbf{2}, 61--93 (2008)

\bibitem{Holte93}
Holte, R.C.: Very simple classification rules perform well on most commonly
  used datasets.
\newblock Machine Learning \textbf{11}(1), 63--91 (1993)

\bibitem{jennings1982informal}
Jennings, D., Amabile, T., Ross, L.: Informal covariation assessment:
  Data-based vs. theory-based judgments.
\newblock Judgment under uncertainty: Heuristics and biases pp. 211--230 (1982)

\bibitem{knaus1985apache}
Knaus, W.A., Draper, E.A., Wagner, D.P., Zimmerman, J.E.: {APACHE II}: a
  severity of disease classification system.
\newblock Critical Care Medicine \textbf{13}(10), 818--829 (1985)

\bibitem{knaus1991apache}
Knaus, W.A., Wagner, D., Draper, E., Zimmerman, J., Bergner, M., Bastos, P.,
  Sirio, C., Murphy, D., Lotring, T., Damiano, A.: The {APACHE III} prognostic
  system. risk prediction of hospital mortality for critically ill hospitalized
  adults.
\newblock Chest Journal \textbf{100}(6), 1619--1636 (1991)

\bibitem{knaus1981apache}
Knaus, W.A., Zimmerman, J.E., Wagner, D.P., Draper, E.A., Lawrence, D.E.:
  {APACHE}-acute physiology and chronic health evaluation: a physiologically
  based classification system.
\newblock Critical Care Medicine \textbf{9}(8), 591--597 (1981)

\bibitem{kohavi1997wrappers}
Kohavi, R., John, G.H.: Wrappers for feature subset selection.
\newblock Artificial intelligence \textbf{97}(1), 273--324 (1997)

\bibitem{kuhn2012c50}
Kuhn, M., Weston, S., code for C5.0~by R.~Quinlan, N.C.C.: C50: C5.0 Decision
  Trees and Rule-Based Models (2012).
\newblock \urlprefix\url{http://CRAN.R-project.org/package=C50}.
\newblock R package version 0.1.0-013

\bibitem{le1993new}
Le~Gall, J.R., Lemeshow, S., Saulnier, F.: A new simplified acute physiology
  score ({SAPS} {II}) based on a european/north american multicenter study.
\newblock The Journal of the American Medical Association \textbf{270}(24),
  2957--2963 (1993)

\bibitem{le1984simplified}
Le~Gall, J.R., Loirat, P., Alperovitch, A., Glaser, P., Granthil, C., Mathieu,
  D., Mercier, P., Thomas, R., Villers, D.: A simplified acute physiology score
  for icu patients.
\newblock Critical Care Medicine \textbf{12}(11), 975--977 (1984)

\bibitem{LethamRuMcMa13}
Letham, B., Rudin, C., McCormick, T.H., Madigan, D.: An interpretable stroke
  prediction model using rules and bayesian analysis.
\newblock In: Proceedings of AAAI Late Breaking Track (2013)

\bibitem{light1972pleural}
Light, R.W., Macgregor, M.I., Luchsinger, P.C., Ball, W.C.: Pleural effusions:
  the diagnostic separation of transudates and exudates.
\newblock Annals of Internal Medicine \textbf{77}(4), 507--513 (1972)

\bibitem{Lip10}
Lip, G., Nieuwlaat, R., Pisters, R., Lane, D., Crijns, H.: Refining clinical
  risk stratification for predicting stroke and thromboembolism in atrial
  fibrillation using a novel risk factor-based approach: the euro heart survey
  on atrial fibrillation.
\newblock Chest \textbf{137}, 263--272 (2010)

\bibitem{liu2009estimation}
Liu, H., Zhang, J.: Estimation consistency of the group lasso and its
  applications.
\newblock In: Proceedings of the Twelfth International Conference on Artificial
  Intelligence and Statistics (2009)

\bibitem{MangaWo90}
Mangasarian, O.L., Wolberg, W.H.: Cancer diagnosis via linear programming.
\newblock {SIAM} News \textbf{23}(5), 1,18 (1990)

\bibitem{mao2002fast}
Mao, K.: Fast orthogonal forward selection algorithm for feature subset
  selection.
\newblock IEEE Transactions on Neural Networks \textbf{13}(5), 1218--1224
  (2002)

\bibitem{mao2004orthogonal}
Mao, K.: Orthogonal forward selection and backward elimination algorithms for
  feature subset selection.
\newblock {IEEE} Transactions on Systems, Man, and Cybernetics, Part B:
  Cybernetics \textbf{34}(1), 629--634 (2004)

\bibitem{mateos2010distributed}
Mateos, G., Bazerque, J.A., Giannakis, G.B.: Distributed sparse linear
  regression.
\newblock IEEE Transactions on Signal Processing \textbf{58}(10), 5262--5276
  (2010)

\bibitem{metnitz2005saps}
Metnitz, P.G., Moreno, R.P., Almeida, E., Jordan, B., Bauer, P., Campos, R.A.,
  Iapichino, G., Edbrooke, D., Capuzzo, M., Le~Gall, J.R.: {SAPS 3} - from
  evaluation of the patient to evaluation of the intensive care unit. part 1:
  Objectives, methods and cohort description.
\newblock Intensive Care Medicine \textbf{31}(10), 1336--1344 (2005)

\bibitem{meyer2012e1071}
Meyer, D., Dimitriadou, E., Hornik, K., Weingessel, A., Leisch, F.: e1071: Misc
  Functions of the Department of Statistics (e1071), TU Wien (2012).
\newblock \urlprefix\url{http://CRAN.R-project.org/package=e1071}.
\newblock R package version 1.6-1

\bibitem{miller1984selection}
Miller, A.J.: Selection of subsets of regression variables.
\newblock Journal of the Royal Statistical Society. Series A (General) pp.
  389--425 (1984)

\bibitem{moreno2005saps}
Moreno, R.P., Metnitz, P.G., Almeida, E., Jordan, B., Bauer, P., Campos, R.A.,
  Iapichino, G., Edbrooke, D., Capuzzo, M., Le~Gall, J.R.: {SAPS 3} - from
  evaluation of the patient to evaluation of the intensive care unit. part 2:
  Development of a prognostic model for hospital mortality at icu admission.
\newblock Intensive Care Medicine \textbf{31}(10), 1345--1355 (2005)

\bibitem{morrow2000timi}
Morrow, D.A., Antman, E.M., Charlesworth, A., Cairns, R., Murphy, S.A.,
  de~Lemos, J.A., Giugliano, R.P., McCabe, C.H., Braunwald, E.: {TIMI} risk
  score for {ST}-elevation myocardial infarction: a convenient, bedside,
  clinical score for risk assessment at presentation an intravenous {nPA} for
  treatment of infarcting myocardium early {II} trial substudy.
\newblock Circulation \textbf{102}(17), 2031--2037 (2000)

\bibitem{neylon2006sparse}
Neylon, T.: Sparse solutions for linear prediction problems.
\newblock Ph.D. thesis, New York University (2006)

\bibitem{quinlan1986induction}
Quinlan, J.R.: Induction of decision trees.
\newblock Machine learning \textbf{1}(1), 81--106 (1986)

\bibitem{quinlan1993c4}
Quinlan, J.R.: C4. 5: programs for machine learning, vol.~1.
\newblock Morgan kaufmann (1993)

\bibitem{ranson1974prognostic}
Ranson, J., Rifkind, K., Roses, D., Fink, S., Eng, K., Spencer, F., et~al.:
  Prognostic signs and the role of operative management in acute pancreatitis.
\newblock Surgery, gynecology \& obstetrics \textbf{139}(1), 69 (1974)

\bibitem{pitfall}
Ridgeway, G.: The pitfalls of prediction.
\newblock NIJ Journal, {\rm National Institute of Justice} \textbf{271}, 34--40
  (2013)

\bibitem{rivest1987learning}
Rivest, R.L.: Learning decision lists.
\newblock Machine learning \textbf{2}(3), 229--246 (1987)

\bibitem{ruping2006learning}
R{\"u}ping, S.: Learning interpretable models.
\newblock Ph.D. thesis, Universit{\"a}t Dortmund (2006)

\bibitem{schwarz1978estimating}
Schwarz, G.: Estimating the dimension of a model.
\newblock The Annals of Statistics \textbf{6}(2), 461--464 (1978)

\bibitem{simon2011glmnet}
Simon, N., Friedman, J., Hastie, T., Tibshirani, R.: Regularization paths for
  cox's proportional hazards model via coordinate descent.
\newblock Journal of Statistical Software \textbf{39}(5), 1--13 (2011).
\newblock \urlprefix\url{http://www.jstatsoft.org/v39/i05/}

\bibitem{sommer1996theory}
Sommer, E.: Theory restructuring: A perspective on design and maintenance of
  knowledge based systems.
\newblock Ph.D. thesis, Universit{\"a}t Dortmund (1996)

\bibitem{steinhart2006juvenile}
Steinhart, D.: Juvenile detention risk assessment: A practice guide to juvenile
  detention reform.
\newblock Juvenile Detention Alternatives Initiative. A project of the Annie E.
  Casey Foundation. Retrieved on April \textbf{28}, 2011 (2006)

\bibitem{the2012rpart}
Therneau, T., Atkinson, B., Ripley, B.: rpart: Recursive Partitioning (2012).
\newblock \urlprefix\url{http://CRAN.R-project.org/package=rpart}.
\newblock R package version 4.1-0

\bibitem{tibshirani1996regression}
Tibshirani, R.: Regression shrinkage and selection via the lasso.
\newblock Journal of the Royal Statistical Society. Series B (Methodological)
  pp. 267--288 (1996)

\bibitem{tipping2001sparse}
Tipping, M.E.: Sparse bayesian learning and the relevance vector machine.
\newblock The Journal of Machine Learning Research \textbf{1}, 211--244 (2001)

\bibitem{turing2004intelligent}
Turing, A.: Intelligent machinery (1948).
\newblock B. Jack Copeland p. 395 (2004)

\bibitem{utgoff1989incremental}
Utgoff, P.E.: Incremental induction of decision trees.
\newblock Machine Learning \textbf{4}(2), 161--186 (1989)

\bibitem{vapnik1998statistical}
Vapnik, V.: Statistical Learning Theory.
\newblock Wiley, New York (1998)

\bibitem{VellidoEtAl12}
Vellido, A., Mart\'{i}n-Guerrero, J.D., Lisboa, P.J.: Making machine learning
  models interpretable.
\newblock In: Proc. European Symposium on Artificial Neural Networks,
  Computational Intelligence and Machine Learning (2012)

\bibitem{psych}
Webster, C.: Risk assessment: Actuarial instruments \& structured clinical
  guides (2013)

\bibitem{webster1995hcr}
Webster, C.D., Eaves, D.: The HCR-20 scheme: The assessment of dangerousness
  and risk.
\newblock Mental Health, Law and Policy Institute, Department of Psychology,
  Simon Fraser University and Forensic Psychiatric Services Commission of
  British Columbia (1995)

\bibitem{wells1997value}
Wells, P.S., Anderson, D.R., Bormanis, J., Guy, F., Mitchell, M., Gray, L.,
  Clement, C., Robinson, K.S., Lewandowski, B., et~al.: Value of assessment of
  pretest probability of deep-vein thrombosis in clinical management.
\newblock Lancet \textbf{350}(9094), 1795--1798 (1997)

\bibitem{wells2000derivation}
Wells, P.S., Anderson, D.R., Rodger, M., Ginsberg, J.S., Kearon, C., Gent, M.,
  Turpie, A., Bormanis, J., Weitz, J., Chamberlain, M., et~al.: Derivation of a
  simple clinical model to categorize patients probability of pulmonary
  embolism-increasing the models utility with the {SimpliRED D-dimer}.
\newblock Thrombosis and Haemostasis \textbf{83}(3), 416--420 (2000)

\bibitem{top10}
Wu, X., Kumar, V., Quinlan, R., Ghosh, J., Yang, Q., Motoda, H., Mclachlan, G.,
  Ng, A., Liu, B., Yu, P., Zhou, Z.H., Steinbach, M., Hand, D., Steinberg, D.:
  Top 10 algorithms in data mining.
\newblock Knowledge and Information Systems \textbf{14}(1), 1--37 (2008)

\bibitem{xu2001comparison}
Xu, L., Zhang, W.J.: Comparison of different methods for variable selection.
\newblock Analytica Chimica Acta \textbf{446}(1), 475--481 (2001)

\bibitem{yu2004efficient}
Yu, L., Liu, H.: Efficient feature selection via analysis of relevance and
  redundancy.
\newblock The Journal of Machine Learning Research \textbf{5}, 1205--1224
  (2004)

\bibitem{zhao2007model}
Zhao, P., Yu, B.: On model selection consistency of lasso.
\newblock Journal of Machine Learning Research \textbf{7}(2), 25--41 (2007)

\bibitem{zou2005regularization}
Zou, H., Hastie, T.: Regularization and variable selection via the elastic net.
\newblock Journal of the Royal Statistical Society: Series B (Statistical
  Methodology) \textbf{67}(2), 301--320 (2005)

\end{thebibliography}

\end{document}